\documentclass[12pt,fleqn]{ucithesis}
\usepackage[square,sort,comma,numbers]{natbib}
\makeatletter
\def\hlinew#1{%
	\noalign{\ifnum0=`}\fi\hrule \@height #1 \futurelet
	\reserved@a\@xhline}
\makeatother
\usepackage{wrapfig} 
\usepackage{lipsum} 
\usepackage{todonotes} 
\newcommand{\tabincell}[2]{\begin{tabular}{@{}#1@{}}#2\end{tabular}} 
\usepackage{url}
\usepackage{bm}
\usepackage{amsmath}
\usepackage{amssymb}
\usepackage{algorithm}
\usepackage[noend]{algpseudocode}
\makeatletter
\def\BState{\State\hskip-\ALG@thistlm}
\algnewcommand\algorithmicinput{\textbf{Input:}}
\algnewcommand\INPUT{\item[\algorithmicinput]}
\algnewcommand\algorithmicfulltrain{\hspace{3em} $\triangleright\triangleright\triangleright$\textbf{ Fully supervised training}}
\algnewcommand\FULLTRAIN{\item[\algorithmicfulltrain]}
\algnewcommand\algorithmicweaktrain{\hspace{3em} $\triangleright\triangleright\triangleright$\textbf{ Weakly supervised training}}
\algnewcommand\WEAKTRAIN{\item[\algorithmicweaktrain]}
\makeatother
\usepackage{amsmath}
\usepackage{amsthm}
\usepackage{array}
\usepackage{graphicx}
\usepackage{natbib}
\usepackage{relsize}

\usepackage{caption}
\usepackage{subcaption}  
\usepackage{multirow}
\usepackage{tabularx}

\usepackage[plainpages=false,pdfborder={0 0 0}]{hyperref}

\thesistitle{Deep Learning for Automated Medical Image Analysis}

\documenttitle{Dissertation}

\degreename{Doctor of Philosophy}

\degreefield{Computer Science}

\authorname{Wentao Zhu}

\committeechair{Professor Xiaohui Xie}
\othercommitteemembers
{
  Professor Charless C. Fowlkes \\
  Professor Sameer Singh
}

\degreeyear{2019}

\copyrightdeclaration
{
  {\copyright} {\Degreeyear} \Authorname
}

\dedications
{
  
  To my wife, little son and my parents
}

\acknowledgments
{
  First and foremost, I would like to thank my advisor, Professor Xiaohui Xie. Your guidance, expertise, insights, wisdom and optimism have been instrumental to my PhD career. Next, thank you to my dear wife, Dr. Yufang Huang. She gives me a family with my lovely little son. It is the greatest happiness being a husband and a father. And it is the kind of happiness that accompanies me during my PhD study. Thank you to my mother-in-law helping me take care of my little son. Thank you to my parents supporting me all the time. Thank you to Professor Charless C. Fowlkes, Professor Sameer Singh, Professor Jack Xin and Professor Ramesh Jain for your expertise and insights in my thesis and advancement. Finally, I want to thank all the friends I made and mentors during my PhD study: Yeeleng Scott Vang, Elmira Forouzmand, Yifei Chen, Cuiling Lan and Wenjun Zeng in Microsoft Research, Yi Wang and Xiaodi Hou in TuSimple, Chaochun Liu (now at JD.com) in Baidu Research, Wei Fan and Zhen Qian in Tencent Medical AI Labs, Le Lu in PingAn Tech, and many many other friends. It is your help and the friendship that support me in the fruitful and exciting PhD life.
  
  I came into UCI through the Computer Science PhD program. Thank you to everyone involved in the program a tremendous amount of gratitude providing such a high quality and standard program. 
  
  I gratefully acknowledge Breast Research Group, INESC Porto, Portugal, LUng Nodule Analysis 2016 (LUNA16), National Institute of Health (NIH), National Cancer Institute (NCI), The National Lung Screening Trial (NLST), The Cancer Imaging Archive (TCIA), Alibaba Cloud TianChi providing datasets for my PhD research. I gratefully acknowledge UCI Associated Graduate students (AGS), AAAI 2016, NIPS 2018, Tencent giving me travel grants and awards to attend conferences.
  
}

\newcommand{\mypubentry}[3]{
  \begin{tabular*}{1\textwidth}{@{\extracolsep{\fill}}p{4.5in}r}
    \textbf{#1} & \textbf{#2} \\ 
    \multicolumn{2}{@{\extracolsep{\fill}}p{.95\textwidth}}{#3}\vspace{6pt} \\
  \end{tabular*}
}

\curriculumvitae
{

\textbf{EDUCATION}
  
  \begin{tabular*}{1\textwidth}{@{\extracolsep{\fill}}lr}
    \textbf{Doctor of Philosophy in Computer Science} & \textbf{2019} \\
    \vspace{6pt}
    University of California, Irvine & \emph{Irvine, California} \\
    \textbf{Master of Science in Computer Science} & \textbf{2015} \\
    \vspace{6pt}
    Institute of Computing Technology, Chinese Academy of Sciences & \emph{Beijing, China} \\
    \textbf{Bachelor of Science in Computational Mathematics} & \textbf{2012} \\
    \vspace{6pt}
    Shandong University & \emph{Shandong, China} \\
  \end{tabular*}

\vspace{12pt}
\textbf{RESEARCH EXPERIENCE}

  \begin{tabular*}{1\textwidth}{@{\extracolsep{\fill}}lr}
    \textbf{Graduate Research Assistant} & \textbf{2015--2019} \\
    \vspace{6pt}
    University of California, Irvine & \emph{Irvine, California} \\
  \end{tabular*}

\vspace{12pt}
\textbf{TEACHING EXPERIENCE}

  \begin{tabular*}{1\textwidth}{@{\extracolsep{\fill}}lr}
    \textbf{Reader} & \textbf{2015--2017} \\
    \textbf{Reader} & \textbf{2019} \\
    \vspace{6pt}
    University of California, Irvine & \emph{Irvine, California} \\
  \end{tabular*}

\pagebreak

\textbf{REFEREED JOURNAL PUBLICATIONS}

  \mypubentry{AnatomyNet: Deep learning for fast and fully automated whole-volume segmentation of head and neck anatomy}{2018}{Medical Physics}

\vspace{12pt}
\textbf{REFEREED CONFERENCE PUBLICATIONS}

  \mypubentry{DeepEM: Deep 3D ConvNets with EM for weakly supervised pulmonary nodule detection}{September 2018}{MICCAI}
  \mypubentry{Deeplung: Deep 3d dual path nets for automated pulmonary nodule detection and classification}{March 2018}{WACV}
  \mypubentry{Adversarial deep structured nets for mass segmentation from mammograms}{April 2018}{ISBI}
  \mypubentry{Deep multi-instance networks with sparse label assignment for whole mammogram classification}{September 2017}{MICCAI}
  \mypubentry{Co-Occurrence feature learning for skeleton based action recognition using regularized deep LSTM networks}{February 2016}{AAAI}

\vspace{12pt}
\textbf{SOFTWARE}

  {\bf{AnatomyNet}}
  
  {https://github.com/wentaozhu/AnatomyNet-for-anatomical-segmentation}
  
  {Deep learning for fast and fully automated whole-volume segmentation of head and neck anatomy.}
  
  {\bf{DeepEM}}
  
  {https://github.com/wentaozhu/DeepEM-for-Weakly-Supervised-Detection}
  
  {Deep 3D ConvNets with EM for weakly supervised pulmonary nodule detection.}
  
  {\bf{DeepLung}}
  
  {https://github.com/wentaozhu/DeepLung}
  
  {Deep 3d dual path nets for automated pulmonary nodule detection and classification.}
  
  {\bf{Adversarial DSN}}
  
  {https://github.com/wentaozhu/adversarial-deep-structural-networks}
  
  {Adversarial Deep Structural Networks for Mammographic Mass Segmentation.}

  {\bf{Deep MIL}}
  
  {https://github.com/wentaozhu/deep-mil-for-whole-mammogram-classification}
  
  {Deep multi-instance networks with sparse label assignment for whole mammogram classification.}

  {\bf{Regularized Deep LSTM}}
  
  {http://www.escience.cn/system/file?fileId=87579}
  
  {Co-Occurrence feature learning for skeleton based action recognition using regularized deep LSTM networks.}

}

\thesisabstract
{
  Medical imaging is an essential tool in many areas of medical applications, used for both diagnosis and treatment. However, reading medical images and making diagnosis or treatment recommendations require specially trained medical specialists. The current practice of reading medical images is labor-intensive, time-consuming, costly, and error-prone. It would be more desirable to have a computer-aided system that can automatically make diagnosis and treatment recommendations.
  
  Recent advances in deep learning enable us to rethink the ways of clinician diagnosis based on medical images. Early detection has proven to be critical to give patients the best chance of recovery and survival. Advanced computer-aided diagnosis systems are expected to have high sensitivities and small low positive rates. How to provide accurate diagnosis results and explore different types of clinical data is an important topic in the current computer-aided diagnosis research.
  
  In this thesis, we will introduce 1) mammograms for detecting breast cancers, the most frequently diagnosed solid cancer for U.S. women, 2) lung Computed Tomography (CT) images for detecting lung cancers, the most frequently diagnosed malignant cancer, and 3) head and neck CT images for automated delineation of organs at risk in radiotherapy. First, we will show how to employ the adversarial concept to generate the hard examples improving mammogram mass segmentation. Second, we will demonstrate how to use the weakly labelled data for the mammogram breast cancer diagnosis by efficiently design deep learning for multi-instance learning. Third, the thesis will walk through DeepLung system which combines deep 3D ConvNets and Gradient Boosting Machine (GBM) for automated lung nodule detection and classification. Fourth, we will show how to use weakly labelled data to improve existing lung nodule detection system by integrating deep learning with a probabilistic graphic model. Lastly, we will demonstrate the AnatomyNet which is thousands of times faster and more accurate than previous methods on automated anatomy segmentation.
}

\hypersetup{
	pdftitle={\Thesistitle},
	pdfauthor={\Authorname},
	pdfsubject={\Degreefield},
}

\begin{document}

\preliminarypages

\chapter{Introduction} 
Deep learning has been a powerful and successful tool to lead the era of artificial intelligence (AI) in recent few years. It has achieved surprising or even over human-level performance on image classification \cite{he2016deep}, speech recognition \cite{xiong2018microsoft}, reading comprehension \cite{devlin2018bert} etc. On the other hand, autonomous driving, personal assistant devices (Google home, Alexa etc), AI for health-care  are emerging topics with the aforementioned breakthroughs from deep learning. In this thesis, we focus on several topics, including mammograms, pulmonary computed tomography (CT) images, head and neck CT images, and deep learning for automated medical image analysis which potentially assists radiologists to improve the diagnosis. 

Breast cancer is the second common cancer for women in United States, and mammogram screening is shown to be an effective way to reduce the death caused by breast cancer \cite{oeffinger2015breast}. However, traditional computer-aided diagnosis system is of a high false positive rate. The main difficulty for deep learning based medical image computing research is that the scarcity of data leads to easily over-fitting \cite{dhungel2015deep}. The thesis tries to alleviate the challenge from two directions. Firstly, we try to improve the generalization ability for deep learning based segmentation \cite{zhu2017adversarial}, which is an essential problem for medical image analysis because the mass segmentation produces morphological features which provide effective evidence for mammogram diagnosis. Secondly, compared with detection or segmentation ground-truth, image-level labels can be easily obtained from electronic medical report (EMR). We design a novel framework for breast cancer diagnosis using image-level label. The framework is different from traditional methods relying on regions of interest (ROIs) which require great efforts to annotate \cite{dhungel2016automated,zhu2017deep}.

Lung cancer is the cancer causing most death. Computed tomography (CT) screening is an effective way for accurate and early diagnosis which significantly improves the survival rate. Recently deep learning has been witnessing widespread adoption in various medical image applications. Considering CT images are spatial 3D data, we firstly construct 3D convolutional neural networks for lung CT nodule analysis system which includes nodule detection and classification \cite{zhu2018deeplung}. However, training complicated deep neural nets requires large-scale datasets labeled with ground truth, which are often unavailable or costly and difficult to scale in many medical image domains. On the other hand, electronic medical records (EMR) consists of plenty of partial information on the content of each medical image. We explore how to tap this vast amount of underutilized weakly labeled data from electronic medical records (EMR) to improve pulmonary nodule detection \cite{zhu2018deepem}.

Radiation therapy (RT) is a common treatment option for head and neck (HaN) cancer. An important step involved in RT planning is the delineation of organs-at-risks (OARs) based on HaN computed tomography (CT). However, manually delineating OARs is time-consuming as each slice of CT images needs to be individually examined and a typical CT consists of hundreds of slices. Automating OARs segmentation has the benefit of both reducing the time and improving the quality of RT planning. Existing anatomy auto-segmentation algorithms use primarily atlas-based methods, which are time-consuming, require sophisticated atlas creation and cannot adequately account for anatomy variations among patients. Designing a fast automated OARs delineation system with few radiologist's corrections can reduce the delay time for treatment and potentially improve the performance of radiotherapy.

\section{Dissertation Outline and Contributions}
The thesis is outlined as follows:

In Chapter 2, we propose a novel end-to-end network for mammographic mass segmentation which employs a fully convolutional network (FCN) to model a potential function, followed by a conditional random field (CRF) to perform structured learning \cite{zhu2017adversarial}. Because the mass distribution varies greatly with pixel position in the ROIs, the FCN is combined with a position priori. Further, we employ adversarial training to eliminate over-fitting due to the small sizes of mammogram datasets. Multi-scale FCN is employed to improve the segmentation performance. Experimental results on two public datasets, INbreast and DDSM-BCRP, demonstrate that our end-to-end network achieves better performance than state-of-the-art approaches. These contributions are released as an open-source software package called adversarial-deep-structural-networks, which is publicly available\footnote{https://github.com/wentaozhu/adversarial-deep-structural-networks}. Portions of this chapter were published as part of \cite{zhu2017adversarial}.

In Chapter 3, we propose end-to-end trained deep multi-instance networks for mass classification based on whole mammogram without the aforementioned ROIs \cite{zhu2017deep}, inspired by the success of using deep convolutional features for natural image analysis and multi-instance learning (MIL) for labeling a set of instances/patches. We explore three different schemes to construct deep multi-instance networks for whole mammogram classification. Experimental results on the INbreast dataset demonstrate the robustness of proposed networks compared to previous work using segmentation and detection annotations. These contributions are released as an open-source software package called deep-mil-for-whole-mammogram-classification, which is publicly available\footnote{https://github.com/wentaozhu/deep-mil-for-whole-mammogram-classification}. Portions of this chapter were published as part of \cite{zhu2017deep}.

In Chapter 4, we present a fully automated lung computed tomography (CT) cancer diagnosis system, DeepLung \cite{zhu2018deeplung}. DeepLung consists of two components, nodule detection (identifying the locations of candidate nodules) and classification (classifying candidate nodules into benign or malignant). Considering the 3D nature of lung CT data and the compactness of dual path networks (DPN), two deep 3D DPN are designed for nodule detection and classification respectively. Specifically, a 3D Faster Region Convolutional Neural Net (R-CNN) is designed for nodule detection with 3D dual path blocks and a U-net-like encoder-decoder structure to effectively learn nodule features. For nodule classification, gradient boosting machine (GBM) with 3D dual path network features is proposed. The nodule classification subnetwork is validated on a public dataset from LIDC-IDRI, on which it achieves better performance than the state-of-the-art approaches and surpasses the performance of experienced doctors based on image modality. Within the DeepLung system, candidate nodules are detected first by the nodule detection subnetwork, and nodule diagnosis is conducted by the classification subnetwork. Extensive experimental results demonstrate that DeepLung has performance comparable to the experienced doctors both for the nodule-level and patient-level diagnosis on the LIDC-IDRI dataset. These contributions are released as an open-source software package called DeepLung, which is publicly available\footnote{https://github.com/wentaozhu/DeepLung}. Portions of this chapter were published as part of \cite{zhu2018deeplung}.

In Chapter 5, we propose DeepEM, a novel deep 3D ConvNet framework augmented with expectation-maximization (EM), to mine weakly supervised labels in EMRs for pulmonary nodule detection \cite{zhu2018deepem}. Experimental results show that DeepEM can lead to 1.5\% and 3.9\% average improvement in free-response receiver operating characteristic (FROC) scores on LUNA16 and Tianchi datasets, respectively, demonstrating the utility of incomplete information in EMRs for improving deep learning algorithms. These contributions are released as an open-source software package called DeepEM, which is publicly available\footnote{https://github.com/wentaozhu/DeepEM-for-Weakly-Supervised-Detection}. Portions of this chapter were published as part of \cite{zhu2018deepem}.

In Chapter 6, we propose an end-to-end, atlas-free 3D convolutional deep learning framework for fast and fully automated whole-volume HaN anatomy segmentation \cite{zhu2018anatomynet}. Our deep learning model, called AnatomyNet, segments OARs from head and neck CT images in an end-to-end fashion, receiving whole-volume HaN CT images as input and generating masks of all OARs of interest in one shot. AnatomyNet is built upon the popular 3D U-net architecture, but extends it in three important ways: 1) a new encoding scheme to allow auto-segmentation on whole-volume CT images instead of local patches or subsets of slices, 2) incorporating 3D squeeze-and-excitation residual blocks in encoding layers for better feature representation, and 3) a new loss function combining Dice scores and focal loss to facilitate the training of the neural model. These features are designed to address two main challenges in deep-learning-based HaN segmentation: a)  segmenting small anatomies (i.e., optic chiasm and optic nerves) occupying only a few slices, and b) training with inconsistent data annotations with missing ground truth for some anatomical structures. We collect 261 HaN CT images to train AnatomyNet, and use MICCAI Head and Neck Auto Segmentation Challenge 2015 as a benchmark dataset to evaluate the performance of AnatomyNet. The objective is to segment nine anatomies: brain stem, chiasm, mandible, optic nerve left, optic nerve right, parotid gland left, parotid gland right, submandibular gland left, and submandibular gland right. Compared to previous state-of-the-art results from the MICCAI 2015 competition, AnatomyNet increases Dice similarity coefficient by 3.3\% on average. AnatomyNet takes about 0.12 seconds to fully segment a head and neck CT image of  dimension  $178 \times 302 \times 225$, significantly faster than previous methods. In addition, the model is able to process  whole-volume CT images and delineate all OARs in one pass, requiring little pre- or post-processing. We demonstrate that our proposed model can improve segmentation accuracy and simplify the auto-segmentation pipeline. These contributions are released as an open-source software package called AnatomyNet, which is publicly available\footnote{https://github.com/wentaozhu/AnatomyNet-for-anatomical-segmentation}. Portions of this chapter were published as part of \cite{zhu2018anatomynet}. 
\chapter{Adversarial Deep Structured Nets for Mass Segmentation from Mammograms}

\section{Introduction}
\label{1sec:intro}
According to the American Cancer Society, breast cancer is the most frequently diagnosed solid cancer and the second leading cause of cancer death among U.S. women~\cite{acs}. Mammogram screening has been demonstrated to be an effective way for early detection and diagnosis, which can significantly decrease breast cancer mortality~\cite{oeffinger2015breast}. Mass segmentation provides morphological features, which play crucial roles for diagnosis. 

Traditional studies on mass segmentation rely heavily on hand-crafted features. Model-based methods build classifiers and learn features from masses \cite{beller2005example,cardoso2015closed}. There are few works using deep networks for mammogram~\cite{deepmil}. Dhungel et al. employed multiple deep belief networks (DBNs), Gaussian mixture model (GMM) classifier and a priori as potential functions, and structured support vector machine (SVM) to perform segmentation \cite{dhungel2015deep}. They further used CRF with tree re-weighted belief propagation to boost the segmentation performance \cite{dhungel2015tree}. A recent work used the output from a convolutional network (CNN) as a complimentary potential function, yielding the state-of-the-art performance~\cite{dhungel2015deepmiccai}. However, the two-stage training used in these methods produces potential functions that easily over-fit the training data.

\begin{figure}[t]
	\begin{center}
		\begin{minipage}{\linewidth}
			\centerline{\includegraphics[width=6cm]{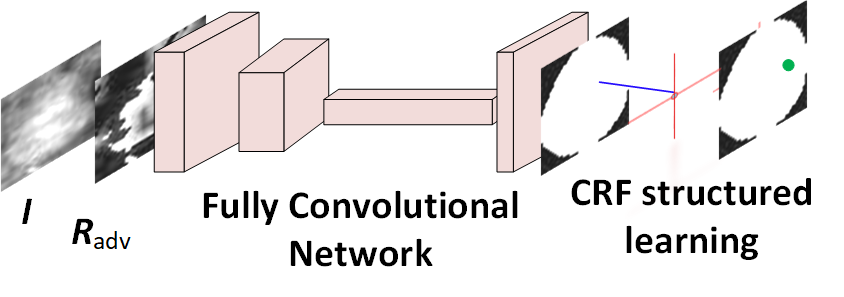}}
		\end{minipage}
		\caption{The proposed adversarial deep FCN-CRF network with four convolutional layers followed by CRF for structured learning.}
		\label{1framework}
	\end{center}
\end{figure}

In this work, we propose an end-to-end trained adversarial deep structured network to perform mass segmentation (Fig. \ref{1framework}). The proposed network is designed to robustly learn from a small dataset with poor contrast mammographic images. Specifically, an end-to-end trained FCN with CRF is applied. Adversarial training is introduced into the network to learn robustly from scarce mammographic images. Different from DI2IN-AN using a generative framework \cite{yang2017automatic}, we directly optimize pixel-wise labeling loss. To further explore statistical property of mass regions, a spatial priori is integrated into FCN. We validate the adversarial deep structured network on two public mammographic mass segmentation datasets. The proposed network is demonstrated to outperform other algorithms for mass segmentation consistently. 

Our main contributions in this work are: (1) We propose an unified end-to-end training framework integrating FCN+CRF and adversarial training. (2) We employ an end-to-end network to do mass segmentation while previous works require a lot of hand-designed features or multi-stage training. (3) Our model achieves the best results on two most commonly used mammographic mass segmentation datasets.
\section{FCN-CRF Network} 
Fully convolutional network (FCN) is a commonly used model for image segmentation, which consists of convolution, transpose convolution, or pooling \cite{long2015fully}. For training, the FCN optimizes maximum likelihood loss function 
\begin{equation}
\mathcal{L}_{FCN} = -\frac{1}{N\times N_i} \sum_{n=1}^{N}\sum_{i=1}^{N_i} \log p_{fcn}{(y_{n,i} | \textbf{I}_n; \bm \theta)},
\end{equation}
where $y_{n,i}$ is the label of $i$th pixel in the $n$th image $\textbf{I}_n$, $N$ is the number of training mammograms, $N_i$ is the number of pixels in the image, and $\bm \theta$ is the parameter of FCN. Here the size of images is fixed to $40 \times 40$ and $N_i$ is 1,600. 

CRF is a classical model for structured learning, well suited for image segmentation. It models pixel labels as random variables in a Markov random field conditioned on an observed input image. To make the annotation consistent, we use $\textbf{y} = (y_1, y_2, \dots, y_i, \dots, y_{1,600})^T$ to denote the random variables of pixel labels in an image, where $y_i\in\{0,1\}$. The zero denotes pixel belonging to background, and one denotes it belonging to mass region. The Gibbs energy of fully connected pairwise CRF is \cite{krahenbuhl2011efficient}
\begin{equation}
E(\textbf{y}) = \sum_{i} \psi_u (y_i) + \sum_{i<j} \psi_p (y_i, y_j),
\end{equation}
where unary potential function $\psi_u (y_i)$ is the loss of FCN in our case, pairwise potential function $\psi_p (y_i, y_j)$ defines the cost of labeling pair $(y_i, y_j)$,
\begin{equation}
\label{1equ:pairwisepotential}
\psi_p (y_i, y_j) = \mu (y_i, y_j) \sum_{m} w^{(m)} k_G^{(m)}(\textbf{f}_i, \textbf{f}_j),
\end{equation}
where label compatibility function $\mu$ is given by the Potts model in our case, $w^{(m)}$ is the learned weight, pixel values $I_i$ and positions $p_i$ can be used as the feature vector $\textbf{f}_i$, $k_G^{(m)}$ is the Gaussian kernel applied to feature vectors \cite{krahenbuhl2011efficient},
\begin{equation}
\label{1equ:gaussiankernel}
k_G(\textbf{f}_i, \textbf{f}_j) = [\exp(-|I_i - I_j|^2/2), \exp(-|p_i - p_j|^2)/2]^T.
\end{equation} 
Efficient inference algorithm can be obtained by mean field approximation $q(\textbf{y}) = \prod_i q_i(y_i) $ \cite{krahenbuhl2011efficient}. The update rule is
\begin{equation}
\label{1equ:messagepassing}
\begin{aligned}
&\tilde{q}_i^{(m)}(l)\leftarrow \sum_{i\neq j}k_G^{(m)}(\textbf{f}_i,\textbf{f}_j)q_j(l) \text{ for all } m,\\
&\check{q}_i(l)\leftarrow \sum_m w^{(m)}\tilde{q}_i^{(m)}(l), \\
&\hat{q}_i(l)\leftarrow \sum_{l^\prime\in\mathcal{L}}\mu(l,l^\prime)\check{q}_i(l),\\
&\breve{q}_i(l)\leftarrow \exp(- \psi_u (y_i = l))-\hat{q}_i(l), \\
&Z_i \leftarrow \sum_{l\in \mathcal{L}} \exp\left(\breve{q}_i(l)\right), \\
&q_i\leftarrow \frac{1}{Z_i}\exp\left(\breve{q}_i(l)\right),
\end{aligned}
\end{equation}
where the first equation is the message passing from label of pixel $i$ to label of pixel $j$, the second equation is re-weighting with the learned weights $w^{(m)}$, the third equation is compatibility transformation, the fourth equation is adding unary potentials, and the last step is normalization. Here $\mathcal{L} = \{0,1\}$ denotes background or mass. The initialization of inference employs unary potential function as $q_i(y_i) = \frac{1}{Z_i} \exp(- \psi_u (y_i))$. The mean field approximation can be interpreted as a recurrent neural network (RNN) \cite{zheng2015conditional}. 
\section{Adversarial FCN-CRF Nets}
\begin{figure}[t]
	\begin{center}
		\begin{minipage}{0.3\linewidth}
			\centerline{\includegraphics[width=1cm]{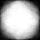}	\includegraphics[width=1cm]{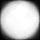}}
			\center{(a)}
		\end{minipage}
		\begin{minipage}{0.6\linewidth}
			\centerline{
				\includegraphics[width=1cm]{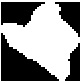}
				\includegraphics[width=1cm]{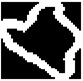}
				\includegraphics[width=1cm]{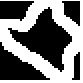}}
			\center{(b)}
		\end{minipage}
		\caption{The empirical estimation of a priori on INbreast (left) and DDSM-BCRP (right) training datasets (a). Trimap visualizations on the DDSM-BCRP dataset, segmentation groundtruth (first column), trimap of width $2$ (second column), trimaps of width $3$ (third column) (b).}
		\label{1prior}
	\end{center}
\end{figure}

The shape and appearance prior play important roles in mammogram mass segmentation \cite{jiang2016mammographic,dhungel2015deepmiccai}. The distribution of labels varies greatly with position in the mammographic mass segmentation. From observation, most of the masses are located in the center of region of interest (ROI), and the boundary areas of ROI are more likely to be background (Fig. \ref{1prior}(a)).

The conventional FCN provides independent pixel-wise predictions. It considers global class distribution difference corresponding to bias in the last layer. Here we employ a priori for position into consideration 
\begin{equation}
p(y_i | \textbf{I}; \bm \theta) \propto w_i p_{fcn}{(y_i | \textbf{I}; \bm \theta)},
\end{equation}
where $w_i$ is the empirical estimation of mass varied with the pixel position $i$, and $p_{fcn}{(y_i | \textbf{I}; \bm \theta)}$ is the predicted mass probability of conventional FCN. In the implementation, we added an image sized bias in the softmax layer as the empirical estimation of mass for FCN to train network. The $-\log p(y_i | \textbf{I}; \bm{\theta})$ is used as the unary potential function for $\psi_u (y_i)$ in the CRF as RNN. For multi-scale FCN as potential functions, the potential function is defined as $\psi_u (y_i) = \sum_{u ^{\prime}} w_{(u ^{\prime})} \psi_{u ^{\prime}}(y_i)$, where $w_{(u ^{\prime})}$ is the learned weight for unary potential function, $\psi_{u ^{\prime}}(y_i)$ is the  potential function provided by FCN of each scale. 

Adversarial training provides strong regularization for deep networks. The idea of adversarial training is that if the model is robust enough, it should be invariant to small perturbations of training examples that yield the largest increase in the loss (adversarial examples \cite{szegedy2013intriguing}). The perturbation $\bm{R}$ can be obtained as $\min_{\textbf{R}, \| \textbf{R} \| \leq \epsilon} \log p(\bm{y} | \textbf{I} + \textbf{R}; \bm \theta)$. In general, the calculation of exact $\bm R$ is intractable especially for complicated models such as deep networks. The linear approximation and $L_2$ norm box constraint can be used for the calculation of perturbation as $\bm{R}_{adv} = - \frac{\epsilon \bm{g}}{\|\bm{g}\|_2}$,
where $\bm{g} = \nabla_{\bm{I}} \log p(\bm{y} | \bm{I}; \bm{\theta})$. For adversarial FCN, the network predicts label of each pixel independently as $p(\bm{y}| \bm{I}; \bm{\theta}) = \prod_{i} p(y_i | \bm{I}; \bm{\theta})$. For adversarial CRF as RNN, the prediction of network relies on mean field approximation inference as $p(\bm{y}| \bm{I}; \bm{\theta}) = \prod_{i} q(y_i | \bm{I}; \bm{\theta})$. 

The adversarial training forces the model to fit examples with the worst perturbation direction. The adversarial loss is 
\begin{equation}
\label{1equ:adversarialloss}
\mathcal{L}_{adv}(\bm{\theta}) = - \frac{1}{N} \sum_{n=1}^{N} \log p(\bm{y}_n | \bm{I}_n + \bm{R}_{adv,n}; \bm{\theta}).
\end{equation}
In the back-propagation, we block the further calculation of gradient of $\bm{R}$ to avoid Hessian computing. In training, the total loss is defined as the sum of adversarial loss and the empirical loss based on training samples as 
\begin{equation}
\label{1equ:totalloss}
\mathcal{L}(\bm{\theta}) = \mathcal{L}_{adv}(\bm{\theta})- \frac{1}{N} \sum_{n=1}^{N} \log p(\bm{y}_n | \bm{I}_n; \bm{\theta}) + \frac{\lambda}{2} \| \bm{\theta} \|^2,
\end{equation}
where $\lambda$ is the \(l_2\) regularization factor for \(\bm \theta \), $p(\bm{y}_n | \bm{I}_n; \bm{\theta})$ is either mass probability prediction in the FCN or a posteriori approximated by mean field inference in the CRF as RNN for the $n$th image $\bm{I}_n$. 
\section{Experiments}\label{1sec:exp}
We validate the proposed model on two most commonly used public mammographic mass segmentation datasets: INbreast \cite{moreira2012inbreast} and DDSM-BCRP dataset \cite{heath1998current}. We use the same ROI extraction and resize principle as \cite{dhungel2015deep,dhungel2015deepmiccai,dhungel2015tree}. Due to the low contrast of mammograms, image enhancement technique is used on the extracted ROI images as the first 9 steps in \cite{ball2007digital}, followed by pixel position dependent normalization. The preprocessing makes training converge quickly. We further augment each training set by flipping horizontally, flipping vertically, flipping horizontally and vertically, which makes the training set 4 times larger than the original training set. 

For consistent comparison, the Dice index metric is used to evaluate segmentation performance and is defined as $\frac{2 \times TP}{2 \times TP + FP + FN}$. For a fair comparison, we re-implement a two-stage model \cite{dhungel2015deepmiccai}, and obtain similar result (Dice index $0.9010$) on the INbreast dataset. 
\begin{itemize}
	\item FCN is the network integrating a position priori into FCN (denoted as FCN 1 in Table \ref{1tab:network}).
	\item Adversarial FCN is FCN  with adversarial training.
	\item Joint FCN-CRF is the FCN followed by CRF as RNN with an end-to-end training scheme. 
	\item Adversarial FCN-CRF is the Jointly FCN-CRF with end-to-end adversarial training. 
	\item Multi-FCN, Adversarial multi-FCN, Joint multi-FCN-CRF, Adversarial multi-FCN-CRF employ 4 FCNs with multi-scale kernels, which can be trained in an end-to-end way using the last prediction. 
\end{itemize}
The prediction of Multi-FCN, Adversarial multi-FCN is the average prediction of the 4 FCNs. The configurations of FCNs are in Table~\ref{1tab:network}. Each convolutional layer is followed by \(2\times2\) max pooling. The last layers of the four FCNs are all two $40\times40$ transpose convolution kernels with soft-max activation function. We use hyperbolic tangent activation function in middle layers. The parameters of FCNs are set such that the number of each layer's parameters is almost the same as that of CNN used in the work \cite{dhungel2015deepmiccai}. We use Adam with learning rate 0.003. The $\lambda$ is $0.5$ in the two datasets. The $\epsilon$ used in adversarial training are $0.1$ and $0.5$ for INbreast and DDSM-BCRP datasets respectively. Because the boundaries of masses on the DDSM-BCRP dataset are smoother than those on the INbreast dataset, we use larger perturbation $\epsilon$. For the CRF as RNN, we use 5 time steps in the training and 10 time steps in the test phase empirically. 
\begin{table}[t]
	\fontsize{9pt}{10pt}\selectfont\centering
	\caption{Kernel sizes of sub-nets (\#kernel$\times$\#width$\times$\#height).}\label{1tab:network}
	\begin{tabular}{c|c|c|c}
		\hlinew{0.9pt}
		Net.&First layer&Second layer&Third layer\\
		\hline
		FCN 1&$6\times5\times5$&$12\times5\times5$ conv.&$588\times7\times7$\\
		\hline
		FCN 2&$9\times4\times4$&$12\times4\times4$ conv.&$588\times7\times7$\\
		\hline
		FCN 3&$16\times3\times3$& $13\times3\times3$ conv.&$415\times8\times8$\\
		\hline
		FCN 4&$37\times2\times2$&$12\times2\times2$ conv.&$355\times9\times9$\\
		\hlinew{0.9pt}
	\end{tabular}
\end{table}

\begin{table}[t]
	\fontsize{9pt}{10pt}\selectfont\centering
	\caption{Dices (\%) on INbreast and DDSM-BCRP datasets.}\label{1tab:inbreast}
	\begin{tabular}{c|c|c}
		\hlinew{0.9pt}
		Methodology&INbreast&DDSM-BCRP\\		
		\hlinew{0.7pt}
		\tabincell{c}{Cardoso et al. \cite{cardoso2015closed}}&88&N/A \\
		\hline
		\tabincell{c}{Beller et al. \cite{beller2005example}}&N/A&70\\
		\hline
		\tabincell{c}{Deep Structure Learning \cite{dhungel2015deep}}&88&87\\
		\hline
		\tabincell{c}{TRW Deep Structure Learning \cite{dhungel2015tree}}&89&89\\
		\hline
		\tabincell{c}{Deep Structure Learning + CNN \cite{dhungel2015deepmiccai}}&90&90\\
		\hlinew{0.9pt}
		FCN & 89.48 & 90.21 \\
		\hline
		\tabincell{c}{ Adversarial FCN} & 89.71 & 90.78\\
		\hline
		\tabincell{c}{Joint FCN-CRF} & 89.78 & 90.97\\
		\hline
		\tabincell{c}{ Adversarial FCN-CRF}& 90.07 & 91.03\\
		\hline
		\tabincell{c}{Multi-FCN} & 90.47 & 91.17\\
		\hline
		\tabincell{c}{Adversarial multi-FCN} & 90.71 & 91.20\\
		\hline
		\tabincell{c}{Joint multi-FCN-CRF} & 90.76 & 91.26\\
		\hline
		\tabincell{c}{Adversarial multi-FCN-CRF} & \textbf{90.97} & \textbf{91.30}\\
		\hlinew{0.9pt}
	\end{tabular}
\end{table}
The INbreast dataset is a recently released mammographic mass analysis dataset, which provides more accurate contours of lesion region and the mammograms are of high quality. For mass segmentation, the dataset contains 116 mass regions. We use the first 58 masses for training and the rest for test, which is of the same protocol as \cite{dhungel2015deep,dhungel2015deepmiccai,dhungel2015tree}. The DDSM-BCRP dataset contains 39 cases (156 images) for training and 40 cases (160 images) for testing~\cite{heath1998current}. After ROI extraction, there are 84 ROIs for training, and 87 ROIs for test. We compare schemes with other recently published mammographic mass segmentation methods in Table \ref{1tab:inbreast}. 

Table \ref{1tab:inbreast} shows the CNN features provide superior performance on mass segmentation, outperforming hand-crafted feature based methods \cite{cardoso2015closed,beller2005example}. Our enhanced FCN achieves 0.25\% Dice index improvement than the traditional FCN on the INbreast dataset. The adversarial training yields 0.4\% improvement on average. Incorporating the spatially structured learning further produces 0.3\% improvement. Using multi-scale model contributes the most to segmentation results, which shows multi-scale features are effective for pixel-wise classification in mass segmentation. Combining all the components together achieves the best performance with 0.97\%, 1.3\% improvement on INbreast, DDSM-BCRP datasets respectively. The possible reason for the improvement is adversarial scheme eliminates the over-fitting.
We calculate the p-value of McNemar’s Chi-Square Test to compare our model with ~\cite{dhungel2015deepmiccai} on the INbreast dataset. We obtain p-value $< 0.001$, which shows our model is significantly better than
model~\cite{dhungel2015deepmiccai}. 

\begin{figure}[t]
	\begin{center}
		\begin{minipage}{\linewidth}
			\centerline{\includegraphics[width=8cm]{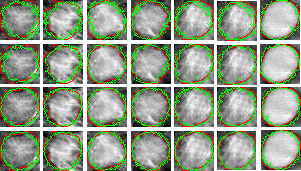}}
		\end{minipage}
		\caption{Visualization of segmentation results using the FCN (first row), Adversarial FCN (second row), Joint FCN-CRF (third row), Adversarial FCN-CRF (fourth row) on the test sets of INbreast dataset. Each column denotes a test sample. Red lines denote the ground truth. Green lines or points denote the segmentation results. Adversarial training provides sharper and more accurate segmentation boundaries than methods without adversarial training.}
		\label{1fig:all}
	\end{center}
\end{figure}

To better understand the adversarial training, we visualize segmentation results in Fig. \ref{1fig:all}. We observe that the segmentations in the second and fourth rows have more accurate boundaries than those of the first and third rows. It demonstrates the adversarial training improves FCN and FCN-CRF.

\begin{figure}[t]
	\begin{center}
		\begin{minipage}{0.46\linewidth}
			\centerline{\includegraphics[width=4.713cm]{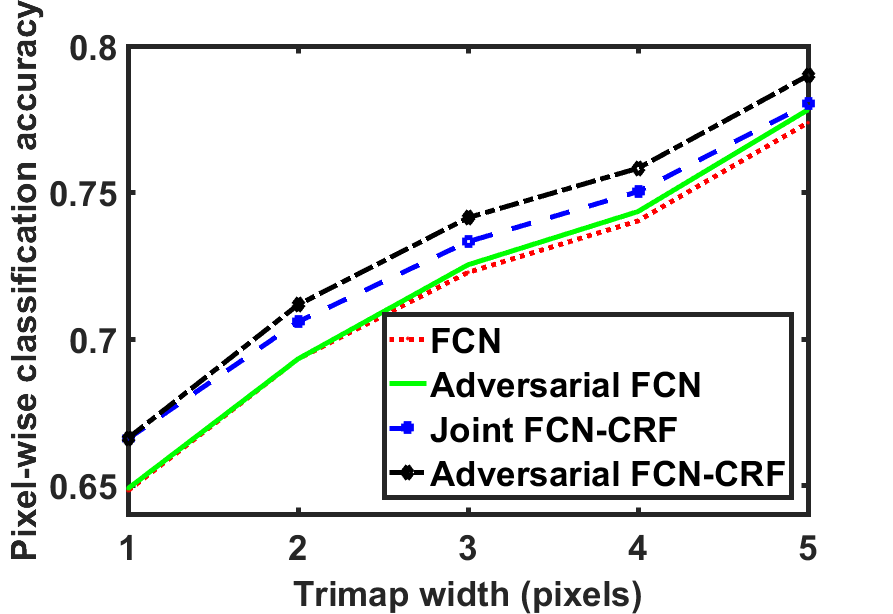}}
			\center{(a)}
		\end{minipage}
		\hspace{0.45cm}
		\begin{minipage}{0.46\linewidth}
			\centerline{\includegraphics[width=4.713cm]{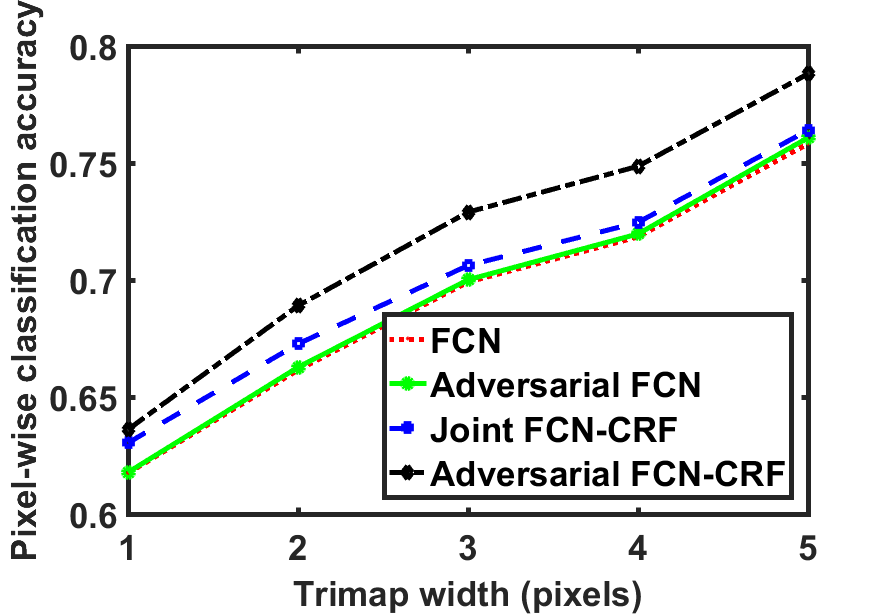}}
			\center{(b)}
		\end{minipage}
		\caption{Accuracy comparisons among FCN, Adversarial FCN, Joint FCN-CRF and Adversarial FCN-CRF in trimaps with pixel width $1$, $2$, $3$, $4$, $5$ on the INbreast dataset (a) and the DDSM-BCRP dataset (b). The adversarial training improves segmentation accuracy around boundaries.}
		\label{1fig:trimapinbreast}
	\end{center}
\end{figure}

We further employ the prediction accuracy based on trimap to specifically evaluate segmentation accuracy in boundaries \cite{kohli2009robust}. We calculate the accuracies within trimap surrounding the actual mass boundaries (groundtruth) in Fig. \ref{1fig:trimapinbreast}. Trimaps on the DDSM-BCRP dataset is visualized in Fig.~\ref{1prior}(b). From the figure, accuracies of Adversarial FCN-CRF are 2-3 \% higher than those of Joint FCN-CRF on average and the accuracies of Adversarial FCN are better than those of FCN. The above results demonstrate that the adversarial training improves the FCN and Joint FCN-CRF both for whole image and boundary region segmentation. 
\section{Conclusion}\label{1sec:con}
In this work, we propose an end-to-end adversarial FCN-CRF network for mammographic mass segmentation. To integrate the priori distribution of masses and fully explore the power of FCN, a position priori is added to the network. Furthermore, adversarial training is used to handle the small size of training data by reducing over-fitting and increasing robustness. Experimental results demonstrate the superior performance of adversarial FCN-CRF on two commonly used public datasets.


\chapter{Deep Multi-instance Networks with Sparse Label Assignment for Whole Mammogram Classification}

\section{Introduction}\label{2sec:intro}
Traditional mammogram classification requires extra annotations such as bounding box for detection or mask ground truth for segmentation~\cite{varela2006use,carneiro2015unregistered,jiao2016deep}. Other work have employed different deep networks to detect ROIs and obtain mass boundaries in different stages \cite{dhungel2016automated}. However, these methods require hand-crafted features to complement the system~\cite{kooi2017large}, and training data to be annotated with bounding boxes and segmentation ground truth which require expert domain knowledge and costly effort to obtain. In addition, multi-stage training cannot fully explore the power of deep networks. 

Due to the high cost of annotation, we intend to perform classification based on a raw whole mammogram. Each patch of a mammogram can be treated as an instance and a whole mammogram is treated as a bag of instances. The whole mammogram classification problem can then be thought of as a standard MIL problem. Due to the great representation power of deep features~\cite{greenspan2016guest,zhu2016adversarial,zhu2016co}, combining MIL with deep neural networks is an emerging topic. Yan et al. used a deep MIL to find discriminative patches for body part recognition~\cite{yan2016multi}. Patch based CNN added a new layer after the last layer of deep MIL to learn the fusion model for multi-instance predictions~\cite{hou2015patch}. Shen et al. employed two stage training to learn the deep multi-instance networks for pre-detected lung nodule classification~\cite{shen2016learning}. The above approaches used max pooling to model the general multi-instance assumption which only considers the patch of max probability. In this paper, more effective task-related deep multi-instance models with end-to-end training are explored for whole mammogram classification. We investigate three different schemes, i.e., max pooling, label assignment, and sparsity, to perform deep MIL for the whole mammogram classification task.
\begin{figure}[t]
	\setlength{\abovecaptionskip}{0.cm}
	\setlength{\belowcaptionskip}{-0.cm}
	\begin{center}
		\begin{minipage}{0.8\linewidth}
			\centerline{\includegraphics[width=\textwidth]{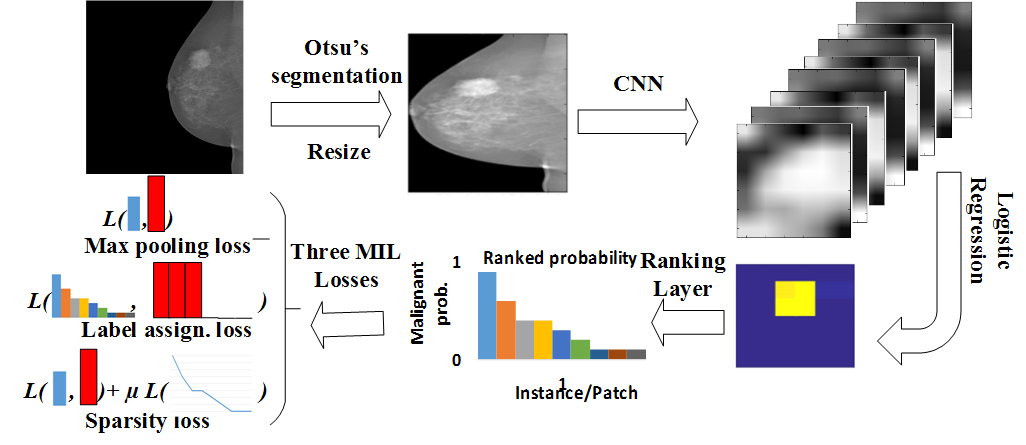}}
		\end{minipage}
		\caption{The framework of whole mammogram classification. First, we use Otsu's segmentation to remove the background and resize the mammogram to $227\times227$. Second, the deep MIL accepts the resized mammogram as input to the convolutional layers. Here we use the convolutional layers in AlexNet~\cite{krizhevsky2012imagenet}. Third, logistic regression with weight sharing over different patches is used to quantify the probability of malignancy of each position from the convolutional neural network (CNN) feature maps of high channel dimensions. Then the responses of the instances/patches are ranked. Lastly, the learning loss is calculated using max pooling loss, label assignment, or sparsity loss for the three different schemes.}
		\label{2fig:framework}
	\end{center}
\end{figure}
\begin{figure}[t]
	\setlength{\abovecaptionskip}{0.cm}
	\setlength{\belowcaptionskip}{-0.cm}
	\begin{center}
		\begin{minipage}{0.245\linewidth}
			\centerline{\includegraphics[width=0.9\linewidth]{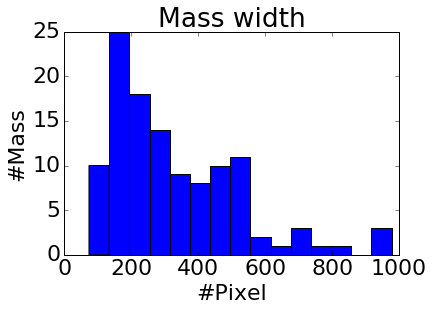}}
			\center{(a)}
		\end{minipage}
			\hspace{0.5cm}
		\begin{minipage}{0.245\linewidth}
			\centerline{\includegraphics[width=0.9\linewidth]{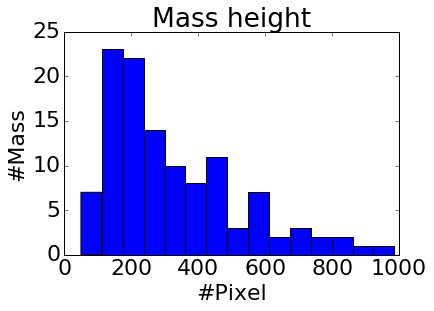}}
			\center{(b)}
		\end{minipage}
		\vfill
		\begin{minipage}{0.245\linewidth}
			\centerline{\includegraphics[width=0.9\linewidth]{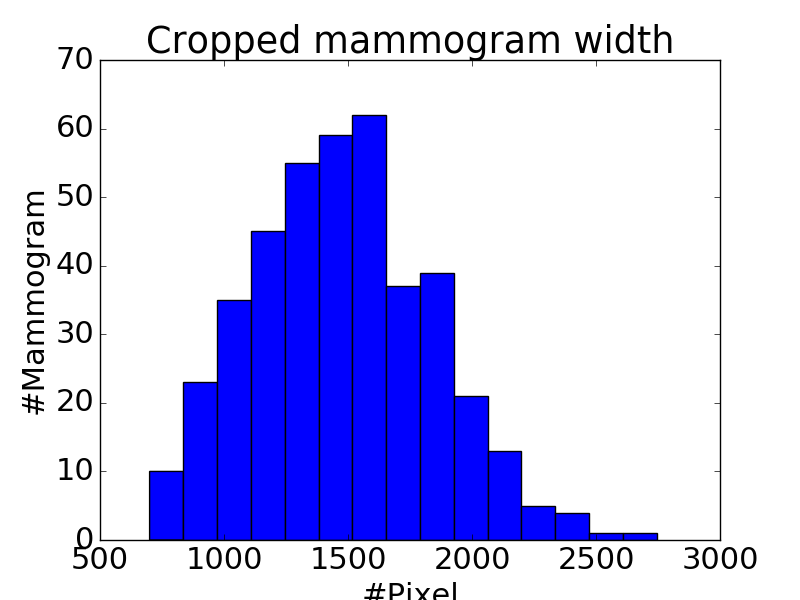}}
			\center{(c)}
		\end{minipage}
		\hspace{0.5cm}
		\begin{minipage}{0.245\linewidth}
			\centerline{\includegraphics[width=0.9\linewidth]{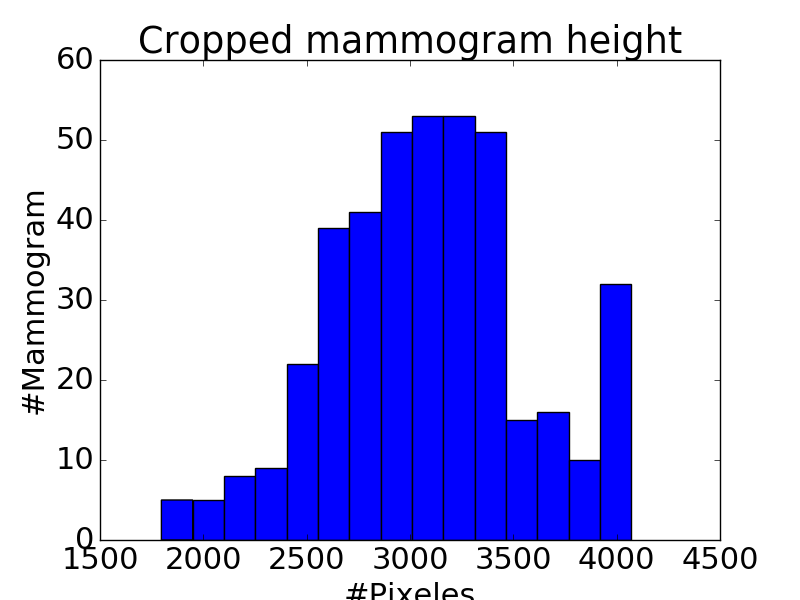}}
			\center{(d)}
		\end{minipage}
		\caption{Histograms of mass width (a) and height (b), mammogram width (c) and height (d). Compared to the size of whole mammogram ($1,474 \times 3,086$ on average after cropping), the mass of average size ($329 \times 325$) is tiny, and takes about 2\% of a whole mammogram. }
		\label{2fig:mass}
	\end{center}
\end{figure}

The framework for our proposed end-to-end trained deep MIL for mammogram classification is shown in Fig.~\ref{2fig:framework}. To fully explore the power of deep MIL, we convert the traditional MIL assumption into a label assignment problem. As a mass typically composes only 2\% of a whole mammogram (see Fig.~\ref{2fig:mass}), we further propose sparse deep MIL. The proposed deep multi-instance networks are shown to provide robust performance for whole mammogram classification on the INbreast dataset~\cite{moreira2012inbreast}.
\section{Deep MIL for Whole Mammogram Mass Classification}\label{2sec:dmlmamm}
Unlike other deep multi-instance networks~\cite{yan2016multi,hou2015patch}, we use a CNN to efficiently obtain features of all patches (instances) at the same time. Given an image $\bm{I}$, we obtain a feature map $\bm{F}$ of multi-channels $N_c$ after multiple convolutional layers and max pooling layers. The $(\bm{F})_{i,j,:}$ represents deep CNN features for a patch $\bm{Q}_{i,j}$ in $\bm{I}$, where $i,j$ represents the pixel row and column index respectively, and the ``$:$'' denotes the channel dimension. 

The goal of our work is to predict whether a whole mammogram contains a malignant mass (BI-RADS\footnote{https://breast-cancer.ca/bi-rads/} $\in \{4, 5, 6\}$ are considered positive examples) or not, which is a standard binary classification problem. We add a logistic regression with weights shared across all the pixel positions following $\bm{F}$, and an element-wise sigmoid activation function is applied to the output. To clarify it, the malignant probability of feature space's pixel $(i,j)$ is
\begin{equation}
\label{2equ:cnn}
r_{i,j} = \text{sigmoid}(\bm{a} \cdot \bm{F}_{i,j,:} + b),
\end{equation}
where $\bm{a}$ is the weights in logistic regression, $b$ is the bias, and $\cdot$ is the inner product of the two vectors $\bm{a}$ and $\bm{F}_{i,j,:}$. The $\bm{a}$ and $b$ are shared for different pixel positions $i,j$. We can combine $r_{i,j}$ into a matrix $\bm{r} = (r_{i,j})$ of range $[0, 1]$ denoting the probabilities of patches being malignant masses. The $\bm{r}$ can be flattened into a one-dimensional vector as $\bm{r} = (r_{1}, r_{2}, ..., r_{m})$ corresponding to flattened patches $(\bm{Q}_{1}, \bm{Q}_{2}, ..., \bm{Q}_{m})$, where $m$ is the number of patches.
\subsection{Max Pooling-Based Multi-Instance Learning}\label{2sec:maxpool}
The general multi-instance assumption is that if there exists an instance that is positive, the bag is positive~\cite{dietterich1997solving}. The bag is negative if and only if all instances are negative. For whole mammogram classification, the equivalent scenario is that if there exists a malignant mass, the mammogram $\bm{I}$ should be classified as positive. Likewise, negative mammogram $\bm{I}$ should not have any malignant masses. If we treat each patch $\bm{Q}_{i}$ of $\bm{I}$ as an instance, the whole mammogram classification is a standard multi-instance task.

For negative mammograms, we expect all the $r_i$ to be close to 0. For positive mammograms, at least one $r_i$ should be close to 1. Thus, it is natural to use the maximum component of $\bm{r}$ as the malignant probability of the mammogram $\bm{I}$
\begin{equation}
\label{2equ:max}
p(y=1|\bm{I}, \bm{\theta}) = \max\{ r_1, r_2, ..., r_m\},
\end{equation}
where $\bm{\theta}$ is the weights in deep networks.

If we sort $\bm{r}$ first in descending order as illustrated in Fig.~\ref{2fig:framework}, the malignant probability of the whole mammogram $\bm{I}$ is the first element of ranked $\bm{r}$ as
\begin{equation}
\label{2equ:sortmax}
\begin{aligned}
&\{{r^\prime}_1, {r^\prime}_2, ..., {r^\prime}_m\} = \text{sort} (\{ r_1, r_2, ..., r_m\}), \\
&p(y=1|\bm{I}, \bm{\theta}) = {r^\prime}_1, \quad\text{and}\quad p(y=0|\bm{I}, \bm{\theta}) = 1-{r^\prime}_1,
\end{aligned}
\end{equation}
where $\bm{r}^\prime = ({r^\prime}_1, {r^\prime}_2, ..., {r^\prime}_m)$ is descending ranked $\bm{r}$. The cross entropy-based cost function can be defined as  
\begin{equation}
\label{2equ:maxloss}
\mathcal{L}_{maxpooling} = -\frac{1}{N}\sum_{n=1}^{N} \log(p(y_n | \bm{I}_n, \bm{\theta})) + \frac{\lambda}{2} \|\bm{\theta}\|^2
\end{equation}
where $N$ is the total number of mammograms, $y_n \in \{0,1\}$ is the true label of malignancy for mammogram $\bm{I}_n$, and $\lambda$ is the regularizer that controls model complexity.

One disadvantage of max pooling-based MIL is that it only considers the patch ${\bm{Q}^\prime}_1$ (patch of the max malignant probability), and does not exploit information from other patches. A more powerful framework should add task-related prior, such as sparsity of mass in whole mammogram, into the general multi-instance assumption and explore more patches for training. 
\subsection{Label Assignment-Based Multi-Instance Learning}\label{2sec:labelassign}
For the conventional classification tasks, we assign a label to each data point. In the MIL scheme, if we consider each instance (patch) $\bm{Q}_i$ as a data point for classification, we can convert the multi-instance learning problem into a label assignment problem.

After we rank the malignant probabilities $\bm{r} = (r_{1}, r_{2}, ..., r_{m})$ for all the instances (patches) in a whole mammogram $\bm{I}$ using the first equation in Eq.~\ref{2equ:sortmax}, the first few ${r^\prime}_i$ should be consistent with the label of whole mammogram as previously mentioned, while the remaining patches (instances) should be negative. Instead of adopting the general MIL assumption that only considers the ${\bm{Q}^\prime}_1$ (patch of malignant probability ${r^\prime}_1$), we assume that 1) patches of the first $k$ largest malignant probabilities $\{{r^\prime}_1, {r^\prime}_2, ..., {r^\prime}_k\}$ should be assigned with the same class label as that of whole mammogram, and 2) other patches should be labeled as negative in the label assignment-based MIL.

After ranking/sorting using the first equation in Eq.~\ref{2equ:sortmax}, we can obtain the malignant probability for each patch
\begin{equation}
\label{2equ:ppatch}
\begin{aligned}
p(y=1 | {\bm{Q}^\prime}_i, \bm{\theta}) = {r^\prime}_i, \quad\text{and}\quad p(y=0 | {\bm{Q}^\prime}_i, \bm{\theta}) = 1-{r^\prime}_i.
\end{aligned}
\end{equation}

The cross entropy loss function of the label assignment-based MIL can be defined
\begin{equation}
\label{2equ:weightedlabelloss}
\begin{aligned}
\mathcal{L}_{labelassign.} = &-\frac{1}{mN}\sum_{n=1}^{N}  \bigg ( \sum_{j=1}^{k} {\log(p(y_n | {\bm{Q}^\prime}_j, \bm{\theta}))}+ \\&\sum_{j=k+1}^{m} {\log(p(y=0 | {\bm{Q}^\prime}_j, \bm{\theta}))}\bigg )+\frac{\lambda}{2} \|\bm{\theta}\|^2.
\end{aligned}
\end{equation}

One advantage of the label assignment-based MIL is that it explores all the patches to train the model. Essentially it acts a kind of data augmentation which is an effective technique to train deep networks when the training data is scarce. From the sparsity perspective, the optimization problem of label assignment-based MIL is exactly a $k$-sparse problem for the positive data points, where we expect $\{{r^\prime}_1, {r^\prime}_2, ..., {r^\prime}_k\}$ being 1 and $\{{r^\prime}_{k+1}, {r^\prime}_{k+2}, ..., {r^\prime}_m\}$ being 0. The disadvantage of label assignment-based MIL is that it is hard to estimate the hyper-parameter $k$. Thus, a relaxed assumption for the MIL or an adaptive way to estimate the hyper-parameter $k$ is preferred. 
\subsection{Sparse Multi-Instance Learning}\label{2sec:sparse}
From the mass distribution, the mass typically comprises about 2\% of the whole mammogram on average (Fig.~\ref{2fig:mass}), which means the mass region is quite sparse in the whole mammogram. It is straightforward to convert the mass sparsity to the malignant mass sparsity, which implies that $\{{r^\prime}_1, {r^\prime}_2, ..., {r^\prime}_m\}$ is sparse in the whole mammogram classification problem. The sparsity constraint means we expect the malignant probability of part patches ${r^ \prime}_i$ being 0 or close to 0, which is equivalent to the second assumption in the label assignment-based MIL. Analogously, we expect ${r^\prime}_1$ to be indicative of the true label of mammogram $\bm{I}$.

After the above discussion, the loss function of sparse MIL problem can be defined
\begin{equation}
\label{2equ:weightedsparseloss}
\mathcal{L}_{sparse} = \frac{1}{N}\sum_{n=1}^{N} \big ( -\log(p(y_n | \bm{I}_n, \bm{\theta})) + \mu \|\bm{r}^{\prime}_n\|_1 \big ) +\frac{\lambda}{2} \|\bm{\theta}\|^2,
\end{equation}
where $p(y_n | \bm{I}_n, \bm{\theta})$ can be calculated in Eq.~\ref{2equ:sortmax}, $\bm{r}_n = ({r^\prime}_1, {r^\prime}_2, ..., {r^\prime}_m)$ for mammogram $\bm{I}_n$, $\|\cdot\|_1$ denotes the $\mathcal{L}_1$ norm, $\mu$ is the sparsity factor, which is a trade-off between the sparsity assumption and the importance of patch ${\bm{Q}^\prime}_1$.

From the discussion of label assignment-based MIL, this learning is a kind of exact $k$-sparse problem which can be converted to $\mathcal{L}_1$ constrain. One advantage of sparse MIL over label assignment-based MIL is that it does not require assign label for each patch which is hard to do for patches where probabilities are not too large or small. The sparse MIL considers the overall statistical property of $\bm{r}$.

Another advantage of sparse MIL is that, it has different weights for general MIL assumption (the first part loss) and label distribution within mammogram (the second part loss), which can be considered as a trade-off between max pooling-based MIL (slack assumption) and label assignment-based MIL (hard assumption).
\section{Experiments}\label{2sec:exp}
We validate the proposed models on the most frequently used mammographic mass classification dataset, INbreast dataset~\cite{moreira2012inbreast}, as the mammograms in other datasets, such as DDSM dataset~\cite{bowyer1996digital}, are of low quality. The INbreast dataset contains 410 mammograms of which 100 containing malignant masses. These 100 mammograms with malignant masses are defined as positive. For fair comparison, we also use 5-fold cross validation to evaluate model performance as~\cite{dhungel2016automated}. For each testing fold, we use three folds for training, and one fold for validation to tune hyper-parameters. The performance is reported as the average of five testing results obtained from cross-validation.

We employ techniques to augment our data. For each training epoch, we randomly flip the mammograms horizontally, shift within 0.1 proportion of mammograms horizontally and vertically, rotate within 45 degree, and set $50 \times 50$ square box as 0. In experiments, the data augmentation is essential for us to train the deep networks.

For the CNN network structure, we use AlexNet and remove the fully connected layers~\cite{krizhevsky2012imagenet}. Through CNN, the mammogram of size $227 \times 227$ becomes 256 $6 \times 6$ feature maps. Then we use steps in Sec.~\ref{2sec:dmlmamm} to do MIL. Here we employ weights pretrained on the ImageNet due to the scarce of data. We use Adam optimization with learning rate $5 \times 10^{-5}$ for training models~\cite{ba2015adam}. The $\lambda$ for max pooling-based and label assignment-based MIL are $1 \times 10^{-5}$. The $\lambda$ and $\mu$ for sparse MIL are $5 \times 10^{-6}$ and $1 \times 10^{-5}$ respectively. For the label assignment-based MIL, we select $k$ from $\{1,2,4,6,8\}$ based on the validation set and do not fix the $k$ on different rounds of cross validation. 

We firstly compare our methods to previous models validated on DDSM dataset and INbreast dataset in Table~\ref{2tab:inbreast}. Previous hand-crafted feature-based methods require manually annotated detection bounding box or segmentation ground truth even in test denoting as manual~\cite{ball2007digital,varela2006use,domingues2012inbreast}. The feat. denotes requiring hand-crafted features. Pretrained CNN uses two CNNs to detect the mass region and segment the mass, followed by a third CNN to do  mass classification on the detected ROI region, which requires hand-crafted features to pretrain the network and needs multi-stages training\cite{dhungel2016automated}. Pretrained CNN+Random Forest further employs random forest and obtained 7\% improvement. These methods are either manually or need hand-crafted features or multi-stages training, while our methods are totally automated, do not require hand-crafted features or extra annotations even on training set, and can be trained in an end-to-end manner.
\begin{table}[t]
	\fontsize{9pt}{10pt}\selectfont\centering
	\caption{Accuracy Comparisons of the proposed deep MILs and related methods on test sets.}\label{2tab:inbreast}
	\begin{tabular}{c|c|c|c|c}
		\hlinew{0.9pt}
		Methodology&Dataset&Set-up&Accu.&AUC\\		
		\hlinew{0.7pt}
		\tabincell{c}{Ball et al. \cite{ball2007digital}}&DDSM&Manual+feat.&0.87&N/A\\
		\hline \tabincell{c}{Varela et al. \cite{varela2006use}}&DDSM&Manual+feat.&0.81&N/A\\
		\hline \tabincell{c}{Domingues et al. \cite{domingues2012inbreast}}&INbr.&Manual+feat.&0.89&N/A\\
		\hline \tabincell{c}{Pretrained CNN \cite{dhungel2016automated}}&INbr.&Auto.+feat.&0.84$\pm{0.04}$&0.69$\pm{0.10}$\\
		\hline \tabincell{c}{Pretrained CNN+Random Forest \cite{dhungel2016automated}}&INbr.&Auto.+feat.&$\bf{0.91\pm{0.02}}$&0.76$\pm{0.23}$\\
		\hlinew{0.9pt}
		AlexNet &INbr.&Auto.&0.81$\pm{0.02}$&0.79$\pm{0.03}$\\
		\hline \tabincell{c}{AlexNet+Max Pooling MIL} &INbr.&Auto.&0.85$\pm{0.03}$&0.83$\pm{0.05}$\\
		\hline \tabincell{c}{AlexNet+Label Assign. MIL} &INbr.&Auto.&0.86$\pm{0.02}$&0.84$\pm{0.04}$\\
		\hline \tabincell{c}{AlexNet+Sparse MIL} &INbr.&Auto.&0.90$\pm{0.02}$&$\bf{0.89\pm{0.04}}$\\
		\hlinew{0.9pt}
	\end{tabular}
\end{table}

The max pooling-based deep MIL obtains better performance than the pretrained CNN using 3 different CNNs and detection/segmentation annotation in the training set. This shows the superiority of our end-to-end trained deep MIL for whole mammogram classification. According to the accuracy metric, the sparse deep MIL is better than the label assignment-based MIL, which is better than the max pooling-based MIL. This result is consistent with previous discussion that the sparsity assumption benefited from not having hard constraints of the label assignment assumption, which employs all the patches and is more efficient than max pooling assumption. Our sparse deep MIL achieves competitive accuracy to random forest-based pretrained CNN, while much higher AUC than previous work, which shows our method is more robust. The main reasons for the robust results using our models are as follows. Firstly, data augmentation is an important technique to increase scarce training datasets and proves useful here. Secondly, the transfer learning that employs the pretrained weights from ImageNet is effective for the INBreast dataset. Thirdly, our models fully explore all the patches to train our deep networks thereby eliminating any possibility of overlooking malignant patches by only considering a subset of patches. This is a distinct advantage over previous networks that employ several stages consisting of detection and segmentation. 
\begin{figure}[!t]
	\setlength{\abovecaptionskip}{0.cm}
	\setlength{\belowcaptionskip}{-0.cm}
	\begin{center}
		\begin{minipage}{0.15\linewidth}
			\centerline{\includegraphics[width=2.5cm]{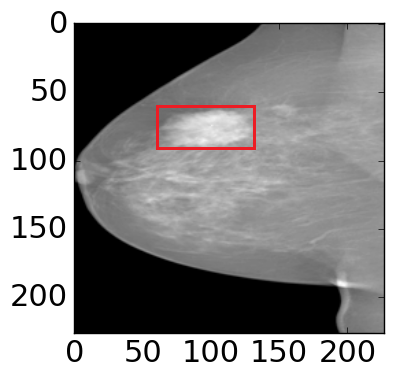}}
			\center{(a)}
		\end{minipage}
		\hspace{1cm}
		\begin{minipage}{0.15\linewidth}
			\centerline{\includegraphics[width=2.5cm]{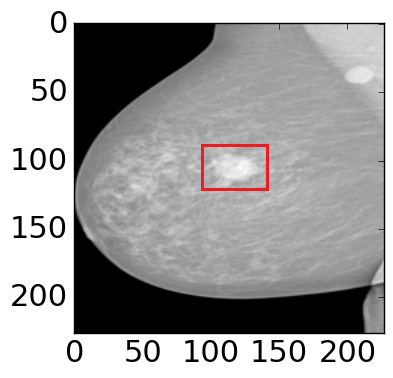}}
			\center{(b)}
		\end{minipage}
		\hspace{1cm}
		\begin{minipage}{0.15\linewidth}
			\centerline{\includegraphics[width=2.5cm]{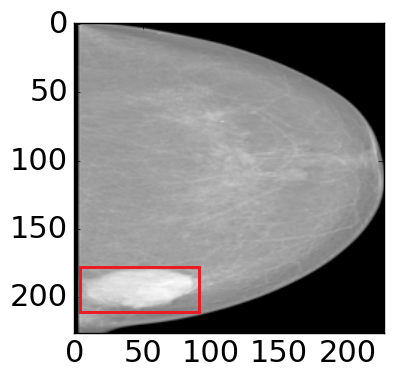}}
			\center{(c)}
		\end{minipage}
		\hspace{1cm}
		\begin{minipage}{0.15\linewidth}
			\centerline{\includegraphics[width=2.5cm]{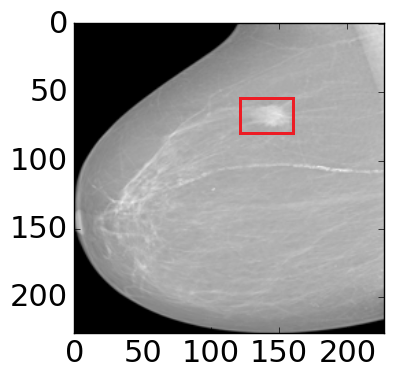}}
			\center{(d)}
		\end{minipage}
		\vfill
		\begin{minipage}{0.15\linewidth}
			\centerline{\includegraphics[width=2.5cm]{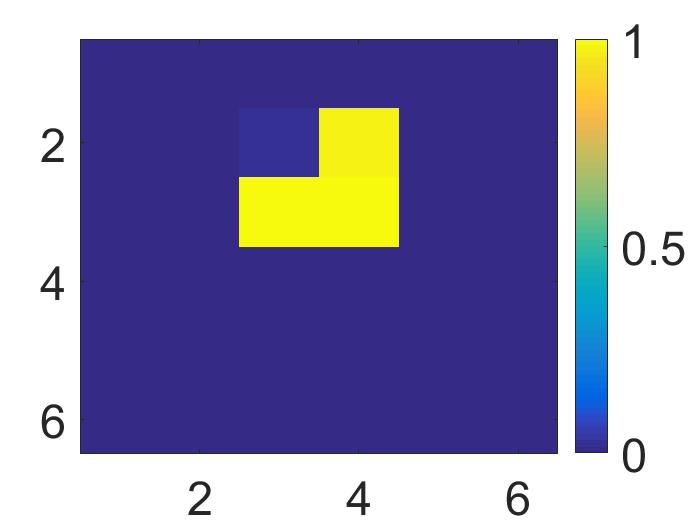}}
		\end{minipage}
		\hspace{1cm}
		\begin{minipage}{0.15\linewidth}
			\centerline{\includegraphics[width=2.5cm]{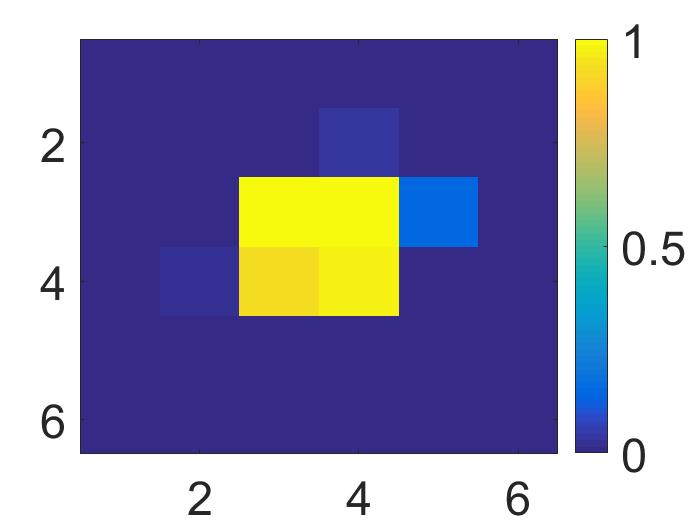}}
		\end{minipage}
		\hspace{1cm}
		\begin{minipage}{0.15\linewidth}
			\centerline{\includegraphics[width=2.5cm]{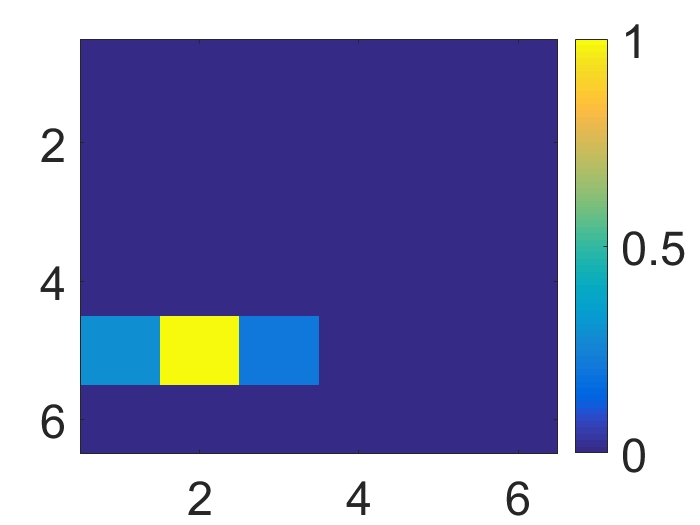}}
		\end{minipage}
		\hspace{1cm}
		\begin{minipage}{0.15\linewidth}
			\centerline{\includegraphics[width=2.5cm]{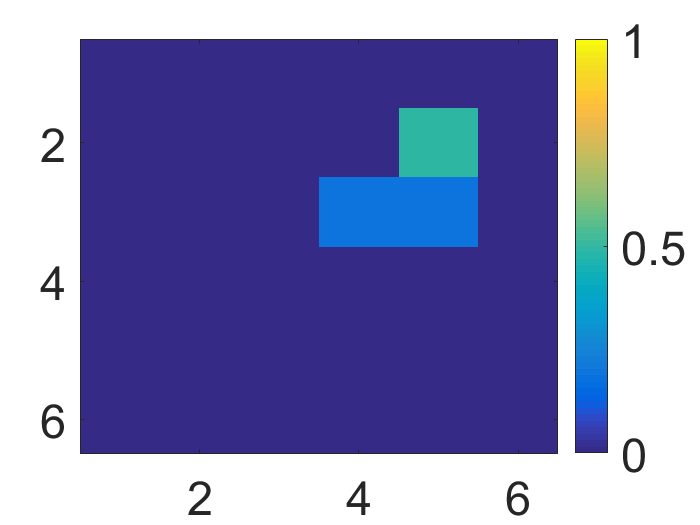}}
		\end{minipage}
		\vfill
		\begin{minipage}{0.15\linewidth}
			\centerline{\includegraphics[width=2.6cm]{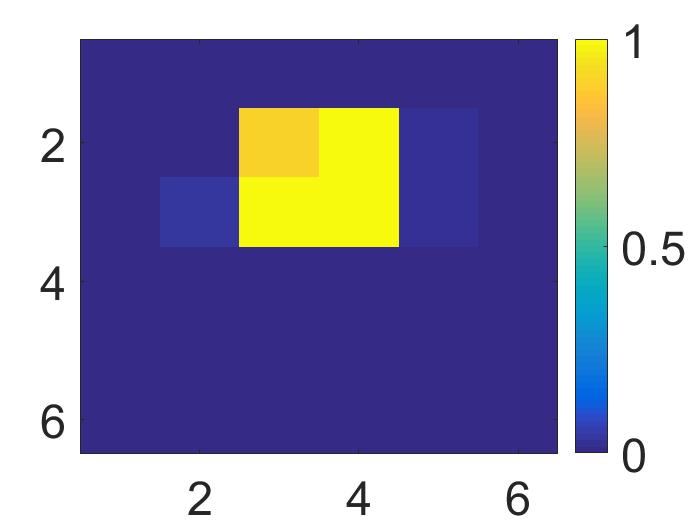}}
		\end{minipage}
		\hspace{1cm}
		\begin{minipage}{0.15\linewidth}
			\centerline{\includegraphics[width=2.6cm]{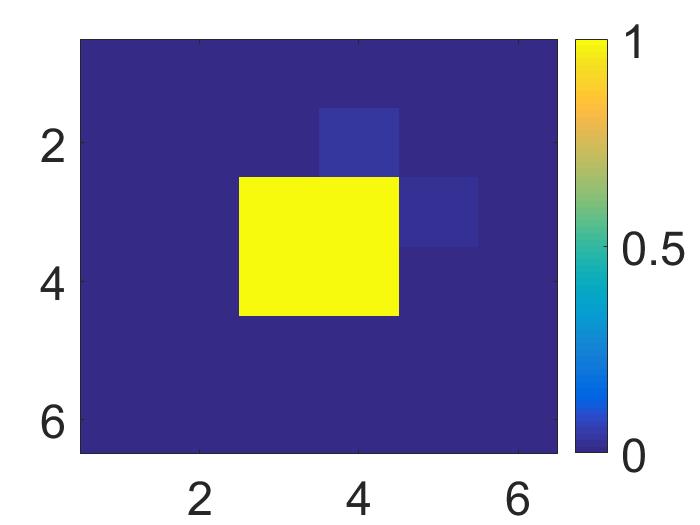}}
		\end{minipage}
		\hspace{1cm}
		\begin{minipage}{0.15\linewidth}
			\centerline{\includegraphics[width=2.6cm]{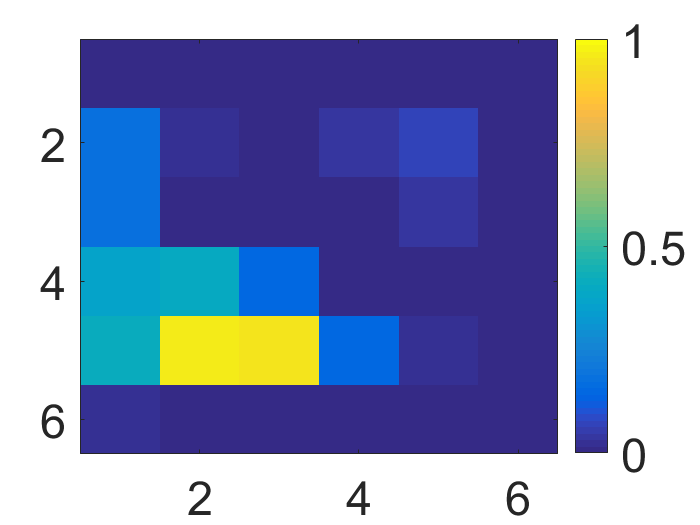}}
		\end{minipage}
		\hspace{1cm}
		\begin{minipage}{0.15\linewidth}
			\centerline{\includegraphics[width=2.6cm]{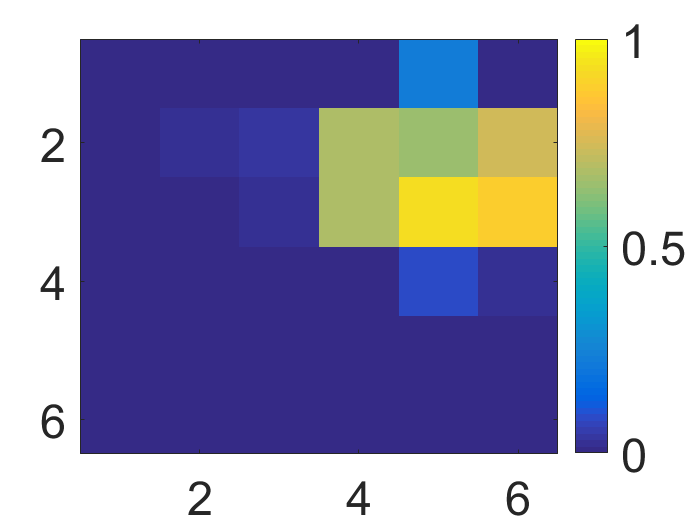}}
		\end{minipage}
		\vfill
		\begin{minipage}{0.15\linewidth}
			\centerline{\includegraphics[width=2.6cm]{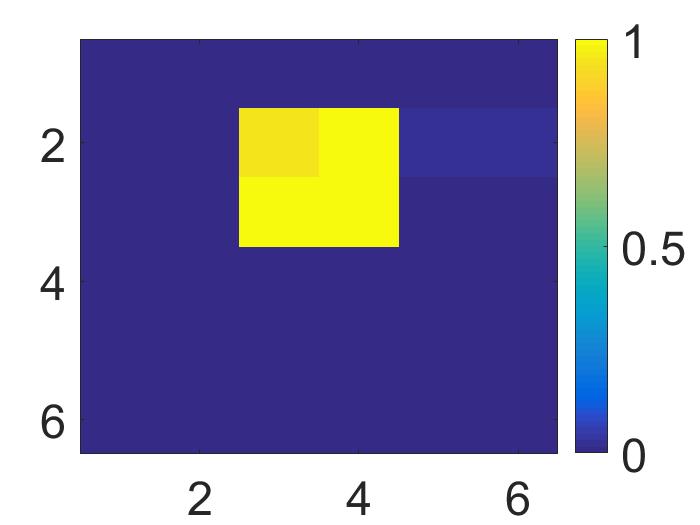}}
		\end{minipage}
		\hspace{1cm}
		\begin{minipage}{0.15\linewidth}
			\centerline{\includegraphics[width=2.6cm]{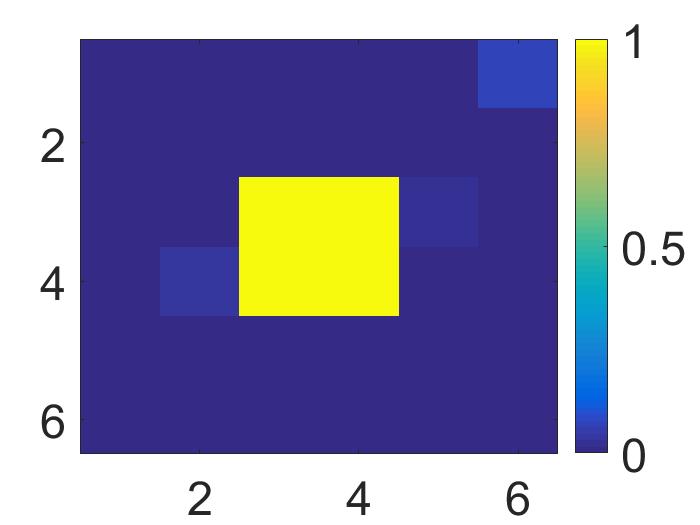}}
		\end{minipage}
		\hspace{1cm}
		\begin{minipage}{0.15\linewidth}
			\centerline{\includegraphics[width=2.6cm]{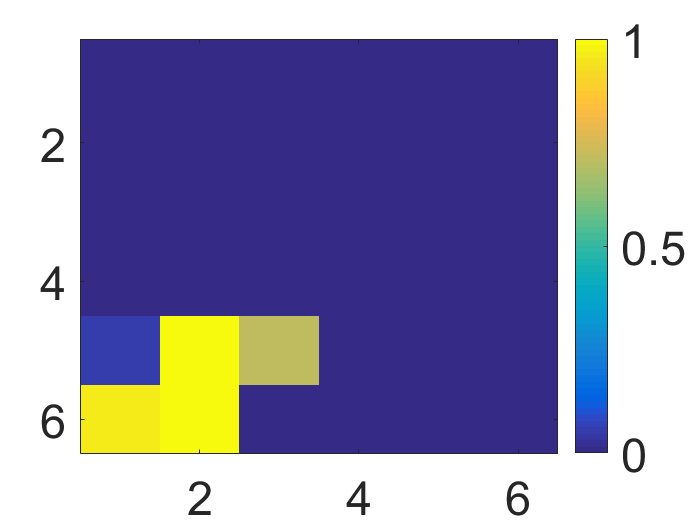}}
		\end{minipage}
		\hspace{1cm}
		\begin{minipage}{0.15\linewidth}
			\centerline{\includegraphics[width=2.6cm]{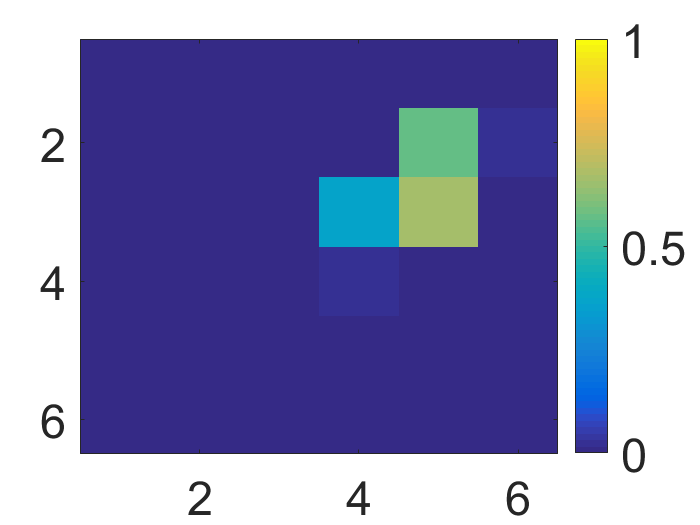}}
		\end{minipage}
		\caption{The visualization of predicted malignant probabilities for instances/patches in four resized mammograms. The first row is the resized mammogram. The red rectangle boxes are mass regions from the annotations on the dataset. The color images from the second row to the last row are the predicted malignant probability from logistic regression layer for (a) to (d) respectively, which are the malignant probabilities of patches/instances. Max pooling-based, label assignment-based, sparse deep MIL are in the second row, third row, fourth row respectively.}
		\label{2fig:visresponse}
	\end{center}
\end{figure}

To further understand our deep MIL, we visualize the responses of logistic regression layer for four mammograms on test set, which represents the malignant probability of each patch, in Fig.~\ref{2fig:visresponse}. We can see the deep MIL learns not only the prediction of whole mammogram, but also the prediction of malignant patches within the whole mammogram. Our models are able to learn the mass region of the whole mammogram without any explicit bounding box or segmentation ground truth annotation of training data. The max pooling-based deep multi-instance network misses some malignant patches in (a), (c) and (d). The possible reason is that it only considers the patch of max malignant probability in training and the model is not well learned for all patches. The label assignment-based deep MIL mis-classifies some patches in (d). The possible reason is that the model sets a constant $k$ for all the mammograms, which causes some mis-classifications for small masses. One of the potential applications of our work is that these deep MIL networks could be used to do weak mass annotation automatically, which provides evidence for the diagnosis.
\section{Conclusion}\label{2sec:con}
In this paper, we propose end-to-end trained deep MIL for whole mammogram classification. Different from previous work using segmentation or detection annotations, we conduct mass classification based on whole mammogram directly. We convert the general MIL assumption to label assignment problem after ranking. Due to the sparsity of masses, sparse MIL is used for whole mammogram classification. Experimental results demonstrate more robust performance than previous work even without detection or segmentation annotation in the training.

In future work, we plan to extend the current work by: 1) incorporating multi-scale modeling such as spatial pyramid to further improve whole mammogram classification, 2) employing the deep MIL to do annotation or provide potential malignant patches to assist diagnoses, and 3) applying to large datasets and expected to have improvement if the big dataset is available.


\chapter{DeepLung: Deep 3D Dual Path Nets for Automated Pulmonary Nodule Detection and Classification}

\begin{figure*}
	\begin{center}
		\includegraphics[width=\linewidth]{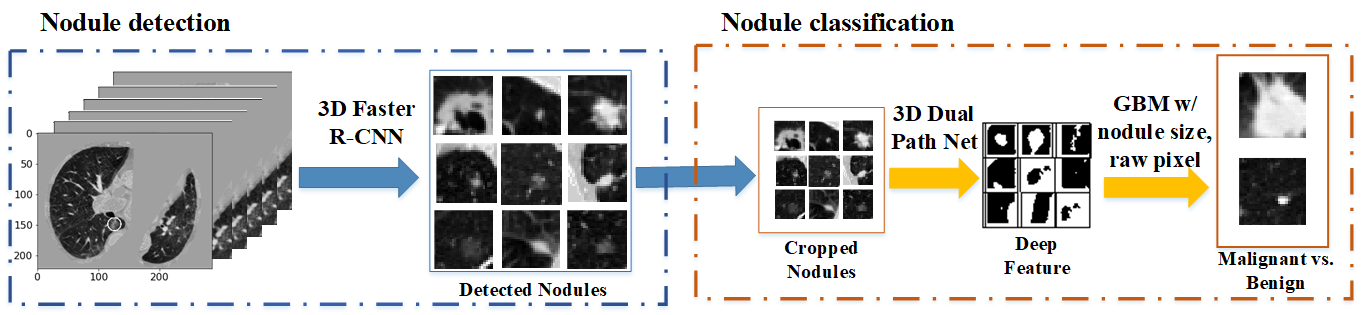}
		\caption{The framework of DeepLung. DeepLung first employs 3D Faster R-CNN to generate candidate nodules. Then it uses deep 3D DPN to extract deep features from the detected and cropped nodules. Lastly, GBM with deep features, detected nodule size, and raw pixels is employed for classification. Patient-level diagnosis can be achieved by fusing the classification results of detected nodules in the CT.}\label{3fig:framework}
	\end{center}
\end{figure*}
\section{Introduction}

Lung cancer is the most common cause of cancer-related death in men. Low-dose lung CT screening provides an effective way for early diagnosis, which can sharply reduce the lung cancer mortality rate. Advanced computer-aided diagnosis systems (CADs) are expected to have high sensitivities while at the same time maintaining low false positive rates. Recent advances in deep learning enable us to rethink the ways of clinician lung cancer diagnosis.

Current lung CT analysis research mainly includes nodule detection \cite{dou2017automated,ding2017accurate}, and nodule classification \cite{shen2015multi,dlcls,hussein2017risk,yan2016classification}. There has been little work previously on building a complete lung CT cancer diagnosis system for fully automated lung CT cancer diagnosis using deep learning, integrating both nodule detection and nodule classification. It is worth exploring a whole lung CT cancer diagnosis system and understanding how far the performance of current deep learning technology differs from that of experienced doctors. To our best knowledge, this is the first work for a fully automated and complete lung CT cancer diagnosis system using deep nets. 

The emergence of large-scale dataset, LUNA16 \cite{setio2016pulmonary}, accelerates the nodule detection related research. Typically, nodule detection consists of two stages, region proposal generation and false positive reduction. Traditional approaches generally require manually designed features such as morphological features, voxel clustering and pixel thresholding \cite{murphy2009large,jacobs2014automatic}. Recently, deep ConvNets, such as Faster R-CNN \cite{kaimingfasterrcnn,liao2017evaluate} and fully ConvNets \cite{long2015fully,zhu2017adversarial,zhe_vqa,zhe_scene,zhe_action}, are employed to generate candidate bounding boxes \cite{ding2017accurate,dou2017automated}. In the second stage, more advanced methods or complex features, such as carefully designed texture features, are used to remove false positive nodules. Because of the 3D nature of CT data and the effectiveness of Faster R-CNN for object detection in 2D natural images \cite{iccv17detectorcompare}, we design a 3D Faster R-CNN for nodule detection with 3D convolutional kernels and a U-net-like encoder-decoder structure to effectively learn latent features \cite{ronneberger2015u}. The U-Net structure is basically a convolutional autoencoder, augmented with skip connections between encoder and decoder layers \cite{ronneberger2015u}. Although it has been widely used in the context of semantic segmentation, being able to capture both contextual and local information should be very helpful for nodule detections as well. Because 3D ConvNet has too many parameters and is hard to train on public lung CT datasets of relatively small sizes, 3D dual path network is employed as the building block since deep dual path network is more compact and provides better performance than deep residual network at the same time \cite{chen2017dual}. 

Before the era of deep learning, feature engineering followed by classifiers is a general pipeline for nodule classification \cite{han2013texture}. After the public large-scale dataset, LIDC-IDRI \cite{armato2011lung}, becomes available, deep learning based methods have become dominant for nodule classification research \cite{dlcls,zhu2016co}. Multi-scale deep ConvNet with shared weights on different scales has been proposed for the nodule classification \cite{shen2015multi}. The weight sharing scheme reduces the number of parameters and forces the multi-scale deep ConvNet to learn scale-invariant features. Inspired by the recent success of dual path network (DPN) on ImageNet \cite{chen2017dual,imagenet}, we propose a novel framework for CT nodule classification. First, we design a deep 3D dual path network to extract features. Considering the excellent power of gradient boosting machines (GBM) given effective features, we use GBM with deep 3D dual path features, nodule size and cropped raw nodule CT pixels for the nodule classification \cite{gbt}. 


Finally, we build a fully automated lung CT cancer diagnosis system, DeepLung, by combining the nodule detection network and nodule classification network together, as illustrated in Fig. \ref{3fig:framework}. For a CT image, we first use the detection subnetwork to detect candidate nodules. Next, we employ the classification subnetwork to classify the detected nodules into either malignant or benign. Finally, the patient-level diagnosis result can be achieved for the whole CT by fusing the diagnosis result of each nodule.

Our main contributions are as follows: 1) To fully exploit the 3D CT images, two deep 3D ConvNets are designed for nodule detection and classification respectively. Because 3D ConvNet contains too many parameters and is hard to train on relatively small public lung CT datasets, we employ 3D dual path networks as the components since DPN uses less parameters and obtains better performance than residual network \cite{chen2017dual}. Specifically, inspired by the effectiveness of Faster R-CNN for object detection \cite{iccv17detectorcompare}, we propose 3D Faster R-CNN for nodule detection based on 3D dual path network and U-net-like encoder-decoder structure, and deep 3D dual path network for nodule classification. 2) Our classification framework achieves better performance compared with state-of-the-art approaches, and the performance surpasses the performance of experienced doctors on the largest public dataset, LIDC-IDRI dataset. 3) The fully automated DeepLung system, nodule classification based on detection, is comparable to the performance of experienced doctors both on nodule-level and patient-level diagnosis. 

\section{Related Work}

Traditional nodule detection requires manually designed features or descriptors \cite{lopez2015large}. Recently, several works have been proposed to use deep ConvNets for nodule detection to automatically learn features, which is proven to be much more effective than hand-crafted features. Setio et al. proposes multi-view ConvNet for false positive nodule reduction \cite{setio2016pulmonarymultiview}. Due to the 3D nature of CT scans, some work propose 3D ConvNets to handle the challenge. The 3D fully ConvNet (FCN) is proposed to generate region candidates, and deep ConvNet with weighted sampling is used in the false positive candidates reduction stage \cite{dou2017automated}. Ding et al. and Liao et al. use the Faster R-CNN to generate candidate nodules, followed by 3D ConvNets to remove false positive nodules \cite{ding2017accurate,liao2017evaluate}. Due to the effective performance of Faster R-CNN \cite{iccv17detectorcompare,kaimingfasterrcnn}, we design a novel network, 3D Faster R-CNN with 3D dual path blocks, for the nodule detection. Further, a U-net-like encoder-decoder scheme is employed for 3D Faster R-CNN to effectively learn the features \cite{ronneberger2015u}.

Nodule classification has traditionally been based on segmentation \cite{el20113d} and manual feature design \cite{aerts2014decoding}. Several works designed 3D contour feature, shape feature and texture feature for CT nodule diagnosis \cite{way2006computer,el20113d,han2013texture}. Recently, deep networks have been shown to be effective for medical images. Artificial neural network was implemented for CT nodule diagnosis \cite{suzuki2005computer}. More computationally effective network, multi-scale ConvNet with shared weights for different scales to learn scale-invariant features, is proposed for nodule classification \cite{shen2015multi}. Deep transfer learning and multi-instance learning is used for patient-level lung CT diagnosis \cite{dlcls,zhu2017deep}. A comparison on 2D and 3D ConvNets is conducted and shown that 3D ConvNet is better than 2D ConvNet for 3D CT data \cite{yan2016classification}. Further, a multi-task learning and transfer learning framework is proposed for nodule diagnosis \cite{hussein2017risk}. Different from their approaches, we propose a novel classification framework for CT nodule diagnosis. Inspired by the recent success of deep dual path network (DPN) on ImageNet \cite{chen2017dual}, we design a novel totally 3D DPN to extract features from raw CT nodules. Due to the superior power of gradient boost machine (GBM) with complete features, we employ GBM with different levels of granularity ranging from raw pixels, DPN features, to global features such as nodule size for the nodule diagnosis. Patient-level diagnosis can be achieved by fusing the nodule-level diagnosis. 

\section{DeepLung Framework}
The fully automated lung CT cancer diagnosis system, DeepLung, consists of two parts, nodule detection and classification. We design a 3D Faster R-CNN for nodule detection, and propose GBM with deep 3D DPN features, raw nodule CT pixels and nodule size for nodule classification.

\subsection{3D Faster R-CNN with Deep 3D Dual Path Net for Nodule Detection}
\begin{figure}
	\includegraphics[width=\linewidth]{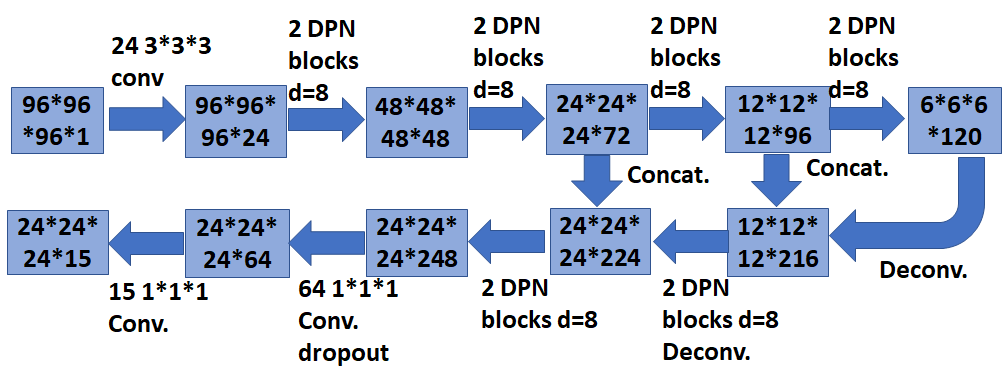}
	\caption{The 3D Faster R-CNN framework contains 3D dual path blocks and a U-net-like encoder-decoder structure. We design 26 layers 3D dual path network for the encoder subnetwork. The model employs 3 anchors and multi-task learning loss, including coordinates \((x, y, z)\) and diameter \(d\) regression, and candidate box classification. The numbers in boxes are feature map sizes in the format (\(\#\)slices*\(\#\)rows*\(\#\)cols*\(\#\)maps). The numbers above the connections are in the format (\(\#\)filters \quad \(\#\)slices*\(\#\)rows*\(\#\)cols).}
	\label{3fig:3dfasterrcnn}
\end{figure}

Inspired by the success of dual path network on the ImageNet \cite{chen2017dual,imagenet}, we design a deep 3D DPN framework for lung CT nodule detection and classification in Fig. \ref{3fig:3dfasterrcnn} and Fig. \ref{3fig:dpncls}. 
\begin{figure}
	\begin{center}
		\includegraphics[width=\linewidth]{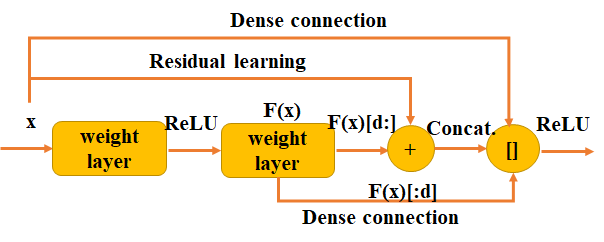}
		\caption{Illustration of dual path connection \cite{chen2017dual}, which benefits both from the advantage of residual learning \cite{he2016deep} and that of dense connection \cite{huang2016densely}.}
		\label{3fig:dualpath}
	\end{center}
\end{figure}
Dual path connection benefits both from the advantage of residual learning and that of dense connection \cite{he2016deep,huang2016densely}. The shortcut connection in residual learning is an effective way to eliminate gradient vanishing phenomenon in very deep networks. From a learned feature sharing perspective, residual learning enables feature reuse while dense connection allows the network to continue to exploit new features \cite{chen2017dual}. The densely connected network has fewer parameters than residual learning because there is no need to relearn redundant feature maps. The assumption of dual path connection is that there might exist some redundancy in the exploited features. And dual path connection uses part of feature maps for dense connection and part of them for residual learning. In implementation, the dual path connection splits its feature maps \({\bf{F}(\bf{x})}\) into two parts. Here we use Python's vector notation where \({\bf{F}(\bf{x})[:k]}\) means we subset the first $k$ channel of \({\bf{F}(\bf{x})}\), and \({\bf{F}(\bf{x})[k:]}\) means the \(k+1\) to last channel of \({\bf{F}(\bf{x})}\). The first \(d\) channels, \({\bf{F}(\bf{x})}[:d]\), are used for dense connection, and other channels, \({\bf{F}(\bf{x})}[d:]\), are used for residual learning as shown in Fig. \ref{3fig:dualpath}. Here \(d\) is a hyper-parameter for deciding how many new features to be exploited. The dual path connection can be formulated as
\begin{equation}
{\bf{y}} = {\bf{G}}([{\bf{x}}[:d], {\bf{F}(\bf{x})}[:d], {\bf{F}(\bf{x})}[d:]+{\bf{x}}[d:]),
\end{equation}
where \(\bf{y}\) is the feature map for dual path connection, \(\bf{G}\) is used as ReLU activation function, \(\bf{F}\) is convolutional layer functions, and \(\bf{x}\) is the input of dual path connection block. Dual path connection integrates the advantages of the two advanced frameworks, residual learning for feature reuse and dense connection for keeping exploiting new features, into a unified structure, which obtains success on the ImageNet dataset\cite{imagenet}. We design deep 3D neural nets based on 3D DPN because of its compactness and effectiveness.

The 3D Faster R-CNN with a U-net-like encoder-decoder structure and 3D dual path blocks is illustrated in Fig. \ref{3fig:3dfasterrcnn}. Due to the GPU memory limitation, the input of 3D Faster R-CNN is cropped from 3D reconstructed CT images with pixel size \(96 \times 96 \times 96\). The encoder network is derived from 2D DPN \cite{chen2017dual}. Before the first max-pooling, two convolutional layers are used to generate features. After that, eight dual path blocks are employed in the encoder subnetwork. We integrate the U-net-like encoder-decoder design concept in the detection to learn the deep nets efficiently \cite{ronneberger2015u}. In fact, for the region proposal generation, the 3D Faster R-CNN conducts pixel-wise multi-scale learning and the U-net is validated as an effective way for pixel-wise labeling. This integration makes candidate nodule generation more effective. In the decoder network, the feature maps are processed by deconvolution layers and dual path blocks, and are subsequently concatenated with the corresponding layers in the encoder network \cite{zeiler2010deconvolutional}. Then a convolutional layer with dropout (dropout probability 0.5) is used for the second last layer. In the last layer, we design 3 anchors, 5, 10, 20, for scale references which are designed based on the distribution of nodule sizes. For each anchor, there are 5 parts in the loss function, classification loss \(L_{cls}\) for whether the current box is a nodule or not, regression loss \(L_{reg}\) for nodule coordinates \(x, y, z\) and nodule size \(d\).

If an anchor overlaps a ground truth bounding box with the intersection over union (IoU) higher than 0.5, we consider it as a positive anchor (\(p^{\star}=1\)). On the other hand, if an anchor has IoU with all ground truth boxes less than 0.02, we consider it as a negative anchor (\(p^{\star}=0\)). The multi-task loss function for the anchor \(i\) is defined as
\begin{equation}
L(p_i, {\bf{t}_i}) = \lambda L_{cls}(p_i, p_i^{\star}) + p_i^{\star} L_{reg}({\bf{t}_i}, {\bf{t}_i}^{\star}),
\end{equation}
where \(p_i\) is the predicted probability for current anchor \(i\) being a nodule, \(\bf{t}_i\) is the predicted relative coordinates for nodule position, which is defined as
\begin{equation}
{\bf{t}_i} = (\frac{x-x_a}{d_a}, \frac{y-y_a}{d_a}, \frac{z-z_a}{d_a}, \log(\frac{d}{d_a})),
\end{equation}
where \((x, y, z, d)\) are the predicted nodule coordinates and diameter in the original space, \((x_a, y_a, z_a, d_a)\) are the coordinates and scale for the anchor \(i\). For ground truth nodule position, it is defined as
\begin{equation}
{{\bf{t}}_i^{\star}} = (\frac{x^{\star}-x_a}{d_a}, \frac{y^{\star}-y_a}{d_a}, \frac{z^{\star}-z_a}{d_a}, \log(\frac{d^{\star}}{d_a})),
\end{equation}
where \((x^{\star}, y^{\star}, z^{\star}, d^{\star})\) are nodule ground truth coordinates and diameter. \(\lambda\) is set as \(0.5\). For \(L_{cls}\), we used binary cross entropy loss function. For \(L_{reg}\), we used smooth \(l_1\) regression loss function \cite{girshick2015fast}. 
\subsection{Gradient Boosting Machine with 3D Dual Path Net Feature for Nodule Classification}

\begin{figure}
	\begin{center}
		\includegraphics[width=0.9\linewidth]{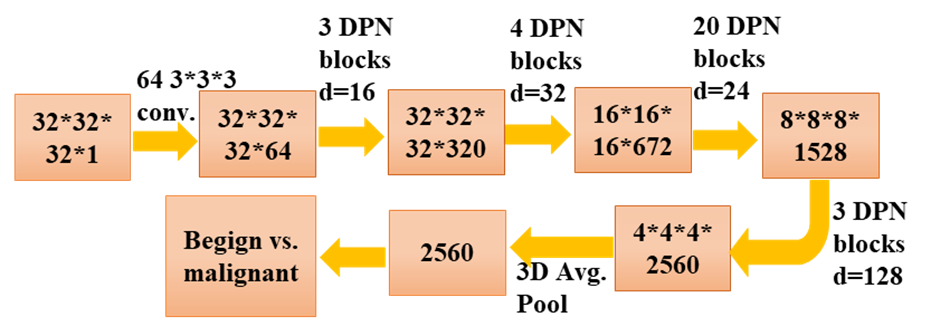}
		\caption{The deep 3D dual path network framework in the nodule classification subnetwork, which contains 30 3D dual path connection blocks. After the training, the deep 3D dual path network feature is extracted for gradient boosting machine to do nodule diagnosis. The numbers are of the same formats as Fig. \ref{3fig:3dfasterrcnn}.}
		\label{3fig:dpncls}
	\end{center}
\end{figure}

For CT data, advanced method should be effective to extract 3D volume feature \cite{yan2016classification}. We design a 3D deep dual path network for the 3D CT lung nodule classification in Fig. \ref{3fig:dpncls}. The main reason we employ dual modules for detection and classification is that classifying nodules into benign and malignant requires the system to learn finer-level features, which can be achieved by focusing only on nodules. Additionally, it introduces extra features in the final classification. We firstly crop CT data centered at predicted nodule locations with size \(32\times 32 \times 32\). After that, a convolutional layer is used to extract features. Then 30 3D dual path blocks are employed to learn higher level features. Lastly, the 3D average pooling and binary logistic regression layer are used for benign or malignant diagnosis. 

The deep 3D dual path network can be used as a classifier for nodule diagnosis directly. And it can also be employed to learn effective features. We construct features by concatenating the learned deep 3D DPN features (the second last layer, 2,560 dimension), nodule size, and raw 3D cropped nodule pixels. Given complete and effective features, GBM learns a sequence of tree classifiers for residual errors and is an effective method to build an advanced classifier from these features \cite{gbt}. We validate the feature combining nodule size with raw 3D cropped nodule pixels, employ GBM as a classifier, and obtain \(86.12 \%\) test accuracy averagely. Lastly, we use the previously constructed features with the GBM classifier to achieve the best diagnosing performance. 

\subsection{DeepLung System: Fully Automated Lung CT Cancer Diagnosis}
The DeepLung system includes the nodule detection using the 3D Faster R-CNN, and nodule classification using GBM with constructed feature (deep 3D dual path features, nodule size and raw nodule CT pixels) in Fig. \ref{3fig:framework}. 

Due to the GPU memory limitation, we first split the whole CT into several \(96 \times 96 \times 96\) patches, process them through the detector, and combine the detected results together. We only keep the detected boxes of detection probabilities larger than 0.12 (threshold as -2 before sigmoid function). After that, non-maximum suppression (NMS) is adopted based on detection probability with the intersection over union (IoU) threshold as 0.1. Here we expect to not miss too many ground truth nodules. 

After we get the detected nodules, we crop the nodule with the center as the detected center and size as \(32 \times 32 \times 32\). The detected nodule size is kept for the classification model as a part of features. The deep 3D DPN is employed to extract features. We use the GBM and construct features to conduct diagnosis for the detected nodules. For pixel feature, we use the cropped size as \(16 \times 16 \times 16\) and center as the detected nodule center in the experiments. For patient-level diagnosis, if one of the detected nodules is positive (cancer), the patient is a cancer patient, and if all the detected nodules are negative, the patient is a negative patient.

\section{Experiments}
We conduct extensive experiments to validate the DeepLung system. We perform 10-fold cross validation using the detector on LUNA16 dataset. For nodule classification, we use the LIDC-IDRI annotation, and employ the LUNA16's patient-level dataset split. Finally, we also validate the whole system based on the detected nodules both on patient-level diagnosis and nodule-level diagnosis. 

In the training, for each model, we use 150 epochs in total with stochastic gradient descent optimization and momentum as 0.9. The used batch size is set based on the GPU memory. We use weight decay as \(1 \times 10^{-4}\). The initial learning rate is 0.01, 0.001 at the half of training, and 0.0001 after the epoch 120.

\subsection{Datasets}

LUNA16 dataset is a subset of the largest public dataset for pulmonary nodules, the LIDC-IDRI dataset \cite{armato2011lung,setio2016pulmonary}. LUNA16 dataset only has the detection annotations, while LIDC-IDRI contains almost all the related information for low-dose lung CTs including several doctors' annotations on nodule sizes, locations, diagnosis results, nodule texture, nodule margin and other informations. LUNA16 dataset removes CTs with slice thickness greater than 3mm, slice spacing inconsistent or missing slices from LIDC-IDRI dataset, and explicitly gives the patient-level 10-fold cross validation split of the dataset. LUNA16 dataset contains 888 low-dose lung CTs, and LIDC-IDRI contains 1,018 low-dose lung CTs. Note that LUNA16 dataset removes the annotated nodules of size smaller than 3mm. 

For nodule classification, we extract nodule annotations from LIDC-IDRI dataset, find the mapping of different doctors' nodule annotations with the LUNA16's nodule annotations, and get the ground truth of nodule diagnosis by taking different doctors' diagnosis equally (Do not count the 0 score for diagnosis, which means N/A.). If the final average score is equal to 3 (uncertain about malignant or benign), we remove the nodule. For the nodules with score greater than 3, we label them as positive. Otherwise, we label them as negative. Because CT slides were annotated by anonymous doctors, the identities of doctors (referred to as Drs 1-4 as the 1st-4th annotations) are not strictly consistent. As such, we refer them as ``simulated'' doctors. To make our results reproducible, we only keep the CTs within LUNA16 dataset, and use the same cross validation split as LUNA16 for classification. 
\subsection{Preprocessing}
Three automated preprocessing steps are employed for the input CT images. We firstly clip the raw data into \([-1200, 600]\). Secondly, we transform the range linearly into \([0, 1]\). Thirdly, we use the LUNA16's given segmentation ground truth and remove the useless background. 
\subsection{DeepLung for Nodule Detection}
We train and evaluate the detector on LUNA16 dataset following 10-fold cross validation with given patient-level split. In the training, we use flipping, randomly scale from 0.75 to 1.25 for the cropped patches to augment the data. The evaluation metric, FROC, is the average recall rate at the average number of false positives as 0.125, 0.25, 0.5, 1, 2, 4, 8 per scan, which is the official evaluation metric for LUNA16 dataset \cite{setio2016pulmonary}. In the test phase, we use detection probability threshold as -2 (before sigmoid function), followed by NMS with IoU threshold as 0.1. 

\begin{figure}
	\includegraphics[width=\linewidth]{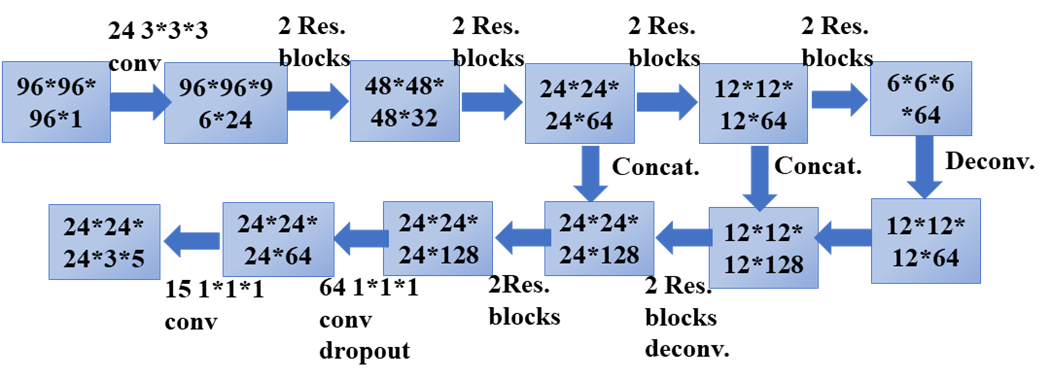}
	\caption{The 3D Faster R-CNN network with 3D residual blocks. It contains several 3D residual blocks. We employ a deep 3D residual network of 18 layers as the encoder subnetwork, which is an extension from 2D Res18 net \cite{he2016deep}.}
	\label{3fig:3dfasterrcnncmp}
\end{figure}

To validate the superior performance of proposed deep 3D dual path network for detection, we employ a deep 3D residual network as a comparison in Fig. \ref{3fig:3dfasterrcnncmp}. The encoder part of compared network is a deep 3D residual network of 18 layers, which is an extension from 2D Res18 net \cite{he2016deep}. Note that the 3D Res18 Faster R-CNN contains \(5.4\)M trainable parameters, while the 3D DPN26 Faster R-CNN employs \(1.4\)M trainable parameters, which is only \( \frac{\textbf{1}}{\textbf{4}}\) of that in 3D Res18 Faster R-CNN. 
\begin{figure}
	\begin{center}
		\includegraphics[width=0.9\linewidth]{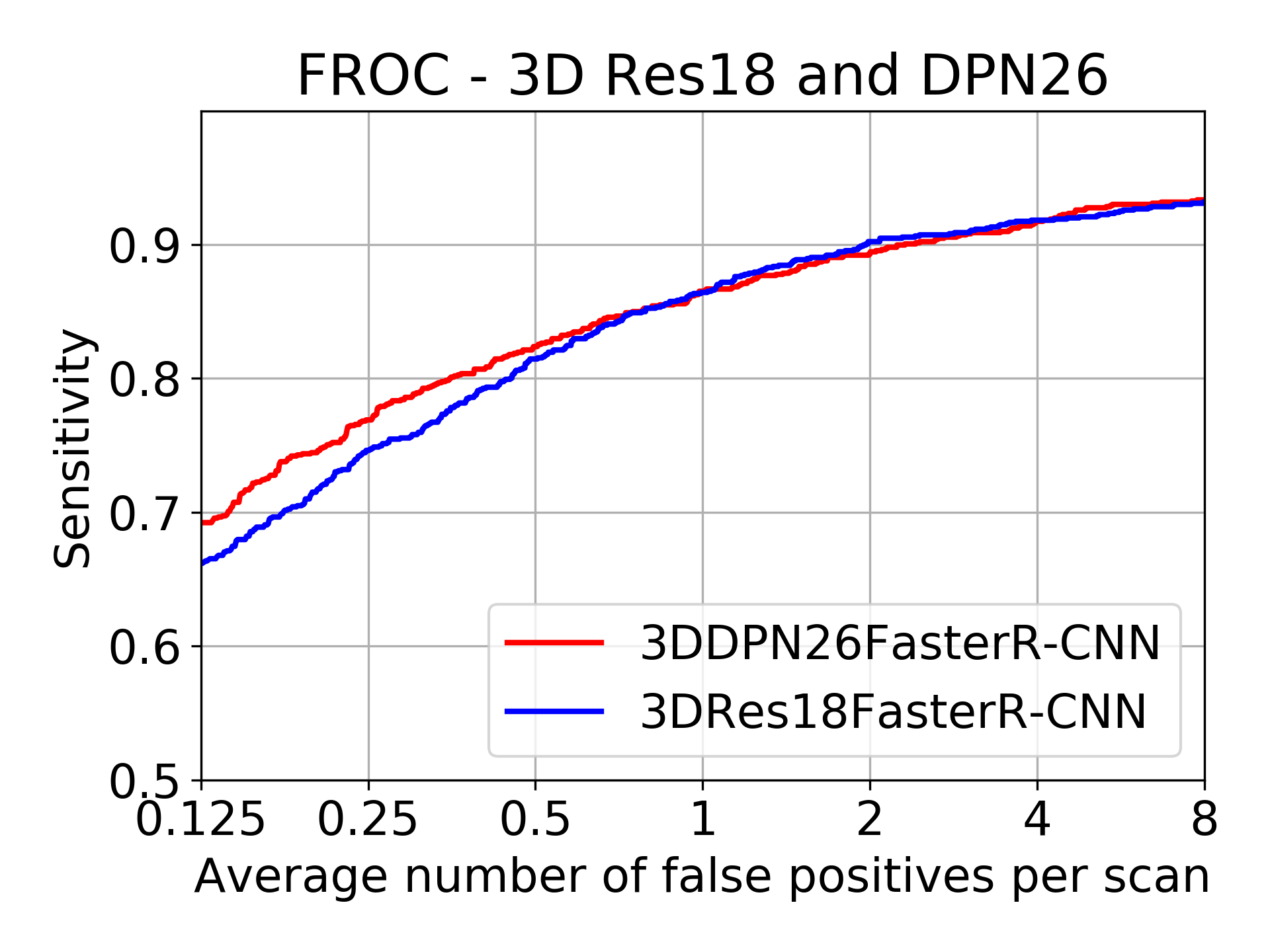}
		\caption{Sensitivity (Recall) rate with respect to false positives per scan. The FROC (average recall rate at the false positives as 0.125, 0.25, 0.5, 1, 2, 4, 8) of 3D Res18 Faster R-CNN is 83.4\%, while the FROC of 3D DPN26 Faster R-CNN is \textbf{84.2\%} with only \( \frac{\textbf{1}}{\textbf{4}}\) of the parameters as 3D Res18 Faster R-CNN. The 3D Res18 Faster R-CNN has a total recall rate 94.6\% for all the detected nodules, while 3D DPN26 Faster R-CNN has a recall rate \textbf{95.8\%}.}
		\label{3fig:froc}
	\end{center}
\end{figure}

The FROC performance on LUNA16 is visualized in Fig. \ref{3fig:froc}. The solid line is interpolated FROC based on true prediction. The 3D DPN26 Faster R-CNN achieves {\textbf{84.2\%}} FROC without any false positive nodule reduction stage, which is better than 83.9\% using two-stage training \cite{dou2017automated}. The 3D DPN26 Faster R-CNN using only \( \frac{\textbf{1}}{\textbf{4}}\) of the parameters preforms better than the 3D Res18 Faster R-CNN, which demonstrates the superior suitability of the 3D DPN for detection. Ding et al. obtains 89.1\% FROC using 2D Faster R-CNN followed by extra false positive reduction classifier \cite{ding2017accurate}, while we only employ enhanced Faster R-CNN with deep 3D dual path for detection. We have recently applied the 3D model to Alibaba Tianchi Medical AI on nodule detection challenge and were able to achieve top accuracy on a hold-out dataset. 

\subsection{DeepLung for Nodule Classification}

We validate the nodule classification performance of the DeepLung system on the LIDC-IDRI dataset with the LUNA16's split principle, 10-fold patient-level cross validation. There are 1,004 nodules left and 450 nodules are positive. In the training, we firstly pad the nodules of size \(32 \times 32 \times 32\) into \(36 \times 36 \times 36\), randomly crop \(32 \times 32 \times 32\) from the padded data, horizontal flip, vertical flip, z-axis flip the data for augmentation, randomly set \(4\times4\times4\) patch as 0, and normalize the data with the mean and standard deviation obtained from training data. The total number of epochs is 1,050. The learning rate is 0.01 at first, then became 0.001 after epoch 525, and turned into 0.0001 after epoch 840. Due to time and resource limitation for training, we use the fold 1, 2, 3, 4, 5 for test, and the final performance is the average performance on the five test folds. The nodule classification performance is concluded in Table \ref{3tab:nodcls}.
\begin{table}[]
	\centering
	\caption{Nodule classification comparisons on LIDC-IDRI dataset.}
	\label{3tab:nodcls}
	\begin{tabular}{c|c|c}
		\hlinew{0.9pt}
		Models & Accuracy (\%) & Year \\  
		\hlinew{0.7pt}
		Multi-scale CNN \cite{shen2015multi} & 86.84 & 2015 \\ \hline
		Slice-level 2D CNN \cite{yan2016classification} & 86.70 & 2016 \\ \hline
		Nodule-level 2D CNN \cite{yan2016classification} & 87.30 & 2016 \\ \hline
		Vanilla 3D CNN \cite{yan2016classification} & 87.40 & 2016 \\ \hline
		Multi-crop CNN \cite{shen2017multi} & 87.14 & 2017 \\ 
		\hlinew{0.9pt}
		Deep 3D DPN & 88.74 & 2017 \\ \hline
		Nodule Size+Pixel+GBM & 86.12 & 2017 \\ \hline
		All feat.+GBM & \textbf{90.44} & 2017 \\ 
		\hlinew{0.9pt}
	\end{tabular}
\end{table}

From the table \ref{3tab:nodcls}, our deep 3D dual path network (DPN) achieves better performance than those of Multi-scale CNN \cite{shen2015multi}, Vanilla 3D CNN \cite{yan2016classification} and Multi-crop CNN \cite{shen2017multi}, because of the strong power of 3D structure and deep dual path network. GBM with nodule size and raw nodule pixels with crop size as \(16 \times 16 \times 16\) achieves comparable performance as multi-scale CNN \cite{shen2015multi} because of the superior classification performance of gradient boosting machine (GBM). Finally, we construct feature with deep 3D dual path network features, 3D Faster R-CNN detected nodule size and raw nodule pixels, and obtain {\bf{90.44\%}} accuracy, which shows the effectiveness of deep 3D dual path network features. 
\subsubsection{Compared with Experienced Doctors on Their Individually Confident Nodules}

We compare our predictions with those of four ``simulated'' experienced doctors on their individually confident nodules (with individual score not 3). Note that about 1/3 annotations are 3. Comparison results are concluded in Table \ref{3tab:nodclsdr}.

\begin{table}[]
	\centering
	\caption{Nodule-level diagnosis accuracy (\%) between nodule classification subnetwork in DeepLung and experienced doctors on doctor's individually confident nodules.}
	\label{3tab:nodclsdr}
	\begin{tabular}{c|c|c|c|c|c}
		\hline
		& Dr 1 & Dr 2 & Dr 3 & Dr 4 & Average \\ \hline
		Doctors   & 93.44    & 93.69    & 91.82    & 86.03    & 91.25   \\ \hline
		DeepLung & 93.55    & 93.30    & 93.19    & 90.89    & \textbf{92.74}   \\ \hline
	\end{tabular}
\end{table}

From Table \ref{3tab:nodclsdr}, these doctors' confident nodules are easy to be diagnosed nodules from the performance comparison between our model's performances in Table \ref{3tab:nodcls} and Table \ref{3tab:nodclsdr}. To our surprise, the average performance of our model is {\textbf{1.5\%}} better than that of experienced doctors even on their individually confident diagnosed nodules. In fact, our model's performance is better than 3 out of 4 doctors (doctor 1, 3, 4) on the confident nodule diagnosis task. The result validates deep network surpasses human-level consistency for image classification \cite{he2016deep}, and the DeepLung is better suited for nodule diagnosis than experienced doctors. 

\begin{table}[]
	\centering
	\caption{Statistical property of predicted malignant probability for borderline nodules (\%)}
	\label{3tab:statborderlinenod}
	\begin{tabular}{c|c|c|c|c}
		\hline
		Prediction & \begin{tabular}[c]{@{}l@{}}\(<0.1\) or\\ \(>0.9\)\end{tabular}   &\begin{tabular}[c]{@{}l@{}}\(<0.2\) or\\ \(>0.8\)\end{tabular} & \begin{tabular}[c]{@{}l@{}}\(<0.3\) or\\ \(>0.7\)\end{tabular}  & \begin{tabular}[c]{@{}l@{}}\(<0.4\) or\\ \(>0.6\)\end{tabular}  \\ \hline
		Frequency & 64.98    & 80.14    & 89.75    & 94.80   \\ \hline
	\end{tabular}
\end{table}
We also employ Kappa coefficient, which is a common approach to evaluate the agreement between two raters, to test the agreement between DeepLung and the ground truth \cite{smeeton1985early}. The kappa coefficient of DeepLung is 85.07\%, which is significantly better than the average kappa coefficient of doctors (81.58\%). To evaluate the performance for all nodules including borderline nodules (labeled as 3, uncertain between malignant and benign), we compute the log likelihood (LL) scores of DeepLung and doctors' diagnosis. We randomly sample 100 times from the experienced doctors' annotations as 100 ``simulated'' doctors. The mean LL of doctors is -2.563 with a standard deviation of 0.23. By contrast, the LL of DeepLung is -1.515, showing that the performance of DeepLung is 4.48 standard deviation better than the average performance of doctors, which is highly statistically significant. It is important to analysis the statistical property of predictions for borderline nodules that cannot be conclusively classified by doctors. Interestingly, 64.98\% of the borderline nodules are classified to be either malignant (with probability $>$ 0.9) or benign (with probability $<$ 0.1) in Table \ref{3tab:statborderlinenod}.  DeepLung classified most of the borderline nodules of malignant probabilities closer to zero or closer to one. A system that produces the uncertainty estimation of prediction is desired as a tool for assisted diagnosis and we expect such a work to be done in the future. 

\subsection{DeepLung for Fully Automated Lung CT Cancer Diagnosis}
We also validate the DeepLung for fully automated lung CT cancer diagnosis on the LIDC-IDRI dataset with the same protocol as LUNA16's patient-level split. Firstly, we employ our 3D Faster R-CNN to detect suspicious nodules. Then we retrain the model from nodule classification model on the detected nodules dataset. If the center of detected nodule is within the ground truth positive nodule, it is a positive nodule. Otherwise, it is a negative nodule. Through this mapping from the detected nodule and ground truth nodule, we can evaluate the performance and compare it with the performance of experienced doctors. We adopt the test fold 1, 2, 3, 4, 5 to validate the performance the same as that for nodule classification.

\begin{table}[]
	\centering
	\caption{Comparison between DeepLung's nodule classification on all detected nodules and doctors on all nodules.}
	\label{3tab:nodclstpfp}
	\begin{tabular}{c|c|c|c}
		\hline
		Method& TP Set & FP Set & Doctors  \\ \hline
		Acc. (\%)   & 81.42    & 97.02    & 74.05-82.67       \\ \hline
	\end{tabular}
\end{table}

Different from pure nodule classification, the fully automated lung CT nodule diagnosis relies on nodule detection. We evaluate the performance of DeepLung on the detection true positive (TP) set and detection false positive (FP) set individually in Table \ref{3tab:nodclstpfp}. If the detected nodule of center within one of ground truth nodule regions, it is in the TP set. If the detected nodule of center out of any ground truth nodule regions, it is in FP set. From Table \ref{3tab:nodclstpfp}, the DeepLung system using detected nodule region obtains \(\textbf{81.42\%}\) accuracy for all the detected TP nodules. Note that the experienced doctors obtain 78.36\% accuracy for all the nodule diagnosis on average. The DeepLung system with fully automated lung CT nodule diagnosis still achieves above average performance of experienced doctors. On the FP set, our nodule classification subnetwork in the DeepLung can reduce 97.02\% FP detected nodules, which guarantees that our fully automated system is effective for the lung CT cancer diagnosis. 

\subsubsection{Compared with Experienced Doctors on Their Individually Confident CTs}

We employ the DeepLung for patient-level diagnosis further. If the current CT has one nodule that is classified as positive, the diagnosis of the CT is positive. If all the nodules are classified as negative for the CT, the diagnosis of the CT is negative. We evaluate the DeepLung on the doctors' individually confident CTs for benchmark comparison in Table \ref{3tab:ctclsdr}. 
\begin{table}[]
	\centering
	\caption{Patient-level diagnosis accuracy(\%) between DeepLung and experienced doctors on doctor's individually confident CTs.}
	\label{3tab:ctclsdr}
	\begin{tabular}{c|c|c|c|c|c}
		\hline
		& Dr 1 & Dr 2 & Dr 3 & Dr 4 & Average \\ \hline
		Doctors   & 83.03    & 85.65    & 82.75    & 77.80    & \textbf{82.31}   \\ \hline
		DeepLung & 81.82    & 80.69    & 78.86    & 84.28    & 81.41   \\ \hline
	\end{tabular}
\end{table}

From Table \ref{3tab:ctclsdr}, DeepLung achieves {\bf{81.41\%}} patient-level diagnosis accuracy. The performance is {\bf{99\%}} of the average performance of four experienced doctors, and the performance of DeepLung is better than that of doctor 4. Thus DeepLung can be used to help improve some doctors' performance, like that of doctor 4, which is the goal for computer aided diagnosis system. For comparison, we calculate the Kappa coefficient of four individual doctors on their individual confident CTs. The Kappa coefficient of DeepLung is 63.02\%, while the average Kappa coefficient of doctors is 64.46\%. It shows the predictions of DeepLung are in good agreement with human diagnosis for patient-level diagnosis, and are comparable with those of experienced doctors.
\section{Discussion}
In this section, we are trying to explain the DeepLung by visualizing the nodule detection and classification results.
\subsection{Nodule Detection}
\begin{figure*}[h]
	\begin{center}
		\includegraphics[width=\linewidth]{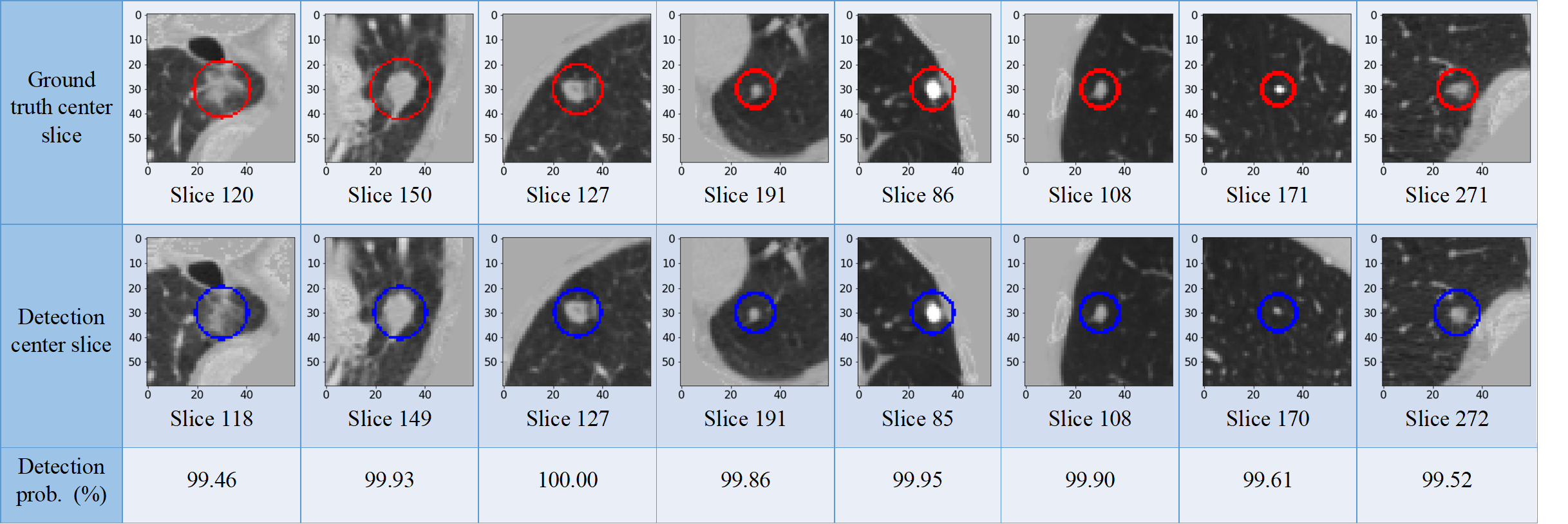}
		\caption{Visualization of central slices for nodule ground truth and detection results. We randomly choose nodules (red circle boxes in the first row) from test fold 1. Detection results are shown in the blue circles of second row. The center slice numbers are shown below the images. The last row shows detection probability. The DeepLung performs well for nodule detection.}
		\label{3fig:detection}
	\end{center}
\end{figure*}

We randomly pick nodules from test fold 1 and visualize them in red circles of the first row in Fig. \ref{3fig:detection}. Detected nodules are visualized in blue circles of the second row. Because CT is 3D voxel data, we can only plot the central slice for visualization. The third row shows the detection probabilities for the detected nodules. The central slice number is shown below each slice. The diameter of the circle is relative to the nodule size.

From the central slice visualizations in Fig. \ref{3fig:detection}, we observe the detected nodule positions including central slice numbers are consistent with those of ground truth nodules. The circle sizes are similar between the nodules in the first row and the second row. The detection probability is also very high for these nodules in the third row. It shows 3D Faster R-CNN works well to detect the nodules from test fold 1. 
\subsection{Nodule Classification}
\begin{figure}[!h]
	\begin{center}
		\includegraphics[width=0.9\linewidth]{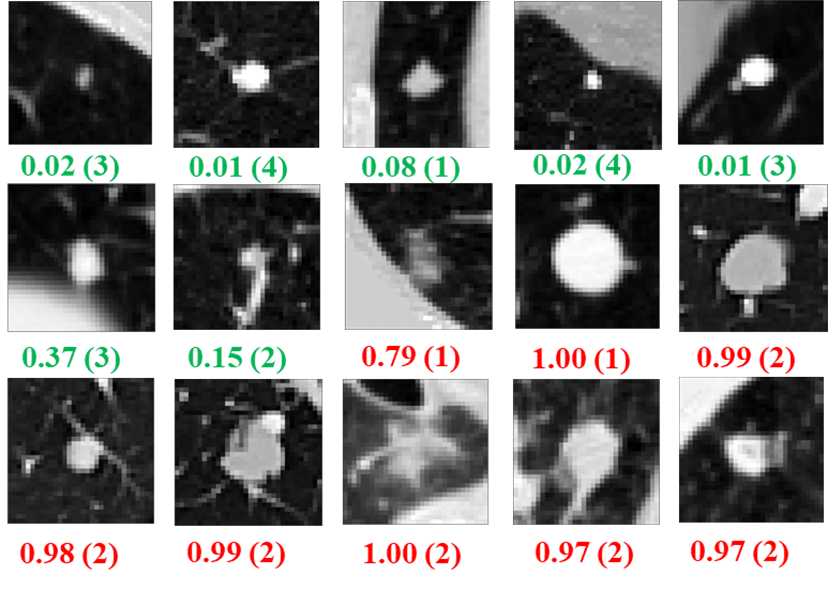} 
		\caption{Visualization of central slices for nodule classification results on test fold 1. We choose nodules that are predicted correct by the DeepLung, but where there is disagreement in the human annotation. The numbers below the nodules are model predicted malignant probabilities followed by which doctor disagreed with the consensus. The first seven nodules are benign nodules. The rest nodules are malignant nodules. The DeepLung performs well for nodule classification.}\label{3fig:classification}
	\end{center}
\end{figure}

We also visualize the nodule classification results from test fold 1 in Fig. \ref{3fig:classification}. We choose nodules that are predicted correct by the DeepLung, but where there is disagreement in the human annotation. The first seven nodules are benign nodules, and the rest nodules are malignant nodules. The numbers below the figures are the DeepLung predicted malignant probabilities, followed by which doctor disagreed with the consensus. For the DeepLung, if the probability is large than 0.5, it predicts malignant. Otherwise, it predicts benign. For an experienced doctor, if a nodule is big and has irregular shape, it has a high probability to be a malignant nodule. 

From Fig. \ref{3fig:classification}, we can observe that doctors mis-diagnose some nodules. The reason is that, humans are not good at processing 3D CT data, which is of low signal to noise ratio. Maybe the doctor cannot find some weak irregular boundaries or consider some tissues as nodule boundaries, which is the possible reason why there are false negatives or false positives for doctors' annotations. In fact, even for high quality 2D natural image, the performance of deep network surpasses that of humans \cite{he2016deep}. They can just observe one slice each time. Some irregular boundaries are vague. The machine learning based methods can learn these complicated rules and high dimensional features from these doctors' annotations, and avoid radiologist's individual biases. From the above analysis, the DeepLung can be considered as a tool to assist the diagnosis for doctors. Combining the DeepLung and doctor's own diagnosis could be an effective way to improve diagnosis accuracy. 

\section{Conclusion}
In this work, we propose a fully automated lung CT cancer diagnosis system, DeepLung, based on deep learning. DeepLung consists of two parts, nodule detection and classification. To fully exploit 3D CT images, we propose two deep 3D convolutional networks based on 3D dual path networks, which is more compact and can yield better performance than residual networks. For nodule detection, we design a 3D Faster R-CNN with 3D dual path blocks and a U-net-like encoder-decoder structure to detect candidate nodules. The detected nodules are subsequently fed to nodule classification network. We use a deep 3D dual path network to extract classification features. Finally, gradient boosting machine with combined features are trained to classify candidate nodules into benign or malignant. Extensive experimental results on public available large-scale datasets, LUNA16 and LIDC-IDRI datasets, demonstrate the superior performance of the DeepLung system.

\chapter{DeepEM: Deep 3D ConvNets With EM For Weakly Supervised Pulmonary Nodule Detection}

\section{Introduction}
{\small{
\begin{figure}[t]
	\begin{center}
		\includegraphics[width=0.9\linewidth]{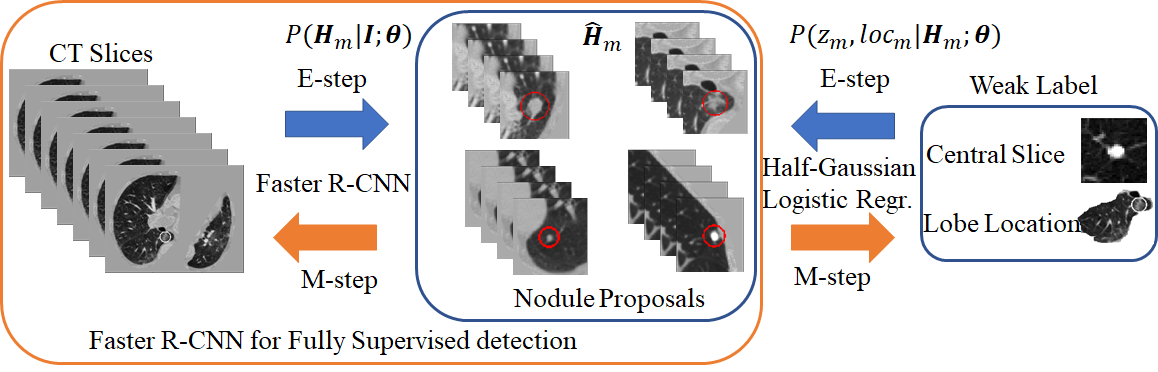}
		\caption{Illustration of DeepEM framework. Faster R-CNN is employed for nodule proposal generation. Half-Gaussian model and logistic regression are employed for central slice and lobe location respectively. In the E-step, we utilize all the observations, CT slices, and weak label to infer the latent variable, nodule proposals, by maximum a posteriori (MAP) or sampling. In the M-step, we employ the estimated proposals to update parameters in the Faster R-CNN and logistic regression.}
		\label{4fig:framework}
	\end{center}
\end{figure}
}}

A prerequisite to utilization of deep learning models is the existence of an abundance of labeled data. However, labels are especially difficult to obtain in the medical image analysis domain. There are multiple contributing factors: a) labeling medical data typically requires specially trained doctors; b) marking lesion boundaries can be hard even for experts because of  low signal-to-noise ratio in many medical images; and c) for CT and magnetic resonance imaging (MRI) images, the annotators need to label the entire 3D volumetric data, which can be costly and time-consuming. Due to these limitations, CT medical image datasets are usually small, which can lead to over-fitting on the training set and, by extension, poor generalization performance on test sets \cite{zhu2017adversarial}.

By contrast, medical institutions have large amount of weakly labeled medical images. In these databases, each medical image is typically associated with an electronic medical report (EMR). Although these reports may not contain explicit information on detection bounding box or segmentation ground truth, it often includes the results of diagnosis, rough locations and summary descriptions of lesions if they exist. We hypothesize that these extra sources of weakly labeled data may be used to enhance the performance of existing detector and improve its generalization capability. 


There are previous attempts to utilize weakly supervised labels to help train machine learning models. Deep multi-instance learning was proposed for lesion localization and whole mammogram classification \cite{zhu2017deep}. The two-stream spatio-temporal ConvNet was proposed to recognize heart frames and localize the heart using only weak labels for whole ultrasound image of fetal heartbeat \cite{gao2017detection}. Different pooling strategies were proposed for weakly supervised localization and segmentation respectively \cite{wang2017chestx,feng2017discriminative,bilen2016weakly}. Papandreou et al. proposed an iterative approach to infer pixel-wise label using image classification label for segmentation \cite{papandreou2015weakly}. Self-transfer learning co-optimized both classification and localization networks for weakly supervised lesion localization \cite{hwang2016self}. Different from these works, we consider nodule proposal as latent variable and propose DeepEM, a new deep 3D convolutional nets with Expectation-Maximization optimization, to mine the big data source of weakly supervised label in EMR as illustrated in Fig. \ref{4fig:framework}. Specifically, we infer the posterior probabilities of the proposed nodules being true nodules, and utilize the posterior probabilities to train nodule detection models. 

\section{DeepEM for Weakly Supervised Detection}
\textbf{Notation} We denote by $\bm{I} \in \mathbb{R}^{h \times w \times s}$ the CT image, where $h$, $w$, and $s$ are image height, width, and number of slices respectively. The nodule bounding boxes for $\bm{I}$ are denoted as ${\bm{H}}=\{\bm{H}_1, \bm{H}_2, \dots, \bm{H}_M\}$, where $\bm{H}_m = \{x_m, y_m, z_m, d_m\}$, the $(x_m, y_m, z_m)$ represents the center of nodule proposal, $d_m$ is the diameter of the nodule proposal, and $M$ is the number of nodules in the image $\bm{I}$. In the weakly supervised scenario, the nodule proposal $\bm{H}$ is a latent variable, and each image $\bm{I}$ is associated with weak label ${\bm{X}}=\{\bm{X}_1, \bm{X}_2, \dots, \bm{X}_M\}$, where $\bm{X}_m=\{{loc}_m, z_m\}$, ${loc}_m \in \{1,2,3,4,5,6\}$ is the location (right upper lobe, right middle lobe, right lower lobe, left upper lobe, lingula, left lower lobe) of nodule $\bm{H}_m$ in the lung, and $z_m$ is the central slice of the nodule. 

For fully supervised detection, the objective function is to maximize the log-likelihood function for observed nodule ground truth $\bm{H}$ given image $\bm{I}$ as 
\begin{equation}
	\mathcal{L}(\bm{\theta}) = \log P(\bm{H} \cup \bm{\bar{H}} | \bm{I}; \bm{\theta}) = \frac{1}{M}\sum_{m=1}^{M} \log P(\bm{H}_m | \bm{I}; \bm{\theta})- \frac{1}{N}\sum_{n=1}^{N} \log P(\bm{\bar{H}}_n | \bm{I}; \bm{\theta}) ,\label{4eq:loglike} %
\end{equation}
where $\bm{\bar{H}}=\{\bm{\bar{H}}_1, \bm{\bar{H}}_2, \dots, \bm{\bar{H}}_N\}$ are hard negative proposals we mine in real time during training \cite{kaimingfasterrcnn}, and $\bm{\theta}$ is the weights of deep 3D ConvNet. We employ Faster R-CNN with 3D Res18 for the fully supervised detection because of its superior performance.

For weakly supervised detection, nodule proposal $\bm{H}$ can be considered as a latent variable. Using this framework, image $\bm{I}$ and weak label \(\bm{X}=\{({loc}_1, z_1), ({loc}_2, \\ z_2), \dots, ({loc}_M, z_M)\}\) can be considered as observations. The joint distribution is
\begin{equation}
\begin{aligned}
P(\bm{I}, \bm{H}, \bm{X}; \bm{\theta}) &= P(\bm{I}) \prod_{m=1}^{M} \big( P(\bm{H}_m|\bm{I}; \bm{\theta}) P(\bm{X}_m | \bm{H}_m; \bm{\theta}) \big) \\
&= P(\bm{I}) \prod_{m=1}^{M} \big( P(\bm{H}_m|\bm{I}; \bm{\theta}) P({loc}_m | \bm{H}_m; \bm{\theta}) P({z}_m | \bm{H}_m; \bm{\theta}) \big).\label{4eq:pgm} 
\end{aligned}
\end{equation}
To model $P({z}_m | \bm{H}_m; \bm{\theta})$, we propose using a half-Gaussian distribution based on nodule size distribution because $z_m$ is correct if it is within the nodule area (center slice of $\bm{H}_m$ as ${z}_{{\bm{H}}_m}$, and nodule size $\sigma$ can be empirically estimated based on existing data) for nodule detection in Fig. \ref{4fig:halfnorm}(a). For lung lobe prediction $P({loc}_m | \bm{H}_m; \bm{\theta})$, a logistic regression model is used based on relative value of nodule center $({x}_{{\bm{H}}_m}, {y}_{{\bm{H}}_m}, {z}_{{\bm{H}}_m})$ after lung segmentation. That is
\begin{equation}
	\begin{aligned}
	P(z_m, {loc}_m | \bm{H}_m ; \bm{\theta}) = \frac{2}{\sqrt{2 \pi {\sigma}^2}} \exp \big( -\frac{|z_m - {z}_{\bm{H}_m} |^2}{2 {\sigma}^2} \big) \frac{\exp(\bm{f}(\bm{H}_m) \bm{\theta}_{{loc}_m})}{\sum_{{{loc}_m}=1}^{6}\exp(\bm{f}(\bm{H}_m) \bm{\theta}_{{loc}_m})},\label{4eq:weaklikeli}
	\end{aligned}
\end{equation}
where $\bm{\theta}_{{loc}_m}$ is the associated weights with lobe location ${loc}_m$ for logistic regression, feature $\bm{f}(\bm{H}_m) = (\frac{{x}_{{\bm{H}}_m}}{{x}_{\bm{I}}}, \frac{{y}_{{\bm{H}}_m}}{{y}_{\bm{I}}}, \frac{{z}_{{\bm{H}}_m}}{{z}_{\bm{I}}})$, and $({x}_{\bm{I}}, {y}_{\bm{I}}, {z}_{\bm{I}})$ is the total size of image $\bm{I}$ after lung segmentation. In the experiments, we found the logistic regression converges quickly and is stable.

The expectation-maximization (EM) is a commonly used approach to optimize the maximum log-likelihood function when there are latent variables in the model. We employ the EM algorithm to optimize deep weakly supervised detection model in equation \ref{4eq:pgm}. The expected complete-data log-likelihood function given previous estimated parameter ${\bm{\theta}}^{\prime}$ in deep 3D Faster R-CNN is
\begin{equation}
\begin{aligned}
Q(\bm{\theta}; \bm{\theta^{\prime}}) = & \frac{1}{M}\sum_{m=1}^{M}  \mathbb{E}_{P(\bm{H}_m | \bm{I}, z_m, {loc}_m; {\bm{\theta}}^{\prime})} \big[ \log P(\bm{H}_m|\bm{I}; \bm{\theta}) \\ &+ \log P(z_m, {loc}_m|{\bm{H}}_m; \bm{\theta}) \big] -  \mathbb{E}_{Q(\bm{\bar{H}}_n | \bm{z})}\big[ \log P(\bm{\bar{H}}_n | \bm{I}; \bm{\theta}) \big], \label{4eq:expect}
\end{aligned}
\end{equation}
where $\bm{z} = \{z_1, z_2, \dots, z_m\}$. In the implementation, we only keep hard negative proposals far away from weak annotation $\bm{z}$ to simplify $Q(\bm{\bar{H}}_n | \bm{z})$. The posterior distribution of latent variable $\bm{H}_m$ can be calculated by
\begin{equation}
\begin{aligned}
P(\bm{H}_m | \bm{I}, z_m, {loc}_m; \bm{{\theta}^{\prime}}) &\propto P(\bm{H}_m | \bm{I}; \bm{{\theta}^{\prime}}) P(z_m, {loc}_m | \bm{H}_m; \bm{{\theta}^{\prime}}).\label{4eq:posteri}
\end{aligned}
\end{equation}
Because Faster R-CNN yields a large number of proposals, we first use hard threshold (-3 before sigmoid function) to remove proposals of small confident probability, then employ non-maximum suppression (NMS) with intersection over union (IoU) as 0.1. We then employ two schemes to approximately infer the latent variable $\bm{H}_m$: maximum a posteriori (MAP) or sampling. \\ 
\textbf{DeepEM with MAP} We only use the proposal of maximal posterior probability to calculate the expectation.
\begin{equation}
	\hat{\bm{H}}_m = {\arg \max }_{\bm{H}_m} P(\bm{H}_m | \bm{I}; \bm{{\theta}^{\prime}}) P(z_m, {loc}_m | \bm{H}_m; \bm{{\theta}^{\prime}})\label{4eq:map} 
\end{equation} 
\textbf{DeepEM with Sampling} We approximate the distribution by sampling $\hat{M}$ proposals $\hat{\bm{H}}_m$ according to normalized equation \ref{4eq:posteri}. The expected log-likelihood function in equation \ref{4eq:expect} becomes
\begin{equation}
\begin{aligned}
Q(\bm{\theta}; \bm{\theta^{\prime}}) = &\frac{1}{M \hat{M}}\sum_{m=1}^{M}  \sum_{\hat{\bm{H}}_m}^{\hat{M}} \big( \log P(\hat{\bm{H}}_m|\bm{I}; \bm{\theta}) + \log P(z_m, {loc}_m|{\hat{\bm{H}}}_m; \bm{\theta}) \big) \\ & + \mathbb{E}_{Q(\bm{\bar{H}}_n | \bm{z})}\big[ \log P(\bm{\bar{H}}_n | \bm{I}; \bm{\theta}) \big]. \label{4eq:deepemsamp}
\end{aligned}
\end{equation}

After obtaining the expectation of complete-data log-likelihood function in equation \ref{4eq:expect}, we can update the parameters $\bm{\theta}$ by 
\begin{equation}
	\hat{\bm{\theta}} = \arg \max Q(\bm{\theta} ; {\bm{\theta}}^{\prime}).\label{4eq:mstep}
\end{equation}
The M-step in equation \ref{4eq:mstep} can be conducted by stochastic gradient descent commonly used in deep network optimization for equation \ref{4eq:loglike}. Our entire algorithm is outlined in algorithm \ref{4alg:deepem}.
{\small{
\begin{algorithm}[t]
	\caption{DeepEM for Weakly Supervised Detection}\label{4alg:deepem}
	\begin{algorithmic}[1]
		\INPUT Fully supervised dataset $D_F = \{({\bm{I}}, {\bm{H}})_i\}_{i=1}^{N_F}$, weakly supervised dataset $D_W = \{({\bm{I}}, {\bm{X}})_i\}_{i=1}^{N_W}$, 3D Faster R-CNN and logistic regression parameters $\bm{\theta}$.
		\BState \emph{Initialization}: Update weights $\bm{\theta}$ by maximizing equation \ref{4eq:loglike} using data from $D_F$.
		\BState \emph{for epoch = 1 to \#TotalEpochs}:
		\WEAKTRAIN
		\State \hspace{1.3em} Use Faster R-CNN model ${\bm{\theta}}^{\prime}$ to obtain proposal probability $P(\bm{H}_m | \bm{I}; \bm{{\theta}^{\prime}})$ for weakly supervised data sampled from $D_W$.
		\State \hspace{1.3em} Remove proposals with small probabilities and NMS. 
		\BState \hspace{1.3em} \emph{for m = 1 to M}:   \hspace{1.3em} $\triangleright\triangleright\triangleright$ Each weak label
		\State \hspace{3.0em} Calculate $P(z_m, {loc}_m | \bm{H}_m ; \bm{\theta})$ for each proposal by equation \ref{4eq:weaklikeli}.
		\State \hspace{3.0em} Estimate posterior distribution $P(\bm{H}_m | \bm{I}, z_m, {loc}_m; \bm{{\theta}^{\prime}})$ by equation \ref{4eq:posteri} with normalization.
		\State \hspace{3.0em} Employ MAP by equation \ref{4eq:map} or Sampling to obtain the inference of $\bm{H}_m$.
		\State \hspace{1.3em} Obtain the expect log-likelihood function by equation \ref{4eq:expect} using the estimated proposal (MAP) or by equation \ref{4eq:deepemsamp} (Sampling).
		\State \hspace{1.3em} Update parameter by equation \ref{4eq:mstep}.
		\FULLTRAIN
		\State	\hspace{1.3em} Update weights $\bm{\theta}$ by maximizing equation \ref{4eq:loglike} using fully supervised data $D_F$. 
	\end{algorithmic}
\end{algorithm}
}}
\section{Experiments}
We used 3 datasets, LUNA16 dataset \cite{setio2017validation} as fully supervised nodule detection, NCI NLST\footnote{https://biometry.nci.nih.gov/cdas/datasets/nlst/} dataset as weakly supervised detection, Tianchi Lung Nodule Detection\footnote{https://tianchi.aliyun.com/} dataset as holdout dataset for test only. LUNA16 dataset is the largest publicly available dataset for pulmonary nodules detection \cite{setio2017validation}. LUNA16 dataset removes CTs with slice thickness greater than 3mm, slice spacing inconsistent or missing slices, and consist of 888 low-dose lung CTs which have explicit patient-level 10-fold cross validation split. NLST dataset consists of hundreds of thousands of lung CT images associated with electronic medical records (EMR). In this work, we focus on nodule detection based on image modality and only use the central slice and nodule location as weak supervision from the EMR. As part of data cleansing, we remove negative CTs, CTs with slice thickness greater than 3mm and nodule diameter less than 3mm. After data cleaning, we have 17,602 CTs left with 30,951 weak annotations. In each epoch, we randomly sample $\frac{1}{16}$ CT images for weakly supervised training because of the large numbers of weakly supervised CTs. Tianchi dataset contains 600 training low-dose lung CTs and 200 validation low-dose lung CTs for nodule detection. The annotations are location centroids and diameters of the pulmonary nodules, and do not have less than 3mm diameter nodule, which are the same with those on LUNA16 dataset. 

\textbf{Parameter estimation in $P({z}_m | \bm{H}_m; \bm{\theta})$} 
If the current $z_m$ is within the nodule, it is a true positive proposal. We can model $|z_m-z_{{\bm{H}}_m}|$ using a half-Gaussian distribution shown as the red dash line in Fig. \ref{4fig:halfnorm}(a). The parameters of the half-Gaussian is estimated from the LUNA16 data empirically. Because LUNA16 removes nodules of diameter less than 3mm, we use the truncated half-Gaussian to model the central slice $z_m$ as $\max(|z_m-z_{{\bm{H}}_m}|-\mu, 0)$, where $\mu$ is the mean of related Gaussian as the minimal nodule radius with 1.63. 
{\small{
\begin{figure}[t]
	\begin{center}
		\begin{minipage}{0.36\textwidth}
			\centerline{\includegraphics[width=0.9\linewidth]{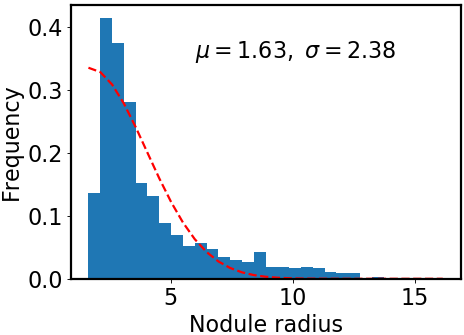}}
			\center{(a)}
		\end{minipage}
		\begin{minipage}{0.63\textwidth}
			\centerline{\includegraphics[width=1.08\linewidth]{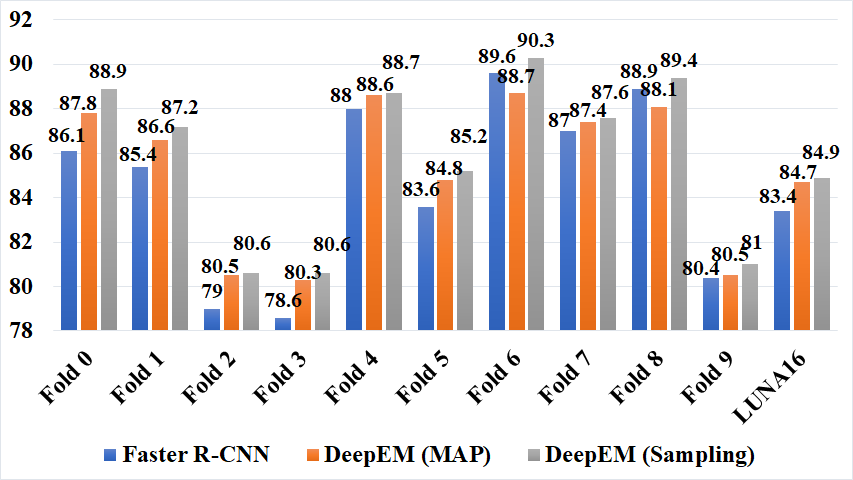}}
			\center{(b)}
		\end{minipage}
		\caption{(a)Empirical estimation of half-Gaussian model for $P({z}_m | \bm{H}_m; \bm{\theta})$ on LUNA16. (b) FROC (\%) comparison among Faster R-CNN, DeepEM with MAP, DeepEM with Sampling on LUNA16.}
		\label{4fig:halfnorm}
	\end{center}
\end{figure}}}

\textbf{Performance comparisons on LUNA16} 
We conduct 10-fold cross validation on LUNA16 to validate the effectiveness of DeepEM. The baseline benchmark method used is Faster R-CNN with 3D Res18 network, henceforth denoted as \textbf{Faster R-CNN}, trained on the supervised data \cite{kaimingfasterrcnn,zhu2018deeplung}. We use the train set of each validation split to train the \textbf{Faster R-CNN}, and obtain ten models in the ten-fold cross validation. Then we employ it to model $P(\bm{H}_m | \bm{I}; \bm{{\theta}^{\prime}})$ for weakly supervised detection scenario. Two inference scheme for ${\bm{H}}_m$ are used in DeepEM denoted as \textbf{DeepEM (MAP)} and \textbf{DeepEM (Sampling)}. In the proposal inference of DeepEM with Sampling, we sample two proposals for each weak label because the average number of nodules each CT is 1.78 on LUNA16. The evaluation metric, Free receiver operating characteristic (FROC), is the average recall rate at the average number of false positives at 0.125, 0.25, 0.5, 1, 2, 4, 8 per scan, which is the official evaluation metric for LUNA16 and Tianchi \cite{setio2017validation}. 

From Fig. \ref{4fig:halfnorm}(b), DeepEM with MAP improves about 1.3\% FROC over Faster R-CNN and DeepEM with Sampling improves about 1.5\% FROC over Faster R-CNN on average on LUNA16 when incorporating weakly labeled data from NLST. We hypothesize the greater improvement of DeepEM with Sampling over DeepEM with MAP is that MAP inference is greedy and can get stuck at a local minimum while the nature of sampling may allow DeepEM with Sampling to escape these local minimums during optimization.

\textbf{Performance comparisons on holdout test set from Tianchi}
We employed a holdout test set from Tianchi to validate each model from 10-fold cross validation on LUNA16. The results are summarized in Table \ref{4tab:tianchifrocperformance}. We can see DeepEM utilizing weakly supervised data improves 3.9\% FROC on average over Faster R-CNN. The improvement on holdout test data validates DeepEM as an effective model to exploit potentially large amount of weak data from electronic medical records (EMR) which would not require further costly annotation by expert doctors and can be easily obtained from hospital associations.
{\small{
\begin{table}[t]
	\centering
	\caption{FROC (\%) comparisons among Faster R-CNN with 3D ResNet18, DeepEM with MAP, DeepEM with Sampling on Tianchi.}\label{4tab:tianchifrocperformance}
	\begin{tabular}{c|c|c|c|c|c|c|c|c|c|c|c}
		\hline
		Fold&0&1&2&3&4&5&6&7&8&9&Average\\
		\hline
		Faster R-CNN&72.8&70.8&69.8&71.9&76.4&73.0&71.3&74.7&72.9&71.3&72.5\\
		\hline
		DeepEM (MAP)&77.2&75.8&75.8&74.9&77.0&75.5&77.2&75.8&76.0&74.7&76.0\\
		\hline
		DeepEM (Sampling)&77.4&75.8&75.9&75.0&77.3&75.0&77.3&76.8&77.7&75.8&76.4\\
		\hline
	\end{tabular}
\end{table}}}
{\small{
\begin{figure}[t]
	\begin{center}
		\centerline{\includegraphics[width=\linewidth]{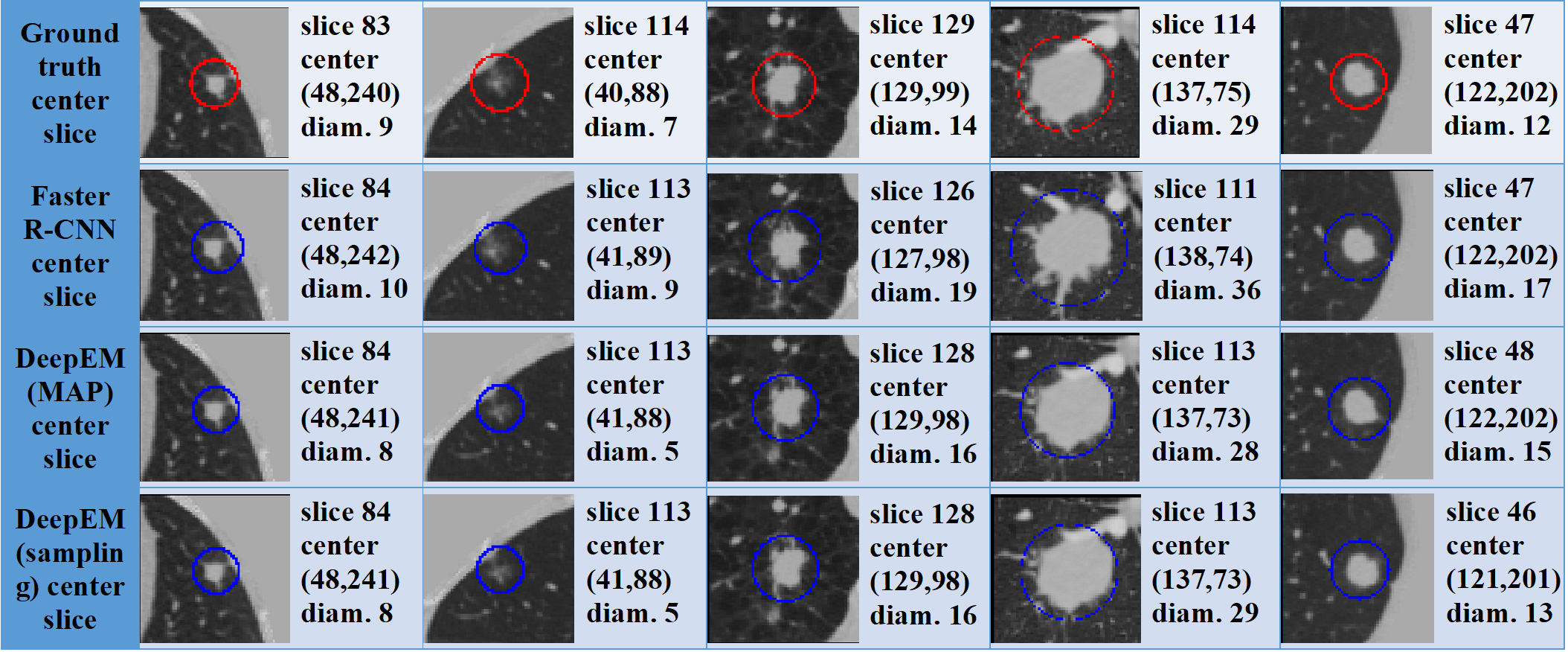}}
		\caption{Detection visual comparison among Faster R-CNN, DeepEM with MAP and DeepEM with Sampling on nodules randomly sampled from Tianchi. DeepEM provides more accurate detection (central slice, center and diameter) than Faster R-CNN.}
		\label{4fig:vis}
	\end{center}
\end{figure}}}

\textbf{Visualizations} 
We compare Faster R-CNN with the proposed DeepEM visually in Fig. \ref{4fig:halfnorm}(b). We randomly choose nodules from Tianchi. From Fig. \ref{4fig:halfnorm}(b), DeepEM yields better detection for nodule center and tighter nodule diameter which demonstrates DeepEM improves the existing detector by exploiting weakly supervised data.
\section{Conclusion}
In this chapter, we have focused on the problem of detecting pulmonary nodules from lung CT images, which previously has been formulated as a supervised learning problem and requires a large amount of training data with the locations and sizes of nodules precisely labeled. Here we propose a new framework, called DeepEM, for pulmonary nodule detection by taking advantage of abundantly available weakly labeled data extracted from EMRs. We treat each nodule proposal as a latent variable, and infer the posterior probabilities of proposal nodules being true ones conditioned on images and weak labels. The posterior probabilities are further fed to the nodule detection module for training. We use an EM algorithm to train the entire model end-to-end. Two schemes, maximum a posteriori (MAP) and sampling, are used for the inference of proposals. Extensive experimental results demonstrate the effectiveness of DeepEM for improving current state of the art nodule detection systems by utilizing readily available weakly supervised detection data. Although our method is built upon the specific application of pulmonary nodule detection, the framework itself is fairly general and can be readily applied to other medical image deep learning applications to take advantage of weakly labeled data.

\chapter{AnatomyNet: Deep Learning for Fast and Fully Automated Whole-volume Segmentation of  Head and Neck Anatomy} 

\section{Introduction}
\label{5sec:intro}  
Head and neck cancer is one of the most common cancers around the world \cite{torre2015global}. Radiation therapy is the primary method for treating patients with head and neck cancers. The planning of the radiation therapy relies on accurate organs-at-risks (OARs) segmentation \cite{han2008atlas}, which is usually undertaken by radiation therapists with laborious manual delineation. Computational tools that automatically segment the anatomical regions can greatly alleviate doctors' manual efforts if these tools can delineate anatomical regions accurately with a reasonable amount of time \cite{sharp2014vision}. 

There is a vast body of literature on automatically segmenting anatomical structures from CT or MRI images. Here we focus on reviewing literature related to head and neck (HaN) CT anatomy segmentation. Traditional anatomical segmentation methods use primarily atlas-based methods, producing segmentations by aligning new images to a fixed set of manually labelled exemplars \cite{raudaschl2017evaluation}. Atlas-based segmentation methods typically undergo a few steps, including preprocessing, atlas creation, image registration, and label fusion. As a consequence, their performances can be affected by various factors involved in each of these steps, such as methods for creating atlas \cite{han2008atlas,voet2011does,isambert2008evaluation,fritscher2014automatic,commowick2008atlas,sims2009pre,fortunati2013tissue,verhaart2014relevance,wachinger2015contour}, methods for label fusion \cite{duc2015validation,duc2015validation,fortunati2015automatic}, and  methods for registration \cite{zhang2007automatic,chen2010combining,han2008atlas,duc2015validation,fritscher2014automatic,fortunati2013tissue,wachinger2015contour,qazi2011auto,leavens2008validation}.  Although atlas-based methods are still very popular and by far the most widely used methods in anatomy segmentation, their main limitation is the difficulty to handle anatomy variations among patients because they use a fixed set of atlas. In addition, it is computationally intensive and can take many minutes to complete one registration task even with most efficient implementations \cite{xu2018use}.



Instead of aligning images to a fixed set of exemplars, learning-based methods trained to directly segment OARs without resorting to reference exemplars have also been tried \cite{tam2018automated,wu2018auto,tong2018hierarchical,pednekar2018image,wang2018hierarchical}. However, most of the learning-based methods require laborious preprocessing steps, and/or hand-crafted image features. As a result, their performances tend to be less robust than registration-based methods.

Recently, deep convolutional models have shown great success for biomedical image segmentation \cite{ronneberger2015u}, and have been introduced to the field of HaN anatomy segmentation \cite{fritscher2016deep,ibragimov2017segmentation,ren2018interleaved,hansch2018comparison}. However, the existing HaN-related deep-learning-based methods either use sliding windows working on patches that cannot capture global features, or rely on atlas registration to obtain highly accurate small regions of interest in the preprocessing. What is more appealing are models that receive the whole-volume image as input without heavy-duty preprocessing, and then directly output the segmentations of all interested anatomies. 

In this work, we study the feasibility and performance of constructing and training a deep neural net model that jointly segment all OARs in a fully end-to-end fashion, receiving  raw whole-volume HaN CT images as input and generating the masks of all OARs in one shot. The success of such a system can improve the current performance of automated anatomy segmentation by simplifying the entire computational pipeline, cutting computational cost and improving segmentation accuracy.


There are, however, a number of obstacles that need to overcome in order to make such a deep convolutional neural net based system successful. First, in designing network architectures, we ought to keep the maximum capacity of GPU memories in mind. Since whole-volume images are used as input, each image feature map will be 3D, limiting the size and number of feature maps at each layer of the neural net due to memory constraints.  Second, OARs contain organs/regions of variable sizes, including some OARs with very small sizes. Accurately segmenting these small-volumed structures is always a challenge. Third, existing datasets of HaN CT images contain data collected from various sources with non-standardized annotations. In particular, many images in the training data contain annotations of only a subset of OARs. How to effectively handle missing annotations needs to be addressed in the design of the training algorithms. 

Here we propose a deep learning based framework, called AnatomyNet, to segment OARs using a single network, trained end-to-end. The network receives whole-volume CT images as input, and outputs the segmented masks of all OARs. Our method requires minimal pre- and post-processing, and utilizes features from all slices to segment anatomical regions. We overcome the three major obstacles outlined above through designing a novel network architecture and utilizing novel loss functions for training the network. 

More specifically, our major contributions include the following. First, we extend the standard U-Net model for 3D HaN image segmentation by incorporating a new feature extraction component, based on squeeze-and-excitation (SE) residual blocks \cite{hu2017squeeze}. Second, we propose a new loss function for better segmenting small-volumed structures. Small volume segmentation suffers from the imbalanced data problem, where the number of voxels inside the small region is much smaller than those outside, leading to the difficulty of training. New classes of loss functions have been proposed to address this issue, including Tversky loss \cite{salehi2017tversky}, generalized Dice coefficients \cite{crum2006generalized,sudre2017generalised}, focal loss \cite{lin2017focal}, sparsity label assignment deep multi-instance learning \cite{zhu2017deep}, and exponential logarithm loss. However, we found none of these solutions alone was adequate to solve the extremely data imbalanced problem (1/100,000) we face in segmenting small OARs, such as optic nerves and chiasm, from HaN images. We propose a new loss based on the combination of Dice scores and focal losses, and empirically show that it leads to better results than other losses. Finally, to tackle the missing annotation problem, we train the AnatomyNet with masked and weighted loss function to account for missing data and to balance the contributions of the losses originating from different OARs.  


To train and evaluate the performance of AnatomyNet, we curated a dataset of 261 head and neck CT images from a number of publicly available sources. We carried out systematic experimental analyses on various components of the network, and demonstrated their effectiveness by comparing with other published methods.  When benchmarked on the test dataset from the MICCAI 2015 competition on HaN segmentation, the AnatomyNet outperformed the state-of-the-art method by 
\textbf{3.3\%} in terms of  Dice coefficient (DSC), averaged over nine anatomical structures.


The rest of the paper is organized as follows. Section \ref{5sec:anatomynet} describes the network structure and SE residual block of AnatomyNet. The designing of the loss function for AnatomyNet is present in Section \ref{5sec:extremeimbalance}. How to handle missing annotations is addressed in Section \ref{5sec:missingannot}. Section \ref{5sec:result} validates the effectiveness of the proposed networks and components. Discussions and limitations are in Section \ref{5sec:discus}. We conclude the work in Section \ref{5sec:conclu}.

\section{Materials and Methods}\label{5sec:method}
Next we describe our deep learning model to delineate OARs from head and neck CT images. Our model receives whole-volume HaN CT images of a patient as input and outputs the 3D binary masks of all OARs at once. The dimension of a typical HaN CT is around $178 \times 512 \times 512$, but the sizes can vary across different patients because of image cropping and different settings.  In this work, we focus on segmenting nine OARs most relevant to head and neck cancer radiation therapy - brain stem, chiasm, mandible, optic nerve left, optic nerve right, parotid gland left, parotid gland right, submandibular gland left, and submandibular gland right. Therefore, our model will produce nine 3D binary masks for each whole volume CT. 

\subsection{Data}\label{5sec:data}
Before we introduce our model, we first describe the curation of training and testing data. Our data consists of whole-volume CT images together with manually generated binary masks of the nine anatomies described above. There were collected from four publicly available sources: 1) DATASET 1 (38 samples) consists of the training set from the MICCAI Head and Neck Auto Segmentation Challenge 2015 \cite{raudaschl2017evaluation}. 2) DATASET 2 (46 samples) consists of CT images from the Head-Neck Cetuximab collection, downloaded from The Cancer Imaging Archive (TCIA)\footnote{https://wiki.cancerimagingarchive.net/} \cite{clark2013cancer}. 3) DATASET 3 (177 samples) consists of CT images from four different institutions in Qu\'ebec, Canada \cite{vallieres2017radiomics}, also downloaded from TCIA \cite{clark2013cancer}. 4) DATATSET 4 (10 samples) consists of the test set from the MICCAI HaN Segmentation Challenge 2015. We combined the first three datasets and used the aggregated data as our training data, altogether yielding 261 training samples.  DATASET 4 was used as our final evaluation/test dataset so that we can benchmark our performance against published results evaluated on the same dataset.  Each of the training and test samples contains both head and neck images and the corresponding manually delineated OARs.

In generating these datasets, We carried out several data cleaning steps, including 1) mapping annotation names named by different doctors in different hospitals into unified annotation names, 2) finding correspondences between the annotations and the CT images, 3) converting annotations in the radiation therapy format into usable ground truth label mask, and 4) removing chest from CT images to focus on head and neck anatomies.  We have taken care to make sure that the four datasets described above are non-overlapping to avoid any potential pitfall of inflating testing or validation performance. 

\subsection{Network architecture}\label{5sec:anatomynet}
\begin{figure*} [ht]
	\begin{center}
		\begin{tabular}{c} 
			\includegraphics[height=7.5cm]{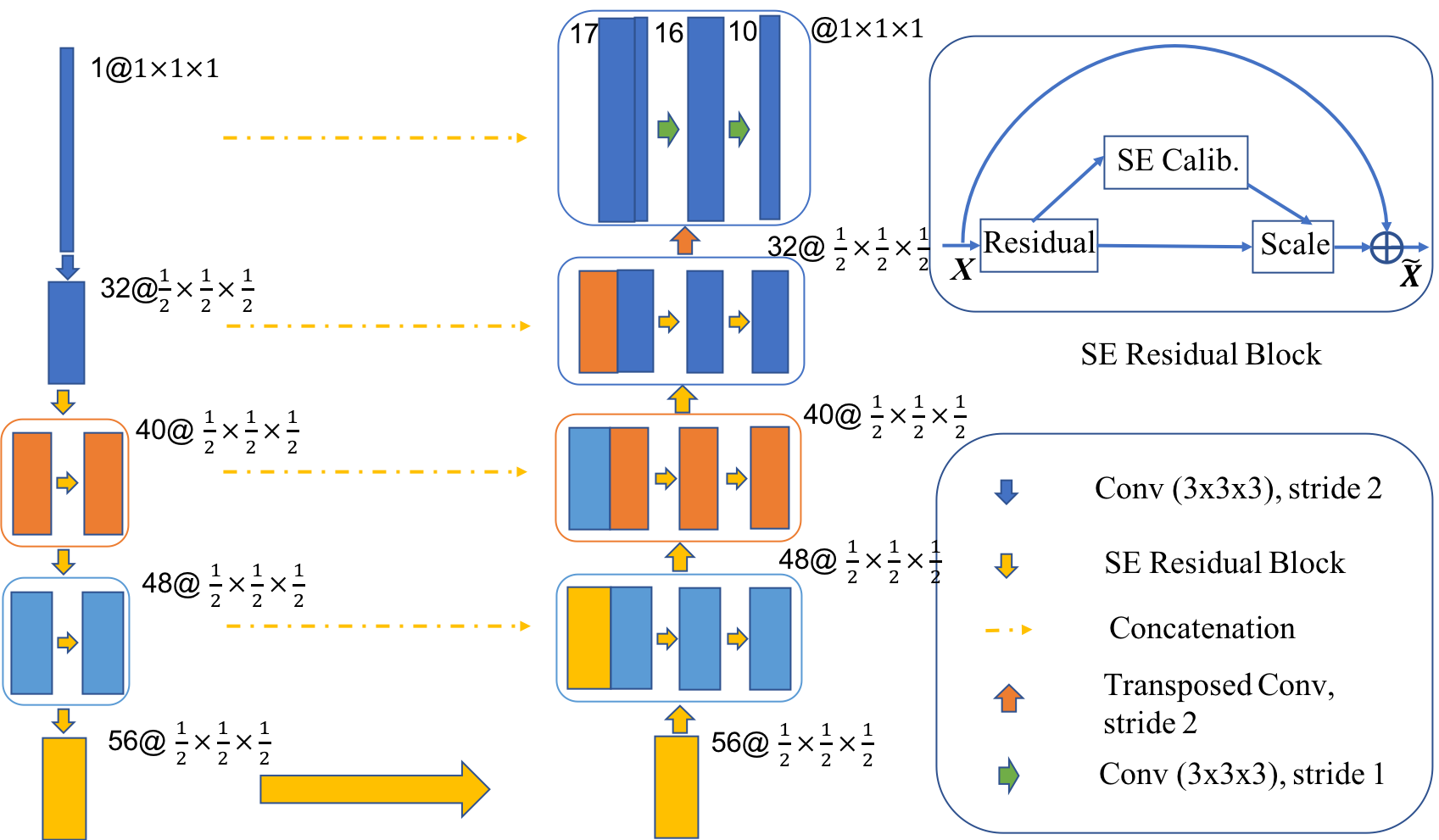}
		\end{tabular}
	\end{center}
	\caption{\label{5fig:seunet} 
		The AnatomyNet is a variant of U-Net with only one down-sampling and squeeze-and-excitation (SE) residual building blocks. The number before symbol @ denotes the number of output channels, while the number after the symbol denotes the size of feature map relative to the input. In the decoder, we use concatenated features. Hybrid loss with dice loss and focal loss is employed to force the model to learn not-well-classified voxels. Masked and weighted loss function is used for ground truth with missing annotations and balanced gradient descent respectively. The decoder layers are symmetric with the encoder layers. The SE residual block is illustrated in the upper right corner.}
\end{figure*}

We take advantage of the robust feature learning mechanisms obtained from squeeze-and-excitation (SE) residual blocks \cite{hu2017squeeze}, and incorporate them into a modified U-Net architecture for medical image segmentation. We propose a novel three dimensional U-Net with squeeze-and-excitation (SE) residual blocks and hybrid focal and dice loss for anatomical segmentation as illustrated in Fig. \ref{5fig:seunet}. 

The AnatomyNet is a variant of 3D U-Net \cite{ronneberger2015u,zhu2018deeplung,zhu2018deepem}, one of the most commonly used neural net architectures in biomedical image segmentation. The standard U-Net contains multiple down-sampling layers via max-pooling or convolutions with strides over two. Although they are beneficial to learn high-level features for segmenting complex, large anatomies, these down-sampling layers can hurt the segmentation of small anatomies such as optic chiasm, which occupy only a few slices in HaN CT images. We design the AnatomyNet with only one down-sampling layer to account for the trade-off between GPU memory usage and network learning capacity. The down-sampling layer is used in the first encoding block so that the feature maps and gradients in the following layers occupy less GPU memory than other network structures. Inspired by the effectiveness of squeeze-and-excitation residual features on image object classification, we design 3D squeeze-and-excitation (SE) residual blocks in the AnatomyNet for OARs segmentation. The SE residual block adaptively calibrates residual feature maps within each feature channel. 
The 3D SE Residual learning extracts 3D features from CT image directly by extending two-dimensional squeeze, excitation, scale and convolutional functions to three-dimensional functions. It can be formulated as 
\begin{equation}
\label{5eq:seres}
\begin{aligned}
{\bm{X}^{r}}&=\bm{F}(\bm{X}) \, , \\
z_{k} &= {\bm{F}}_{sq}({\bm{X}^{r}}_{k}) = \frac{1}{S \times H \times W} \sum_{s=1}^{S} \sum_{h=1}^{H} \sum_{w=1}^{W}{x}^{r}_{k}(s,h,w) \, , \\
\bm{z} &= [z_1, z_2, \cdots, z_{k}, \cdots, z_{K}] \, , \\
\bm{s} &= {\bm{F}}_{ex} (\bm{z}, \bm{W}) = \bm{\sigma}(\bm{W}_2 \bm{G}(\bm{W}_1 \bm{z})) \, , \\
{\tilde{\bm{X}}}_{k} &= {\bm{F}}_{scale}({\bm{X}^{r}}_{k}, s_{k}) = s_{k} {\bm{X}^{r}}_{k} \, , \\
\tilde{\bm{X}} &= [{\tilde{\bm{X}}}_1, {\tilde{\bm{X}}}_2, \cdots, {\tilde{\bm{X}}}_{k}, \cdots, {\tilde{\bm{X}}}_{K}] \, , \\
\bm{Y} &= \bm{G} (\tilde{\bm{X}} + \bm{X}) \, ,
\end{aligned}
\end{equation}
where ${{\bm{X}^{r}}_{k}} \in {\mathbb{R}}^3$ denotes the feature map of one channel from the residual feature $\bm{X}^{r}$. ${\bm{F}}_{sq}$ is the squeeze function, which is global average pooling here. $S, H, W$ are the number of slices, height, and width of $\bm{X}^{r}$ respectively. ${\bm{F}}_{ex}$ is the excitation function, which is parameterized by two layer fully connected neural networks here with activation functions $\bm{G}$ and $\bm{\sigma}$, and weights ${\bm{W}}_1$ and ${\bm{W}}_2$. The $\bm{\sigma}$ is the sigmoid function. The $\bm{G}$ is typically a ReLU function, but we use LeakyReLU in the AnatomyNet \cite{maas2013rectifier}. We use the learned scale value $s_{k}$ to calibrate the residual feature channel ${\bm{X}^{r}}_{k}$, and obtain the calibrated residual feature $\tilde{\bm{X}}$ . The SE block is illustrated in the upper right corner in Fig. \ref{5fig:seunet}.

The AnatomyNet replaces the standard convolutional layers in the U-Net with SE residual blocks to learn effective features. The input of AnatomyNet is a cropped whole-volume head and neck CT image. We remove the down-sampling layers in the second, third, and fourth encoder blocks to improve the performance of segmenting small anatomies. In the output block, we concatenate the input with the transposed convolution feature maps obtained from the second last block. After that, a convolutional layer with 16 $3 \times 3 \times 3$ kernels and LeakyReLU activation function is employed. In the last layer, we use a convolutional layer with 10 $3 \times 3 \times 3$ kernels and soft-max activation function to generate the segmentation probability maps for nine OARs plus background. 

\subsection{Loss function}\label{5sec:extremeimbalance}
\begin{figure} [ht]
	\begin{center}
		\begin{tabular}{c} 
				\includegraphics[height=5.7cm]{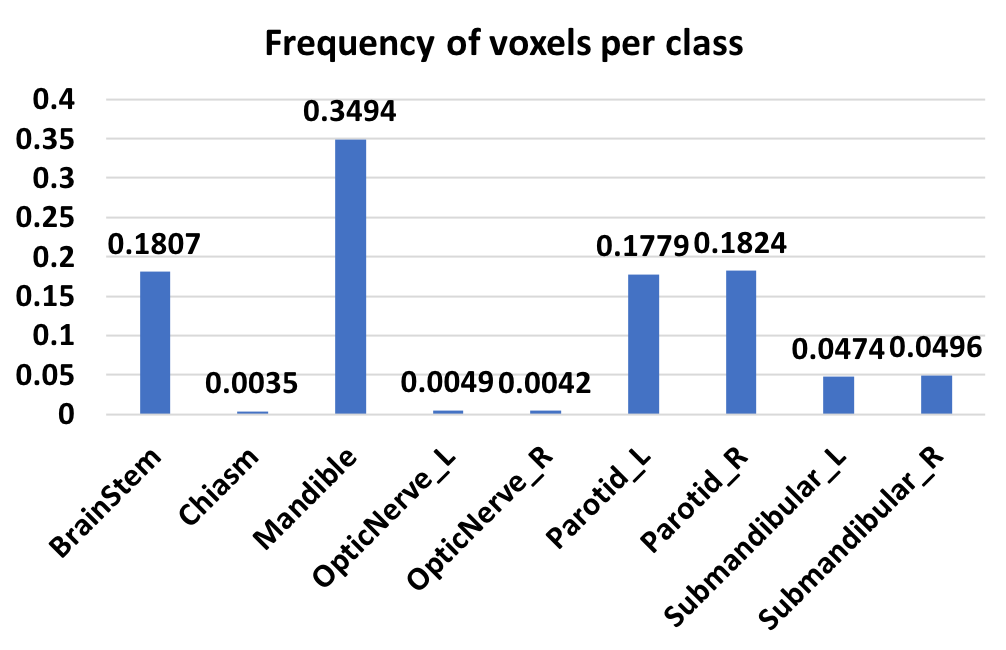}
		\end{tabular}
	\end{center}
	\caption{ \label{5fig:pixeldist} 
		The frequency of voxels for each class on MICCAI 2015 challenge dataset. Background takes up 98.18\% of all the voxels. Chiasm takes only 0.35\% of the foreground which means it only takes about 1/100,000 of the whole-volume CT image. The huge imbalance of voxels in small-volumed organs causes difficulty for small-volumed organ segmentation. }
\end{figure}

Small object segmentation is always a challenge in semantic segmentation. From the learning perspective, the challenge is caused by imbalanced data distribution, because image semantic segmentation requires pixel-wise labeling and small-volumed organs contribute less to the loss. In our case, the small-volumed organs, such as optic chiasm, only take about 1/100,000 of the whole-volume CT images from Fig. \ref{5fig:pixeldist}. The dice loss, the minus of dice coefficient (DSC), can be employed to partly address the problem by turning pixel-wise labeling problem into minimizing class-level distribution distance \cite{salehi2017tversky}. 

Several methods have been proposed to alleviate the small-volumed organ segmentation problem. The generalized dice loss uses squared volume weights. However, it makes the optimization unstable in the extremely unbalanced segmentation \cite{sudre2017generalised}. The exponential logarithmic loss \cite{tang18miccai} is inspired by the focal loss \cite{lin2017focal} for class-level loss as $\mathbb{E}[{(-\ln (D))}^{\gamma}]$, where $D$ is the dice coefficient (DSC) for the interested class, $\gamma$ can be set as 0.3, and $\mathbb{E}$ is the expectation over classes and whole-volume CT images. The gradient of exponential logarithmic loss w.r.t. DSC $D$ is $-\frac{0.3}{D {\ln (D)}^{0.7}}$. The absolute value of gradient is getting bigger for well-segmented class ($D$ close to 1).  Therefore, the exponential logarithmic loss still places more weights on well-segmented class, and is not effective in learning to improve on not-well-segmented class.

In the AnatomyNet, we employ a hybrid loss consisting of contributions from both dice loss and focal loss \cite{lin2017focal}. The dice loss learns the class distribution alleviating the imbalanced voxel problem, where as the focal loss forces the model to learn poorly classified voxels better. The total loss can be formulated as
\begin{equation}
\label{5eq:hybridloss}
\begin{aligned}
{TP}_p(c) =& \sum_{n=1}^{N}{p_{n}(c)g_{n}(c)} \\
{FN}_p(c) =& \sum_{n=1}^{N}{(1-p_{n}(c))g_{n}(c)} \\
{FP}_p(c) =& \sum_{n=1}^{N}{p_{n}(c)(1-g_{n}(c))} \\
\mathcal{L} =& \mathcal{L}_{Dice} + \lambda \mathcal{L}_{Focal} \\
=&C - \sum_{c=0}^{C-1}{\frac{{TP}_p(c)}{{TP}_p(c)
		+ \alpha {FN}_p(c) + \beta {FP}_p(c)}} \\
&- \lambda \frac{1}{N}\sum_{c=0}^{C-1}\sum_{n=1}^{N}g_n(c){(1-p_n(c))}^{2}\log(p_n(c)) \, ,
\end{aligned}
\end{equation}
where ${TP}_p(c)$, ${FN}_p(c)$ and ${FP}_p(c)$ are the true positives, false negatives and false positives for class $c$ calculated by prediction probabilities respectively, $p_n(c)$ is the predicted probability for voxel $n$ being class $c$, $g_n(c)$ is the ground truth for voxel $n$ being class $c$, $C$ is the total number of anatomies plus one (background), $\lambda$ is the trade-off between dice loss $\mathcal{L}_{Dice}$ and focal loss $\mathcal{L}_{Focal}$, $\alpha$ and $\beta$ are the trade-offs of penalties for false negatives and false positives which are set as 0.5 here, $N$ is the total number of voxels in the CT images. $\lambda$ is set to be 0.1, 0.5 or 1 based on the performance on the validation set.

\subsection{Handling missing annotations}\label{5sec:missingannot}
Another challenge in anatomical segmentation is due to missing annotations common in the training datasets, because annotators often include different anatomies in their annotations. For example, we collect 261 head and neck CT images with anatomical segmentation ground truth from 5 hospitals, and the numbers of nine annotated anatomies are very different as shown in Table \ref{5tab:numberanatomy}. To handle this challenge, we mask out the background (denoted as class 0) and the missed anatomy. Let $c \in \{1,2,3,4,5,6,7,8,9\}$ denote the index of anatomies. We employ a mask vector $\bm{m}_i$ for the $i$th CT image, and denote background as label $0$. That is 
$\bm{m}_i(c) = 1$ if anatomy c is annotated, and $0$ otherwise. For the background, the mask is
$\bm{m}_i(0) = 1$ if all anatomies are annotated, and 0 otherwise.
\begin{table}[ht]
	\caption{The numbers of the nine annotated anatomies from 261 head and neck training CT images.} 
	\label{5tab:numberanatomy}
	\begin{center}       
		\begin{tabular}{|c|c|} 
			\hline
			\rule[-1ex]{0pt}{3.5ex}  Anatomy names &\#Annotations\\
			\hline
			\rule[-1ex]{0pt}{3.5ex}  Brain Stem &196\\
			\hline
			\rule[-1ex]{0pt}{3.5ex} Chiasm &129 \\
			\hline
			\rule[-1ex]{0pt}{3.5ex} Mandible &227\\
			\hline
			\rule[-1ex]{0pt}{3.5ex} Opt Ner L &133 \\
			\hline
			\rule[-1ex]{0pt}{3.5ex} Opt Ner R &133 \\
			\hline
			\rule[-1ex]{0pt}{3.5ex} Parotid L &257 \\
			\hline
			\rule[-1ex]{0pt}{3.5ex} Parotid R &256\\
			\hline
			\rule[-1ex]{0pt}{3.5ex} Submand. L &135\\
			\hline
			\rule[-1ex]{0pt}{3.5ex} Submand. R &130 \\
			\hline
		\end{tabular}
	\end{center}
\end{table}
The missing annotations for some anatomical structures cause imbalanced class-level annotations. To address this problem, we employ weighted loss function for balanced weights updating of different anatomies. The weights $\bm{w}$ are set as the inverse of the number of annotations for class $c$, $\bm{w}(c)=1/\sum_{i}\bm{m}_i(c)$, so that the weights in deep networks are updated equally with different anatomies. The dice loss for $i$th CT image in equation \ref{5eq:hybridloss} can be written as 
\begin{equation}
\label{5eq:missing}
\mathcal{\tilde{L}}_{Dice} =C - \sum_{c=0}^{C-1} \bm{m}_i(c) \bm{w}(c) {\frac{{TP}_p(c)}{{TP}_p(c)
		+ \alpha {FN}_p(c) + \beta {FP}_p(c)}}  \, .
\end{equation}
The focal loss for missing annotations in the $i$th CT image can be written as 
\begin{equation}
\label{5eq:focalmissing}
\mathcal{\tilde{L}}_{Focal}
=- \frac{1}{N}\sum_{c=0}^{C-1}\bm{m}_i(c) \bm{w}(c)\sum_{n=1}^{N}g_n(c){(1-p_n(c))}^{2}\log(p_n(c)) \, .
\end{equation}
We use loss $\mathcal{\tilde{L}}_{Dice} + \lambda \mathcal{\tilde{L}}_{Focal}$ in the AnatomyNet.

\subsection{Implementation details and performance evaluation}\label{5method_eval}
We implemented AnatomyNet in PyTorch, and trained it on NVIDIA Tesla P40. Batch size was set to be 1 because of different sizes of whole-volume CT images. We first used RMSprop optimizer \cite{tieleman2012lecture} with learning rate being $0.002$ and the number of epochs being 150. Then we used stochastic gradient descend with momentum 0.9, learning rate $0.001$ and the number of epochs 50. Because of size differences for different HaN whole-volume CT images, we set the batch size to be 1. During training, we used affine transformation and elastic deformation for data augmentation, implemented on the fly. 

We use Dice coefficient (DSC) as the final evaluation metric, defined to be  $2TP/(2TP+FN+FP)$, where $TP$, $FN$, and $FP$ are true positives, false negatives, false positives, respectively. 

\section{Results}\label{5sec:result}
We trained our deep learning model, AnatomyNet, on 261 training samples, and evaluated its performance on the MICCAI head and neck segmentation challenge 2015 test data (10 samples, DATASET 4) and compared it to the performances of previous methods benchmarked on the same test dataset.  Before we present the final results, we first describe the rationale behind several design choices under AnatomyNet, including architectural designs and model training. 
\subsection{Determining down-sampling scheme}
The standard U-Net model has multiple down-sampling layers, which help the model learn high-level image features.  However, down-sampling also reduces image resolution and makes it harder to segment small OARs such as optic nerves and chiasm.  To evaluate the effect of the number of down-sampling layers on the segmentation performance, we experimented with four different down-sampling schemes shown in Table \ref{5tab:netstructpool}.  Pool 1 uses only one down-sampling step, while Pool 2, 3, and 4 use 2, 3 and 4 down-sampling steps, respectively, distributed over consecutive blocks. With each down-sampling, the feature map size is reduced by half.  We incorporated each of the four down-sampling schemes into the standard U-Net model, which was then trained on the training set and evaluated on the test set.  For fair comparisons, we used the same number of filters in each layer. The decoder layers of each model are set to be symmetric with the encoder layers.

\begin{table}[ht]
	\caption{Sizes of encoder blocks in U-Nets with different numbers of down-samplings. The number before symbol @ denotes the number of output channels, while the number after the symbol denotes the size of feature map relative to the input.} 
	\label{5tab:netstructpool}
	\begin{center}       
		\begin{tabular}{|c|c|c|c|c|} 
			\hline
			\rule[-1ex]{0pt}{3.5ex}  Nets &1st block & 2nd block & 3rd block & 4th block\\
			\hline
			\rule[-1ex]{0pt}{3.5ex}  Pool 1 &$32@{(1/2)}^3$&$40@{(1/2)}^3$&$48@{(1/2)}^3$&$56@{(1/2)}^3$\\
			\hline
			\rule[-1ex]{0pt}{3.5ex} Pool 2 &$32@{(1/2)}^3$&$40@{(1/4)}^3$&$48@{(1/4)}^3$&$56@{(1/4)}^3$\\
			\hline
			\rule[-1ex]{0pt}{3.5ex} Pool 3 &$32@{(1/2)}^3$&$40@{(1/4)}^3$&$48@{(1/8)}^3$&$56@{(1/8)}^3$\\
			\hline
			\rule[-1ex]{0pt}{3.5ex} Pool 4 &$32@{(1/2)}^3$&$40@{(1/4)}^3$&$48@{(1/8)}^3$&$56@{(1/16)}^3$\\
			\hline
		\end{tabular}
	\end{center}
\end{table}

The DSC scores of the four down-sampling schemes are shown in Table \ref{5tab:numpool}. On average, one down-sampling block (Pool 1) yields the best average performance, beating other down-sampling schemes in 6 out of 9 anatomies. The performance gaps are most prominent on three small-volumed OARs -  optic nerve left, optic nerve right and optic chiasm, which demonstrates that the U-Net with one down-sampling layer works better on small organ segmentation than the standard U-Net. The probable reason is that small organs reside in only a few slices and more down-sampling layers are more likely to miss features for the small organs in the deeper layers. Based on these results, we decide to use only one down-sampling layer in AnatomyNet  (Fig. \ref{5fig:seunet}). 

	\begin{table}[ht]
		\caption{Performances of U-Net models with different numbers of down-sampling layers, measured with Dice coefficients.} 
		\label{5tab:numpool}
		\begin{center}    
				\begin{tabular}{|c|c|c|c|c|} 
					\hline
					\rule[-1ex]{0pt}{3.5ex}  Anatomy &Pool 1 &Pool 2 & Pool 3 & Pool 4 \\
					\hline
					\rule[-1ex]{0pt}{3.5ex}  Brain Stem&85.1&\textbf{85.3}&84.3&84.9\\
					\hline
					\rule[-1ex]{0pt}{3.5ex} Chiasm &\textbf{50.1}&48.7&47.0&45.3\\
					\hline
					\rule[-1ex]{0pt}{3.5ex} Mand. &\textbf{91.5}&89.9&90.6&90.1\\
					\hline
					\rule[-1ex]{0pt}{3.5ex} Optic Ner L &\textbf{69.1}&65.7&67.2&67.9\\
					\hline
					\rule[-1ex]{0pt}{3.5ex} Optic Ner R &\textbf{66.9}&65.0&66.2&63.7\\
					\hline
					\rule[-1ex]{0pt}{3.5ex} Paro. L &\textbf{86.6}&\textbf{86.6}&85.9&\textbf{86.6}\\
					\hline
					\rule[-1ex]{0pt}{3.5ex} Paro. R &85.6&85.3&84.8&\textbf{85.9}\\
					\hline
					\rule[-1ex]{0pt}{3.5ex} Subm. L &\textbf{78.5}&77.9&77.9&77.3\\
					\hline
					\rule[-1ex]{0pt}{3.5ex} Subm. R &77.7&\textbf{78.4}&76.6&77.8\\
					\hline
					\rule[-1ex]{0pt}{3.5ex} Average &\textbf{76.8}&75.9&75.6&75.5\\
					\hline
				\end{tabular}
		\end{center}
\end{table}

\subsection{Choosing network structures}\label{5sec:expnetstruct}
In addition to down-sample schemes, we also tested several other architecture designing choices. The first one is on how to combine features from horizontal layers within U-Net. Traditional U-Net uses concatenation to combine features from horizontal layers in the decoder, as illustrated with dash lines in Fig. \ref{5fig:seunet}. However, recent feature pyramid network (FPN) recommends summation to combine horizontal features \cite{lin2017feature}.   Another designing choice is on choosing local feature learning blocks with each layer.  The traditional U-Net uses simple 2D convolution, extended to 3D convolution in our case.  To learn more effective features, we tried two other feature learning blocks: a) residual learning, and b) squeeze-and-excitation residual learning.  Altogether, we investigated the performances of the following six architectural designing choices:
\begin{enumerate}
\item \textbf{3D SE Res UNet}, the architecture implemented in AnatomyNet (Fig.\ \ref{5fig:seunet}) with both squeeze-excitation residual learning and concatenated horizontal features. 
\item \textbf{3D Res UNet}, replacing the SE Residual blocks in 3D SE Res UNet with residual blocks.
\item \textbf{Vanilla U-Net}, replacing the SE Residual blocks in 3D SE Res UNet with 3D convolutional layers.
\item \textbf{3D SE Res UNet (sum)}, replacing concatenations in 3D SE Res UNet with summations. When the numbers of channels are different, one additional $1\times 1 \times 1$ 3D convolutional layer is used to map the encoder to the same size as the decoder.
\item \textbf{3D Res UNet (sum)}, replacing the SE Residual blocks in 3D SE Res UNet (sum) with residual blocks.
\item \textbf{Vanilla U-Net (sum)}, replacing the SE Residual blocks in 3D SE Res UNet (sum) with 3D convolutional layers.
\end{enumerate}

The six models were trained on the same training dataset with identical training procedures. The performances measured by DSC on the  test dataset are summarized in Table \ref{5tab:seres_sum}. We notice a few observations from this study. First, feature concatenation shows consistently better performance than feature summation. It seems feature concatenation provides more flexibility in feature learning than the fixed operation through feature summation. Second, 3D SE residual U-Net with concatenation yields the best performance. It demonstrates the power of SE features on 3D semantic segmentation, because the SE scheme learns the channel-wise calibration and helps alleviate the dependencies among channel-wise features as discussed in Section \ref{5sec:anatomynet}. 

The SE residual block learning incorporated in AnatomyNet results in 2-3\%  improvements in DSC over the traditional U-Net model, outperforming U-Net in 6 out of 9 anatomies. 
	\begin{table}[ht]
		\caption{Performance comparison on different network structures} 
		\label{5tab:seres_sum}
		\begin{center}    
			\resizebox{\columnwidth}{!}{   
				\begin{tabular}{|c|c|c|c|c|c|c|} 
					\hline
					\rule[-1ex]{0pt}{3.5ex}  Anatomy &\begin{tabular}{@{}c@{}}Vanilla \\ UNet\end{tabular} & \begin{tabular}{@{}c@{}}3D Res\\ UNet\end{tabular} & \begin{tabular}{@{}c@{}}3D SE Res \\ UNet\end{tabular} & \begin{tabular}{@{}c@{}}Vanilla \\ UNet (sum)\end{tabular} & \begin{tabular}{@{}c@{}}3D Res \\ UNet (sum)\end{tabular}&\begin{tabular}{@{}c@{}}3D SE Res \\ UNet (sum)\end{tabular}\\
					\hline
					\rule[-1ex]{0pt}{3.5ex}  \begin{tabular}{@{}c@{}}Brain \\ Stem\end{tabular}&85.1&85.9&\textbf{86.4}&85.0&85.8&86.0\\
					\hline
					\rule[-1ex]{0pt}{3.5ex} Chiasm &50.1&53.3&53.2&50.8&49.8&\textbf{53.5}\\
					\hline
					\rule[-1ex]{0pt}{3.5ex} Mand. &91.5&90.6&\textbf{92.3}&90.2&90.9&91.3\\
					\hline
					\rule[-1ex]{0pt}{3.5ex} \begin{tabular}{@{}c@{}}Optic \\ Ner L\end{tabular} &69.1&69.8&\textbf{72.0}&66.5&70.4&68.9\\
					\hline
					\rule[-1ex]{0pt}{3.5ex} \begin{tabular}{@{}c@{}}Optic \\ Ner R\end{tabular} &66.9&67.5&\textbf{69.1}&66.1&67.4&45.6\\
					\hline
					\rule[-1ex]{0pt}{3.5ex} Paro. L &86.6&\textbf{87.8}&\textbf{87.8}&86.8&87.5&87.4\\
					\hline
					\rule[-1ex]{0pt}{3.5ex} Paro. R &85.6&86.2&86.8&86.0&86.7&\textbf{86.9}\\
					\hline
					\rule[-1ex]{0pt}{3.5ex} Subm. L &78.5&79.9&81.1&79.3&79.1&\textbf{82.4}\\
					\hline
					\rule[-1ex]{0pt}{3.5ex} Subm. R &77.7&80.2&\textbf{80.8}&77.6&78.4&80.5\\
					\hline
					\rule[-1ex]{0pt}{3.5ex} Average &76.8&77.9&\textbf{78.8}&76.5&77.3&75.8\\
					\hline
				\end{tabular}
			}
		\end{center}
\end{table}

\subsection{Choosing loss functions}
We also validated the effects of different loss functions on training and model performance. To differentiate the effects of loss functions from network design choices,  we used only the vanilla U-Net and trained it with different loss functions. This way, we can focus on studying the impact of loss functions on model performances.  We tried four loss functions, including Dice loss, exponential logarithmic loss, hybrid loss between Dice loss and focal loss, and hybrid loss between Dice loss and cross entropy. The trade-off parameter in hybrid losses ($\lambda$ in Eq.\ \ref{5eq:hybridloss}) was chosen from either 0.1, 0.5 or 1, based on the performance on a validation set. For hybrid loss between Dice loss and focal loss, the best $\lambda$ was found to be 0.5. For hybrid loss between Dice loss and cross entropy, the best $\lambda$ was 0.1.


	\begin{table}[ht]
		\caption{Comparisons of test performances of models trained with different loss functions, evaluated with Dice coefficients.} 
		\label{5tab:lossfunc}
		\begin{center}    
				\begin{tabular}{|c|c|c|c|c|} 
					\hline
					\rule[-1ex]{0pt}{3.5ex}  Anatomy &\begin{tabular}{@{}c@{}}Dice \\ loss\end{tabular} & \begin{tabular}{@{}c@{}}Exp. Log.\\ Dice \end{tabular} & \begin{tabular}{@{}c@{}}Dice + \\ focal \end{tabular} & \begin{tabular}{@{}c@{}}Dice + cross \\ entropy \end{tabular}\\
					\hline
					\rule[-1ex]{0pt}{3.5ex}  Brain Stem&85.1&85.0&\textbf{86.1}&85.2\\
					\hline
					\rule[-1ex]{0pt}{3.5ex} Chiasm &50.1&50.0&\textbf{52.2}&48.8\\
					\hline
					\rule[-1ex]{0pt}{3.5ex} Mand. &\textbf{91.5}&89.9&90.0&91.0\\
					\hline
					\rule[-1ex]{0pt}{3.5ex} Optic Ner L &69.1&67.9&68.4&\textbf{69.6}\\
					\hline
					\rule[-1ex]{0pt}{3.5ex} Optic Ner R &66.9&65.9&\textbf{69.1}&67.4\\
					\hline
					\rule[-1ex]{0pt}{3.5ex} Paro. L &86.6&86.4&87.4&\textbf{88.0}\\
					\hline
					\rule[-1ex]{0pt}{3.5ex} Paro. R &85.6&84.8&86.3&\textbf{86.9}\\
					\hline
					\rule[-1ex]{0pt}{3.5ex} Subm. L &78.5&76.3&\textbf{79.6}&77.8\\
					\hline
					\rule[-1ex]{0pt}{3.5ex} Subm. R &77.7&78.2&\textbf{79.8}&78.4\\
					\hline
					\rule[-1ex]{0pt}{3.5ex} Average &76.8&76.0&\textbf{77.7}&77.0\\
					\hline
				\end{tabular}
		\end{center}
\end{table}

The performance of the model trained with the four loss functions described above are shown in Table \ref{5tab:lossfunc}. The performance is measured in terms of the average DSC on the test dataset. We notice a few observations from this experiment. First, the two hybrid loss functions consistently outperform simple Dice or exponential logarithmic loss, beating the other two losses in 8 out of 9 anatomies. This suggests that taking the voxel-level loss into account can improve performance. Second, between the two hybrid losses, Dice combined with focal loss has better performances.  In particular, it leads to significant improvements (2-3\%) on segmenting two small anatomies - optic nerve R and optic chiasm, consistent with our motivation discussed in the Section \ref{5sec:extremeimbalance}.  

Based on the above observations, the hybrid loss with Dice combined with focal loss was used to train AnatomyNet, and benchmark its performance against previous methods.

\subsection{Comparing to state-of-the-art methods}
After having determined the structure of AnatomyNet and the loss function for training it, we set out to compare its performance with previous state-of-the-art methods. For consistency purpose, all models were evaluated on the MICCAI head and neck challenge 2015 test set. The average DSC of different methods are summarized in Table \ref{5tab:anatomynet}. The best result for each anatomy from the MICCAI 2015 challenge is denoted as MICCAI 2015 \cite{raudaschl2017evaluation}, which may come from different teams with different methods. 
\begin{table}[ht]
	\caption{Performance comparisons with state-of-the-art methods, showing the average DSC on the test set.} 
	\label{5tab:anatomynet}
	\begin{center}    
		\resizebox{\columnwidth}{!}{   
		\begin{tabular}{|c|c|c|c|c|c|} 
			\hline
			\rule[-1ex]{0pt}{3.5ex}  Anatomy &\begin{tabular}{@{}c@{}}\textbf{MICCAI} \\ \textbf{2015} \cite{raudaschl2017evaluation}\end{tabular} & \begin{tabular}{@{}c@{}}Fritscher et al. \\ 2016 \cite{fritscher2016deep}\end{tabular} & \begin{tabular}{@{}c@{}}Ren et al. \\ 2018 \cite{ren2018interleaved}\end{tabular} & \begin{tabular}{@{}c@{}}Wang et al. \\ 2018 \cite{wang2018hierarchical}\end{tabular} & AnatomyNet\\
			\hline
			\rule[-1ex]{0pt}{3.5ex}  \begin{tabular}{@{}c@{}}Brain \\ Stem\end{tabular}&88&N/A&N/A&\textbf{90$\pm$4}&86.65$\pm$2\\
			\hline
			\rule[-1ex]{0pt}{3.5ex} Chiasm &55&49$\pm$9&\textbf{58$\pm$17}&N/A&53.22$\pm$15\\
			\hline
			\rule[-1ex]{0pt}{3.5ex} Mand. &93&N/A&N/A&\textbf{94}$\pm$1&92.51$\pm$2\\
			\hline
			\rule[-1ex]{0pt}{3.5ex} \begin{tabular}{@{}c@{}}Optic \\ Ner L\end{tabular} &62&N/A&72$\pm$8&N/A&\textbf{72.10$\pm$6}\\
			\hline
			\rule[-1ex]{0pt}{3.5ex} \begin{tabular}{@{}c@{}}Optic \\ Ner R\end{tabular} &62&N/A&70$\pm$9&N/A&\textbf{70.64$\pm$10}\\
			\hline
			\rule[-1ex]{0pt}{3.5ex} Paro. L &84&81$\pm$4&N/A&83$\pm$6&\textbf{88.07$\pm$2}\\
			\hline
			\rule[-1ex]{0pt}{3.5ex} Paro. R &84&81$\pm$4&N/A&83$\pm$6&\textbf{87.35$\pm$4}\\
			\hline
			\rule[-1ex]{0pt}{3.5ex} Subm. L &78&65$\pm$8&N/A&N/A&\textbf{81.37$\pm$4}\\
			\hline
			\rule[-1ex]{0pt}{3.5ex} Subm. R &78&65$\pm$8&N/A&N/A&\textbf{81.30$\pm$4}\\
			\hline
			\rule[-1ex]{0pt}{3.5ex} Average &76&N/A&N/A&N/A&\textbf{79.25}\\
			\hline
		\end{tabular}
	}
	\end{center}
\end{table}

MICCAI 2015 competition merged left and right paired organs into one target, while we treat them as two separate anatomies.  As a result, MICCAI 2015 competition is a seven (6 organs + background) class segmentation and ours is a ten-class segmentation, which makes the segmentation task more challenging. Nonetheless, the AnatomyNet achieves an average Dice coefficient of 79.25, which is 3.3\% better than the best result from MICCAI 2015 Challenge (Table \ref{5tab:anatomynet}). In particular, the improvements on optic nerves are about 9-10\%, suggesting that deep learning models are better equipped to handle small anatomies with large variations among patients.   The AnatomyNet also outperforms the atlas based ConvNets in \cite{fritscher2016deep} on all classes, which is likely contributed by the fact that the end-to-end structure in AnatomyNet for whole-volume HaN CT image captures global information for relative spatial locations among anatomies. Compared to the interleaved ConvNets in \cite{ren2018interleaved} on small-volumed organs, such as chiasm, optic nerve left and optic nerve right, AnatomyNet is better on 2 out of 3 cases. The interleaved ConvNets achieved higher performance on chiasm, which is likely contributed by the fact that its prediction was operated on small region of interest (ROI), obtained first through atlas registration, while AnatomyNet operates directly on whole-volume slices. 

Aside from the improvement on segmentation accuracy, another advantage of AnatomyNet is that it is orders of magnitude faster than traditional atlas-based methods using in the MICCAI 2015 challenge.  AnatomyNet takes about 0.12 seconds to fully segment  a head and neck CT image of  dimension  $178 \times 302 \times 225$. By contrast, the atlas-based methods can take a dozen minutes to complete one segmentation depending on implementation details and the choices on the number of atlases. 


\subsection{Visualizations on MICCAI 2015 test}
\begin{figure} [ht]
	\begin{center}
		 		\begin{minipage}{0.15\linewidth}
		 			\includegraphics[width=\textwidth]{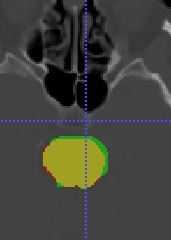}
		 		\end{minipage}
	 		\begin{minipage}{0.15\linewidth}
	 			\includegraphics[width=\textwidth]{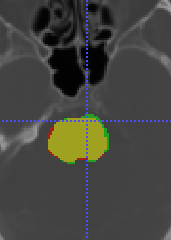}
	 		\end{minipage}
 		\begin{minipage}{0.15\linewidth}
 			\includegraphics[width=\textwidth]{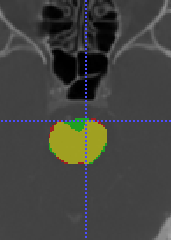}
 		\end{minipage}
 		\begin{minipage}{0.15\linewidth}
 			\includegraphics[width=\textwidth]{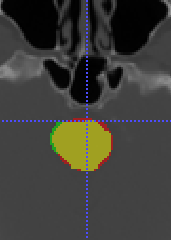}
 		\end{minipage} \\
 		
		 		\begin{minipage}{0.15\linewidth}
		 			\includegraphics[width=\textwidth]{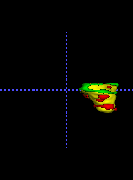}
		 		\end{minipage}
	 		\begin{minipage}{0.15\linewidth}
	 			\includegraphics[width=\textwidth]{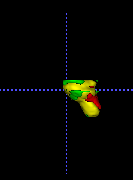}
	 		\end{minipage}
 		\begin{minipage}{0.15\linewidth}
 			\includegraphics[width=\textwidth]{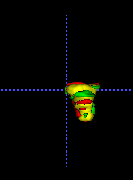}
 		\end{minipage}
 	\begin{minipage}{0.15\linewidth}
 		\includegraphics[width=\textwidth]{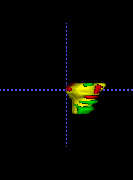}
 	\end{minipage} \\
 	
		 		\begin{minipage}{0.15\linewidth}
		 			\includegraphics[width=\textwidth]{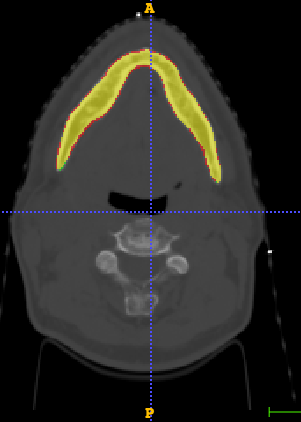}
		 		\end{minipage}
	 		\begin{minipage}{0.15\linewidth}
	 			\includegraphics[width=\textwidth]{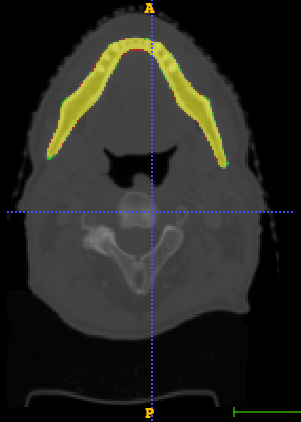}
	 		\end{minipage}
 		\begin{minipage}{0.15\linewidth}
 			\includegraphics[width=\textwidth]{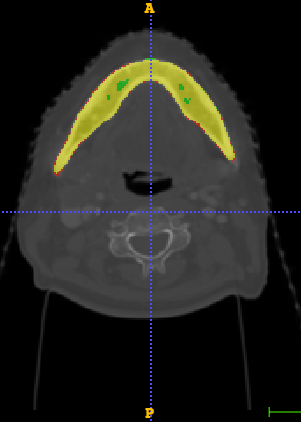}
 		\end{minipage}
 	\begin{minipage}{0.15\linewidth}
 		\includegraphics[width=\textwidth]{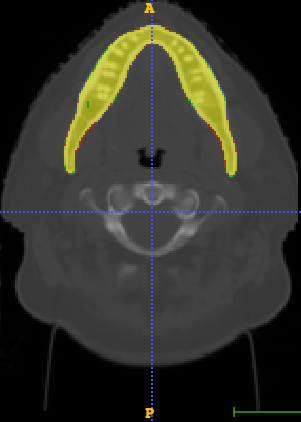}
 	\end{minipage}\\
 
		 		\begin{minipage}{0.15\linewidth}
		 			\includegraphics[width=\textwidth]{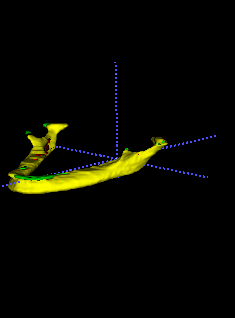}
		 		\end{minipage}
	 		\begin{minipage}{0.15\linewidth}
	 			\includegraphics[width=\textwidth]{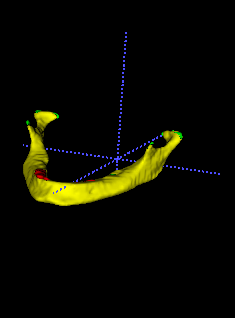}
	 		\end{minipage}
 		\begin{minipage}{0.15\linewidth}
 			\includegraphics[width=\textwidth]{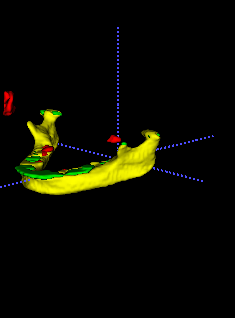}
 		\end{minipage}
 	\begin{minipage}{0.15\linewidth}
 		\includegraphics[width=\textwidth]{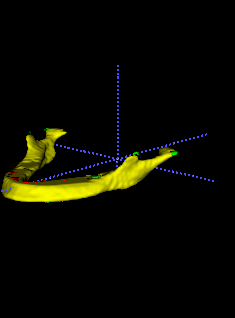}
 	\end{minipage} \\
 
		 		\begin{minipage}{0.15\linewidth}
		 			\includegraphics[width=\textwidth]{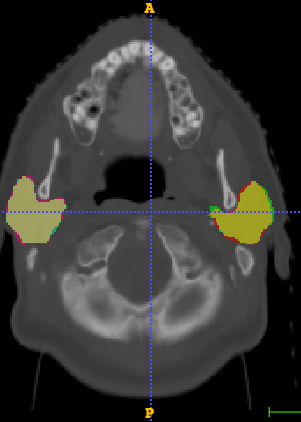}
		 		\end{minipage}
	 		\begin{minipage}{0.15\linewidth}
	 			\includegraphics[width=\textwidth]{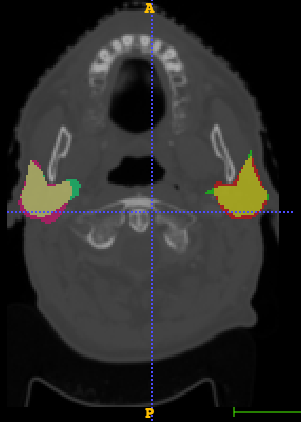}
	 		\end{minipage}
 		\begin{minipage}{0.15\linewidth}
 			\includegraphics[width=\textwidth]{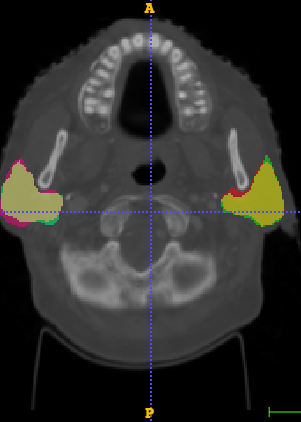}
 		\end{minipage}
 	\begin{minipage}{0.15\linewidth}
 		\includegraphics[width=\textwidth]{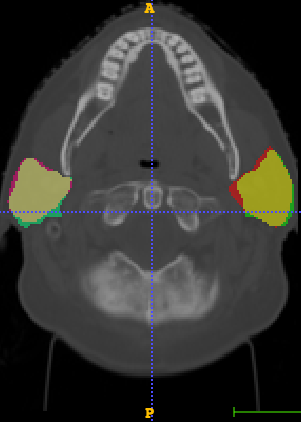}
 	\end{minipage} \\
 
		 		\begin{minipage}{0.15\linewidth}
		 			\includegraphics[width=\textwidth]{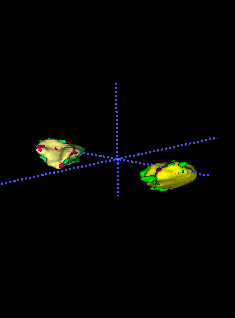}
		 		\end{minipage}
	 		\begin{minipage}{0.15\linewidth}
	 			\includegraphics[width=\textwidth]{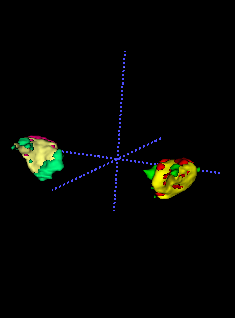}
	 		\end{minipage}
 		\begin{minipage}{0.15\linewidth}
 			\includegraphics[width=\textwidth]{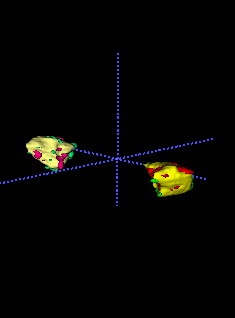}
 		\end{minipage}
 	\begin{minipage}{0.15\linewidth}
 		\includegraphics[width=\textwidth]{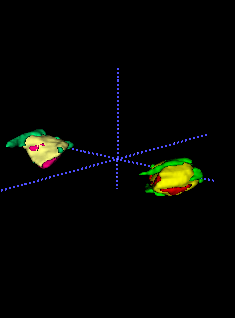}
 	\end{minipage} 
	\end{center}
	\caption{ \label{5fig:vis1} 
	Visualizations on four test CT images. Rows from top to bottom represent brain stem, brain stem 3D, mandibular, mandibular 3D, parotid left and right, and parotid left and right 3D. Each column represents one CT image. Green represents ground truth, and red denotes predictions. Yellow is the overlap. } 
\end{figure}

\begin{figure} [ht]
	\begin{center}
				 		\begin{minipage}{0.15\linewidth}
			\includegraphics[width=\textwidth]{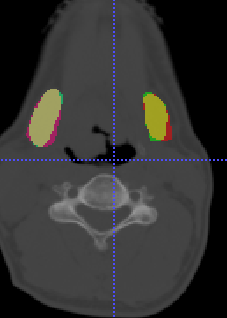}
		\end{minipage}
		\begin{minipage}{0.15\linewidth}
			\includegraphics[width=\textwidth]{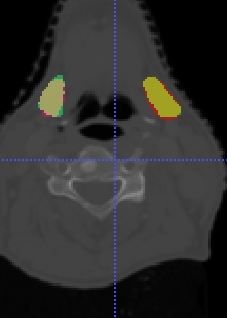}
		\end{minipage}
		\begin{minipage}{0.15\linewidth}
			\includegraphics[width=\textwidth]{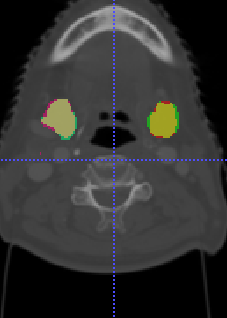}
		\end{minipage}
		\begin{minipage}{0.15\linewidth}
			\includegraphics[width=\textwidth]{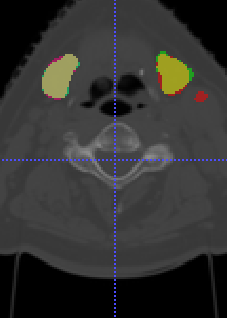}
		\end{minipage}\\
		
		\begin{minipage}{0.15\linewidth}
			\includegraphics[width=\textwidth]{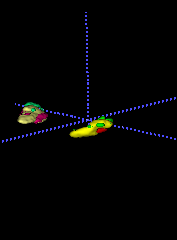}
		\end{minipage}
		\begin{minipage}{0.15\linewidth}
			\includegraphics[width=\textwidth]{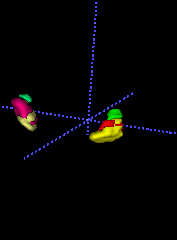}
		\end{minipage}
		\begin{minipage}{0.15\linewidth}
			\includegraphics[width=\textwidth]{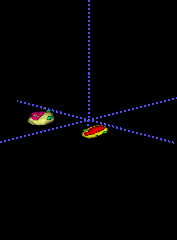}
		\end{minipage}
		\begin{minipage}{0.15\linewidth}
			\includegraphics[width=\textwidth]{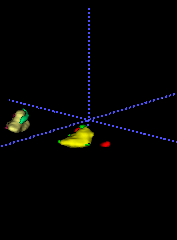}
		\end{minipage}  \\
	
		\begin{minipage}{0.15\linewidth}
			\includegraphics[width=\textwidth]{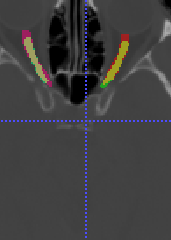}
		\end{minipage}
		\begin{minipage}{0.15\linewidth}
			\includegraphics[width=\textwidth]{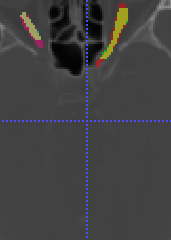}
		\end{minipage}
		\begin{minipage}{0.15\linewidth}
			\includegraphics[width=\textwidth]{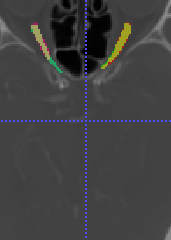}
		\end{minipage}
		\begin{minipage}{0.15\linewidth}
			\includegraphics[width=\textwidth]{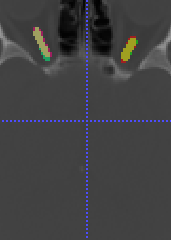}
		\end{minipage}\\
		
		\begin{minipage}{0.15\linewidth}
			\includegraphics[width=\textwidth]{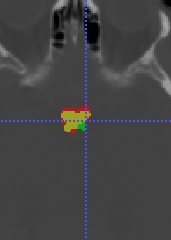}
		\end{minipage}
		\begin{minipage}{0.15\linewidth}
			\includegraphics[width=\textwidth]{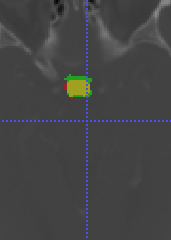}
		\end{minipage}
		\begin{minipage}{0.15\linewidth}
			\includegraphics[width=\textwidth]{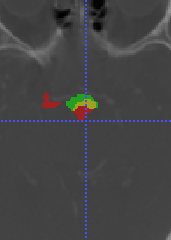}
		\end{minipage}
		\begin{minipage}{0.15\linewidth}
			\includegraphics[width=\textwidth]{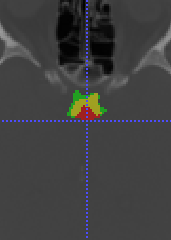}
		\end{minipage}
	\end{center}
	\caption{ \label{5fig:vis1_2} 
		Visualizations on four test CT images. Rows from top to bottom represent submandibular left and right, submandibular left and right 3D, optic nerve left and right, and chiasm. Each column represents one CT image. Green represents ground truth, and red denotes predictions. Yellow is the overlap. The AnatomyNet performs well on small-volumed anatomies.} 
\end{figure}

In Fig.\ \ref{5fig:vis1} and Fig.\ \ref{5fig:vis1_2}, we visualize the segmentation results by AnatomyNet on four cases from the test dataset.   Each row represents one (left and right) anatomy or 3D reconstructed anatomy. Each column denotes one sample. The last two columns show cases where AnatomyNet did not perform well.  The discussions of these cases are presented in Section \ref{5sec:futurework}. Green denotes the ground truth. Red represents predicted segmentation results. Yellow denotes the overlap between ground truth and prediction. We visualize the slices containing the largest area of each related organ.  For small OARs such as optic nerves and chiasm (shown in  Fig.\ \ref{5fig:vis1_2}), only cross-sectional slices are shown. 


\subsection{Visualizations on independent samples}\label{5sec:holdout}
\begin{figure} [ht]
	\begin{center}
		 		\begin{minipage}{0.15\linewidth}
		 			\includegraphics[width=\textwidth]{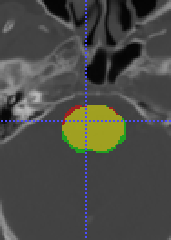}
		 		\end{minipage}
	 		\begin{minipage}{0.15\linewidth}
	 			\includegraphics[width=\textwidth]{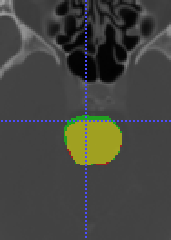}
	 		\end{minipage}
 		\begin{minipage}{0.15\linewidth}
 			\includegraphics[width=\textwidth]{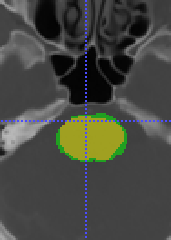}
 		\end{minipage}
 	\begin{minipage}{0.15\linewidth}
 		\includegraphics[width=\textwidth]{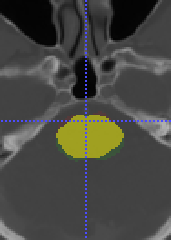}
 	\end{minipage}\\
 
		 		\begin{minipage}{0.15\linewidth}
		 			\includegraphics[width=\textwidth]{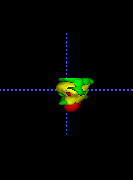}
		 		\end{minipage}
	 		\begin{minipage}{0.15\linewidth}
	 			\includegraphics[width=\textwidth]{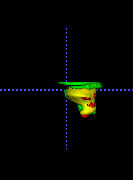}
	 		\end{minipage}
 		\begin{minipage}{0.15\linewidth}
 			\includegraphics[width=\textwidth]{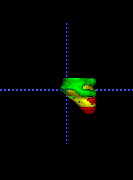}
 		\end{minipage}
 	\begin{minipage}{0.15\linewidth}
 		\includegraphics[width=\textwidth]{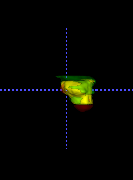}
 	\end{minipage}\\
 
		 		\begin{minipage}{0.15\linewidth}
		 			\includegraphics[width=\textwidth]{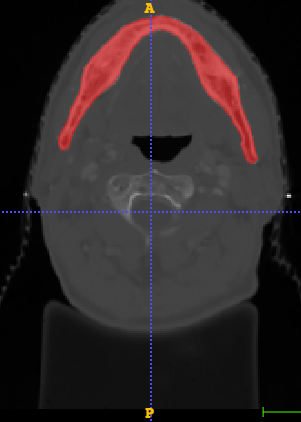}
		 		\end{minipage}
	 		\begin{minipage}{0.15\linewidth}
	 			\includegraphics[width=\textwidth]{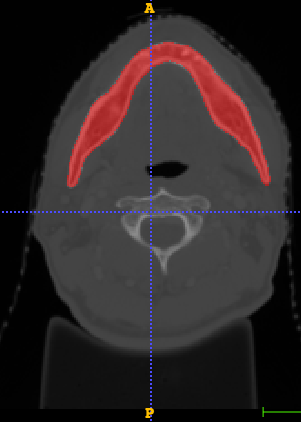}
	 		\end{minipage}
 		\begin{minipage}{0.15\linewidth}
 			\includegraphics[width=\textwidth]{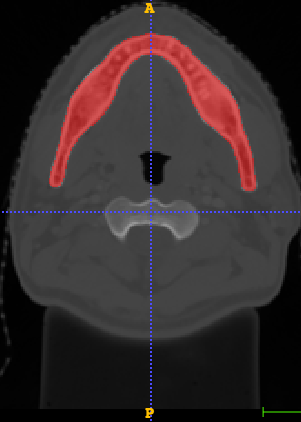}
 		\end{minipage}
 	\begin{minipage}{0.15\linewidth}
 		\includegraphics[width=\textwidth]{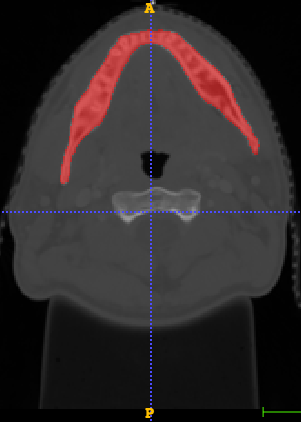}
 	\end{minipage}\\
 
		 		\begin{minipage}{0.15\linewidth}
		 			\includegraphics[width=\textwidth]{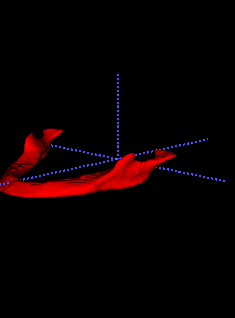}
		 		\end{minipage}
	 		\begin{minipage}{0.15\linewidth}
	 			\includegraphics[width=\textwidth]{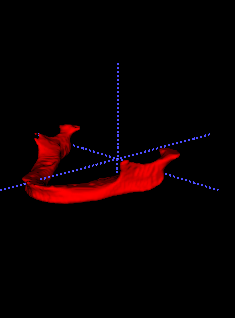}
	 		\end{minipage}
 		\begin{minipage}{0.15\linewidth}
 			\includegraphics[width=\textwidth]{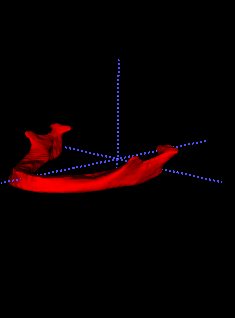}
 		\end{minipage}
 	\begin{minipage}{0.15\linewidth}
 		\includegraphics[width=\textwidth]{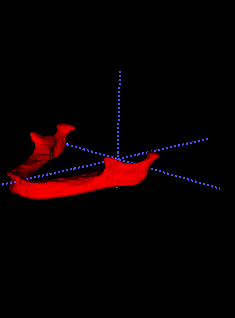}
 	\end{minipage}\\
 
		 		\begin{minipage}{0.15\linewidth}
		 			\includegraphics[width=\textwidth]{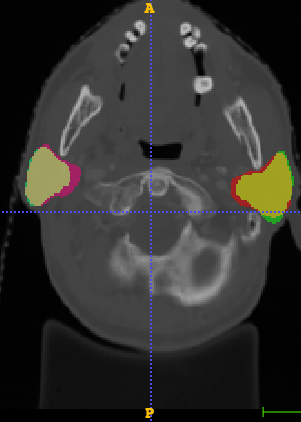}
		 		\end{minipage}
	 		\begin{minipage}{0.15\linewidth}
	 			\includegraphics[width=\textwidth]{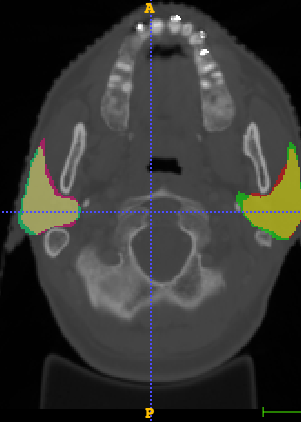}
	 		\end{minipage}
 		\begin{minipage}{0.15\linewidth}
 			\includegraphics[width=\textwidth]{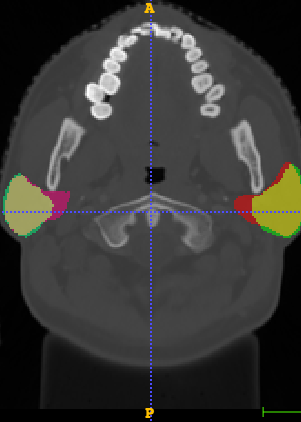}
 		\end{minipage}
 	\begin{minipage}{0.15\linewidth}
 		\includegraphics[width=\textwidth]{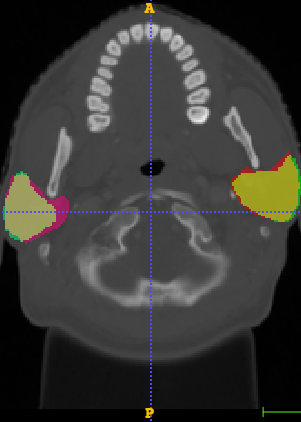}
 	\end{minipage}\\
 
		 		\begin{minipage}{0.15\linewidth}
		 			\includegraphics[width=\textwidth]{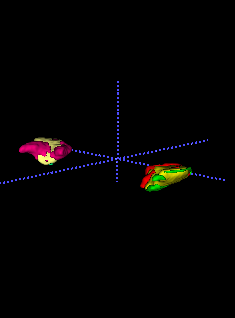}
		 		\end{minipage}
	 		\begin{minipage}{0.15\linewidth}
	 			\includegraphics[width=\textwidth]{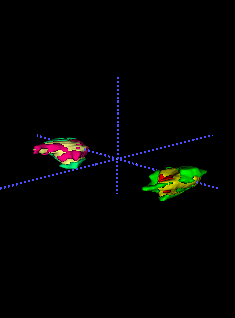}
	 		\end{minipage}
 		\begin{minipage}{0.15\linewidth}
 			\includegraphics[width=\textwidth]{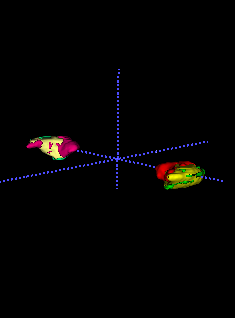}
 		\end{minipage}
 	\begin{minipage}{0.15\linewidth}
 		\includegraphics[width=\textwidth]{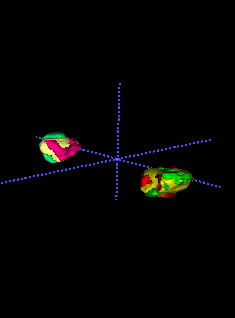}
 	\end{minipage}
	\end{center}
	\caption{ \label{5fig:vis2} 
	Visualizations for the first four anatomy on the first four holdout CT images. There is no ground truth for mandible and submandibular glands. Because this is a different source from MICCAI 2015, the annotations of brain stem and chiasm are inconsistent with those from MICCAI 2015. The AnatomyNet generalizes well for hold out test set.}
\end{figure}

\begin{figure} [ht]
	\begin{center}
		\begin{minipage}{0.15\linewidth}
			\includegraphics[width=\textwidth]{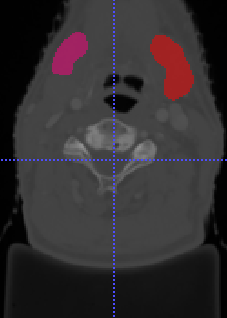}
		\end{minipage}
		\begin{minipage}{0.15\linewidth}
			\includegraphics[width=\textwidth]{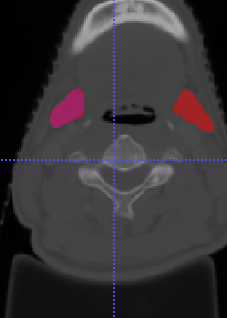}
		\end{minipage}
		\begin{minipage}{0.15\linewidth}
			\includegraphics[width=\textwidth]{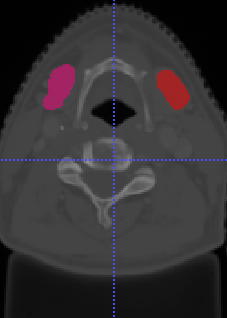}
		\end{minipage}
		\begin{minipage}{0.15\linewidth}
			\includegraphics[width=\textwidth]{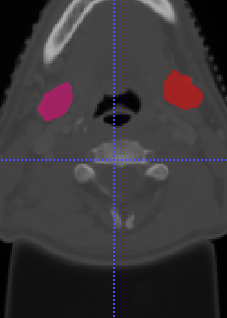}
		\end{minipage}\\
		
		\begin{minipage}{0.15\linewidth}
			\includegraphics[width=\textwidth]{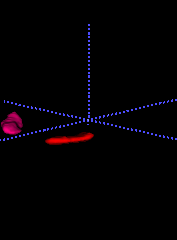}
		\end{minipage}
		\begin{minipage}{0.15\linewidth}
			\includegraphics[width=\textwidth]{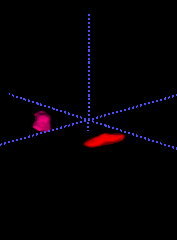}
		\end{minipage}
		\begin{minipage}{0.15\linewidth}
			\includegraphics[width=\textwidth]{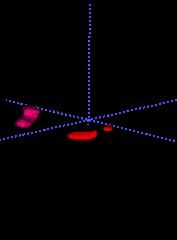}
		\end{minipage}
		\begin{minipage}{0.15\linewidth}
			\includegraphics[width=\textwidth]{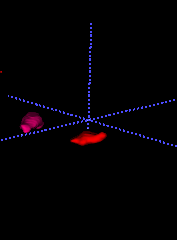}
		\end{minipage}\\
		
		\begin{minipage}{0.15\linewidth}
			\includegraphics[width=\textwidth]{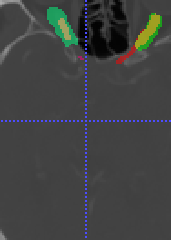}
		\end{minipage}
		\begin{minipage}{0.15\linewidth}
			\includegraphics[width=\textwidth]{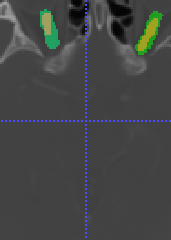}
		\end{minipage}
		\begin{minipage}{0.15\linewidth}
			\includegraphics[width=\textwidth]{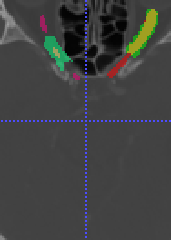}
		\end{minipage}
		\begin{minipage}{0.15\linewidth}
			\includegraphics[width=\textwidth]{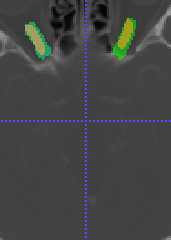}
		\end{minipage}\\
		
		\begin{minipage}{0.15\linewidth}
			\includegraphics[width=\textwidth]{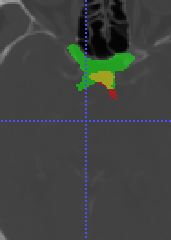}
		\end{minipage}
		\begin{minipage}{0.15\linewidth}
			\includegraphics[width=\textwidth]{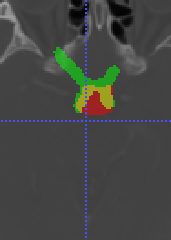}
		\end{minipage}
		\begin{minipage}{0.15\linewidth}
			\includegraphics[width=\textwidth]{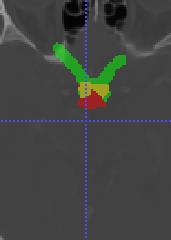}
		\end{minipage}
		\begin{minipage}{0.15\linewidth}
			\includegraphics[width=\textwidth]{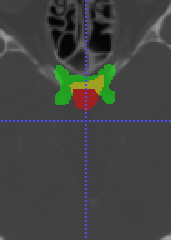}
		\end{minipage}
	\end{center}
	\caption{ \label{5fig:vis2_2} 
		{Visualizations for five anatomies on the first four holdout CT images. Rows from top to bottom represent submandibular left and right, submandibular left and right 3D, optic nerve left and right, and chiasm. The ground truth of holdout test only images is not consistent from the training set from the last row.}}
\end{figure}

To check the generalization ability of the trained model, we also visualize the segmentation results of the trained model on a small internal dataset in Fig. \ref{5fig:vis2} and Fig. \ref{5fig:vis2_2}. Visual inspection suggests that the trained model performed well on this independent test set. In general, the performances on larger anatomies are better than small ones (such as optic chiasm), which can be attributed by both manual annotation inconsistencies and algorithmic challenges in segmenting these small regions.

\section{Discussion}\label{5sec:discus}
\subsection{Impacts of training datasets}\label{5sec:datasetcomp} 
The training datasets we collected come from various sources with annotations done by different groups of physicians with different guiding criteria. It is unclear how the different datasets might contribute the model performance.  For this purpose, we carried out an experiment to test the model performance trained with two different datasets: a) using only the training data provided in the MICCAI head and neck segmentation challenge 2015 (DATASET 1, 38 samples), and b) the combined training data with 216 samples (DATASET 1-3 combined).  In terms of annotations, the first dataset is more consistent with the test dataset, therefore less likely to suffer from annotational inconsistencies.   However, on the other hand, the size of the dataset is much smaller, posing challenges to training deep learning models.


Table \ref{5tab:dataset} shows the test performances of a 3D Res U-Net model trained with the above-mentioned two datasets after applying the same training procedure of minimizing Dice loss. We notice a few observations. First, overall the model trained with the larger dataset (DATATSET 1-3) achieves better performance with a 2.5\% improvement over the smaller dataset, suggesting that the larger sample size does lead to a better performance.  Second, although the larger dataset improves performances on average, there are some OARs on which the smaller dataset actually does better, most noticeably, on mandible and optic nerves.  This suggests that there are indeed significant data annotation inconsistencies between different datasets, whose impact on model performance cannot be neglected.  Third, to further check the generalization ability of the model trained with DATASET 1 only, we checked its performance on DATASETS 2-3 and found its performance was generally poor. Altogether, this suggests both annotation quality and data size are important for training deep learning models. How to address inconsistencies in existing datasets is an interesting open question to be addressed in the future.

\begin{table}[ht]
\caption{Performance comparisons of models trained with different datasets.} 
\label{5tab:dataset}
\begin{center}       
\begin{tabular}{|c|c|c|} 
\hline
\rule[-1ex]{0pt}{3.5ex}  Datasets &DATASET 1& DATASET 1,2,3\\
\hline
\rule[-1ex]{0pt}{3.5ex}  Brain Stem &58.60&85.91\\
\hline
\rule[-1ex]{0pt}{3.5ex}  Chiasm &39.93&53.26  \\
\hline
\rule[-1ex]{0pt}{3.5ex}  Mandible &94.16&90.59 \\
\hline
\rule[-1ex]{0pt}{3.5ex}  Opt Ner L &74.62&69.80 \\
\hline
\rule[-1ex]{0pt}{3.5ex}  Opt Ner R &73.77&67.50 \\
\hline
\rule[-1ex]{0pt}{3.5ex}  Parotid L &88.83&87.84 \\
\hline
\rule[-1ex]{0pt}{3.5ex}  Parotid R &87.24&86.15 \\
\hline
\rule[-1ex]{0pt}{3.5ex}  Submand. L &78.56&79.91 \\
\hline
\rule[-1ex]{0pt}{3.5ex}  Submand. R &81.83&80.24 \\
\hline
\rule[-1ex]{0pt}{3.5ex}  Average  &75.28&77.91 \\
\hline
\end{tabular}
\end{center}
\end{table}

\subsection{Limitations}\label{5sec:futurework}
There are a couple of limitations in the current implementation of AnatomyNet. First, AnatomyNet treats voxels equally in the loss function and network structure. As a result, it cannot model the shape prior and connectivity patterns effectively. The translation and rotation invariance of convolution are great for learning appearance features, but suffer from the loss of spatial information.  For example, the AnatomyNet sometimes misclassifies a small background region into OARs (Fig.\ \ref{5fig:vis1},\ref{5fig:vis1_2}). The mis-classification results in a partial anatomical structures, which can be easily excluded if the overall shape information can also be learned. A network with multi-resolution outputs from different levels of decoders, or deeper layers with bigger local receptive fields should  help alleviate this issue. 

Second, our evaluation of the segmentation performance is primarily based on the Dice coefficient. Although it is a common metric used in image segmentation, it may not be the most relevant one for clinical applications. Identifying a new metric in consultation with the physicians practicing in the field would be an important next step in order for real clinical applications of the method.  Along this direction, we quantitatively evaluated the geometric surface distance by calculating the average 95th percentile Hausdorff distance (unit: mm, detailed formulation in \cite{raudaschl2017evaluation}) (Table \ref{5tab:anatomynethd95}). We should note that this metric imposes more challenges to AnatomyNet than other methods operating on local patches (such as the method by Ren et al. \cite{ren2018interleaved}), because AnatomyNet operates on whole-volume slices and a small outlier prediction outside the normal range of OARs can lead to drastically bigger Hausdorff distance. Nonetheless,  AnatomyNet is roughly within the range of the best MICCAI 2015 challenge results on six out of nine anatomies \cite{raudaschl2017evaluation}. Its performance on this metric can be improved by considering surface and shape priors into the model as discussed above \cite{nikolov2018deep,zhu2017adversarial}. 
\begin{table}[ht]
	\caption{Average 95th percentile Hausdorff distance (unit: mm) comparisons with state-of-the-art methods} 
	\label{5tab:anatomynethd95}
	\begin{center}    
		\resizebox{\columnwidth}{!}{   
		\begin{tabular}{|c|c|c|c|} 
			\hline
			\rule[-1ex]{0pt}{3.5ex}  Anatomy & \begin{tabular}{@{}c@{}}\textbf{MICCAI 2015} \\ Range \cite{raudaschl2017evaluation}\end{tabular}  & Ren et al. 2018 \cite{ren2018interleaved} & AnatomyNet\\
			\hline
			\rule[-1ex]{0pt}{3.5ex}  Brain Stem & 4-6 & N/A & 6.42$\pm$2.38 \\
			\hline
			\rule[-1ex]{0pt}{3.5ex} Chiasm & 3-4 & 2.81$\pm$1.56 & 5.76$\pm$2.49 \\
			\hline
			\rule[-1ex]{0pt}{3.5ex} Mand. & 2-13 & N/A & 6.28$\pm$2.21 \\
			\hline
			\rule[-1ex]{0pt}{3.5ex} Optic Ner L & 3-8 & 2.33$\pm$0.84 & 4.85$\pm$2.32 \\
			\hline
			\rule[-1ex]{0pt}{3.5ex} Optic Ner R & 3-8 & 2.13$\pm$0.96 & 4.77$\pm$4.27 \\
			\hline
			\rule[-1ex]{0pt}{3.5ex} Paro. L & 5-8 & N/A & 9.31$\pm$3.32 \\
			\hline
			\rule[-1ex]{0pt}{3.5ex} Paro. R & 5-8 & N/A & 10.08$\pm$5.09 \\
			\hline
			\rule[-1ex]{0pt}{3.5ex} Subm. L & 4-9 & N/A & 7.01$\pm$4.44 \\
			\hline
			\rule[-1ex]{0pt}{3.5ex} Subm. R & 4-9 & N/A & 6.02$\pm$1.78 \\
			\hline
		\end{tabular}
	}
	\end{center}
\end{table}

\section{Conclusion}\label{5sec:conclu}
In summary, we have proposed an end-to-end atlas-free and fully automated deep learning model for anatomy segmentation from head and neck CT images.  We propose a number of techniques to improve model performance and facilitate model training. To alleviate highly imbalanced challenge for small-volumed organ segmentation, a hybrid loss with class-level loss (dice loss) and focal loss (forcing model to learn not-well-predicted voxels better) is employed to train the network, and one single down-sampling layer is used in the encoder. To handle missing annotations, masked and weighted loss is implemented for accurate and balanced weights updating. The 3D SE block is designed in the U-Net to learn effective features.  Our experiments demonstrate that our model provides new state-of-the-art results on head and neck OARs segmentation, outperforming previous models by 3.3\%.  It is significantly faster, requiring only a fraction of a second to segment nine anatomies from a head and neck CT.  In addition, the model is able to process a whole-volume CT and delineate all OARs in one pass. All together, our work suggests that deep learning offers a flexible and efficient framework for delineating OARs from CT images. With additional training data and improved annotations, it would be possible to further improve the quality of auto-segmentation, bringing it closer to real clinical practice. 

\chapter{Conclusion} 
The rapid progress of deep learning on medical image analysis has the potential to revolutionize the common clinical practice including breast cancer mammogram screening, pulmonary cancer CT screening, and radiotherapy in head and neck cancer treatment planning. In this thesis, we introduce how to develop a mass segmentation system based on mammogram in Chapter 2, how to develop a breast cancer screening system in Chapter 3, how to use 3D convolutional neural networks for lung computed tomography (CT) nodule detection and classification system in Chapter 4, how to use electronic medical report (EMR) to further improve the nodule detection system in Chapter 5, and how to design 3D convolutional neural networks to significantly reduce the delineation time of organ at risks (OARs) for radiotherapy planning in Chapter 6. These methods reduce the false positives, improve sensitivities or reduce the delay time in the current clinical practice, and can be potentially great tools for radiologists.

The main challenge for deep learning based medical image analysis is that, deep learning models typically require a lot of training data but the datasets of medical image computing are typically small because 1) patient privacy prevents building large scale datasets, 2) the ground-truth requires experienced radiologists to laboriously annotate or even biopsy results, 3) some anatomies are small naturally. To alleviate the difficulty, we 1) use adversarial concept to generate the hard example to improve the generalization ability of mass segmentation \cite{zhu2017adversarial}, 2) formulate whole mammogram classification as a deep multi-instance learning and propose three inference schemes to employ image-level label for mammogram diagnosis system, instead of requiring detection and segmentation ground-truth \cite{zhu2017deep}, 3) propose a deep learning based probabilistic graphical model employing weak labels from existing in-house EMR to further improve the lung nodule detection system \cite{zhu2018deepem}, 4) proposed a novel encoding framework with weighted hybrid loss of class-distribution loss (Dice loss) and adaptively weighted pixel-wise loss (focal loss) for small anatomy segmentation in radiotherapy planning. 

However, there are still a few challenges in the deep learning based medical image analysis. First, medical images, such as ultrasound images, are typically of low signal-to-noise ratio compared with natural images, and a ultrasound device is cheap. How to design a framework that can bring breakthrough for echo image analysis is a significant task for medical image computing research. Second, medical images, such as CT or MRI, are typically very large. The spatial 3D data requires 3D models naturally which occupy a lot of GPU memory. How to design light models both for training and inference is an interesting direction for the community. Third, the deep learning models are of great capacity and the big complexity which leads to low generalization ability on outlier samples. In addition, abnormal tissues can vary drastically in size and shape, leading to large variations in test images as well. Using the deep generative models to improve the generalization ability of deep models is another interesting direction for medical image computing. 

\clearpage
\phantomsection

\bibliographystyle{abbrv}
\bibliography{thesis}

\captionsetup[figure]{list=no}
\captionsetup[table]{list=no}


\end{document}